\documentclass[10pt,twocolumn,letterpaper]{article}

\usepackage{cvpr}              

\usepackage{graphicx}
\usepackage{amsmath}
\usepackage{amssymb}
\usepackage{booktabs}
\usepackage{enumitem}
\usepackage{times}
\usepackage{epsfig}
\usepackage{enumitem}
\usepackage{subcaption}
\usepackage{multirow}
\usepackage[accsupp]{axessibility} %
\usepackage[dvipsnames]{xcolor}

\usepackage[pagebackref,breaklinks,colorlinks]{hyperref}

\usepackage[capitalize]{cleveref}
\crefname{section}{Sec.}{Secs.}
\Crefname{section}{Section}{Sections}
\Crefname{table}{Table}{Tables}
\crefname{table}{Tab.}{Tabs.}

\def\confName{Random}
\def\confYear{2023}

\begin{document}

\title{Unite and Conquer: Cross Dataset Multimodal Synthesis using Diffusion Models}

\author{Nithin Gopalakrishnan Nair, Wele Gedara Chaminda Bandara and Vishal M. Patel\\
Johns Hopkins University, Baltimore, MD,USA\\
{\tt\footnotesize \{ngopala2, wbandar1 and vpatel36\}@jhu.edu}\\
}

\twocolumn[{%
\renewcommand\twocolumn[1][]{#1}%
\maketitle
\vspace{1.5cm}
\vspace{-3\baselineskip}
\vspace{-3\baselineskip}
\begin{center}
\centering
\setlength{\tabcolsep}{0.5pt}
\captionsetup{type=figure}
{\footnotesize
\renewcommand{\arraystretch}{0.5} 
\begin{tabular}{c c c c c c c c c c c c}
    {(a) \{ImageNet class, Text\} $\longrightarrow$ Generic Scene}\\
    \tabularnewline
        \raisebox{0.01in}{\rotatebox{90}{\scriptsize A road leading
 }}
\raisebox{0.01in}{\rotatebox{90}{\scriptsize into mountains}}
 \includegraphics[width=0.092\linewidth]{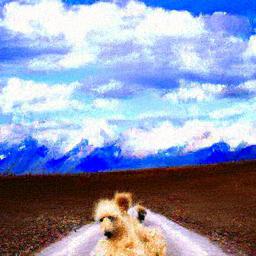}
 \includegraphics[width=0.092\linewidth]{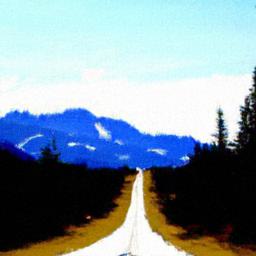}  \includegraphics[width=0.092\linewidth]{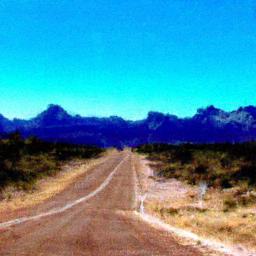}  
\includegraphics[width=0.092\linewidth]{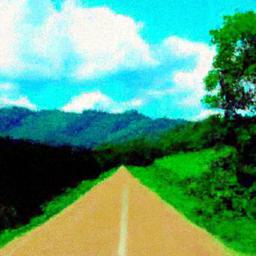}  
\includegraphics[width=0.092\linewidth]{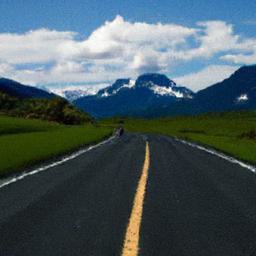} 
\hspace{2mm}
 \includegraphics[width=0.092\linewidth]{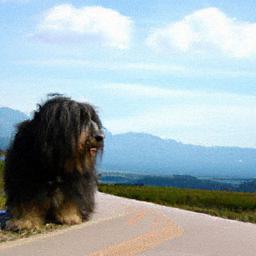}
 \includegraphics[width=0.092\linewidth]{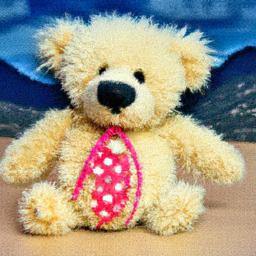}  \includegraphics[width=0.092\linewidth]{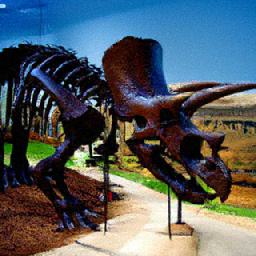}
\includegraphics[width=0.092\linewidth]{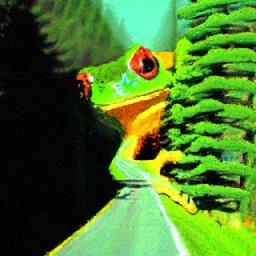}  
\includegraphics[width=0.092\linewidth]{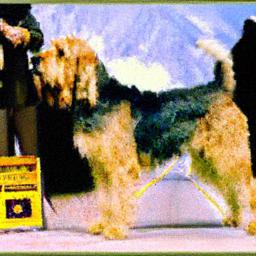}
    \tabularnewline
        \raisebox{0.15in}{\rotatebox{90}{\scriptsize A starry 
 }}
\raisebox{0.1in}{\rotatebox{90}{\footnotesize night sky}}
\hspace{-1.5mm}
 \includegraphics[width=0.092\linewidth]{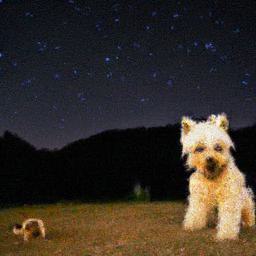}
 \includegraphics[width=0.092\linewidth]{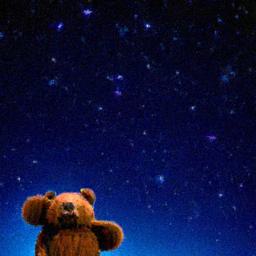}  \includegraphics[width=0.092\linewidth]{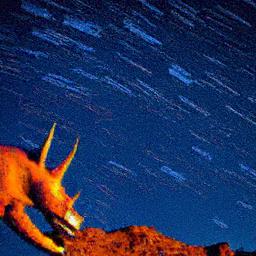}  
\includegraphics[width=0.092\linewidth]{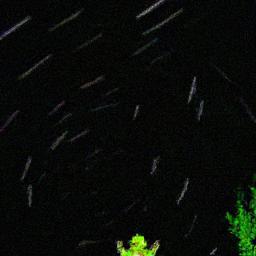}  
\includegraphics[width=0.092\linewidth]{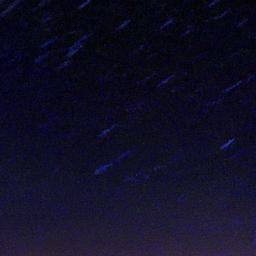} 
\hspace{2mm}
 \includegraphics[width=0.092\linewidth]{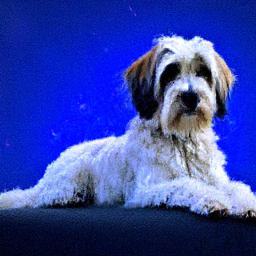}
 \includegraphics[width=0.092\linewidth]{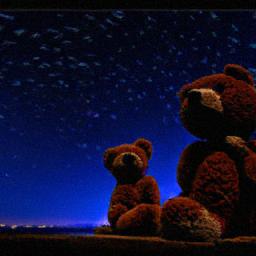}  \includegraphics[width=0.092\linewidth]{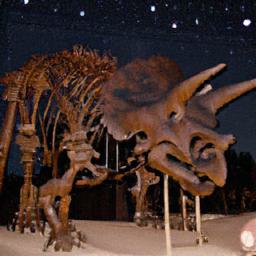}   
\includegraphics[width=0.092\linewidth]{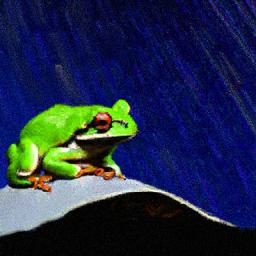}  
\includegraphics[width=0.092\linewidth]{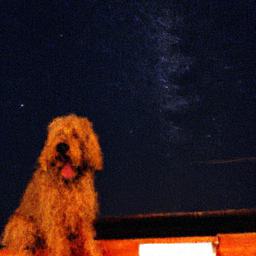}  
    \tabularnewline
    \raisebox{0.01in}{\rotatebox{90}{\scriptsize Forest covered
 }}
\raisebox{0.15in}{\rotatebox{90}{\footnotesize  in snow}}
 \includegraphics[width=0.092\linewidth]{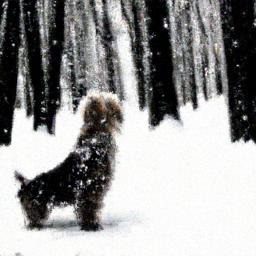}
 \includegraphics[width=0.092\linewidth]{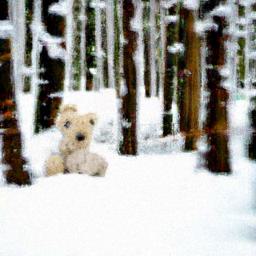}  \includegraphics[width=0.092\linewidth]{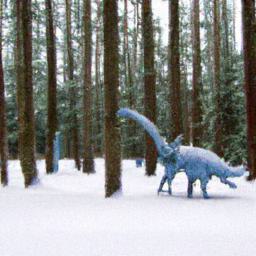}  
\includegraphics[width=0.092\linewidth]{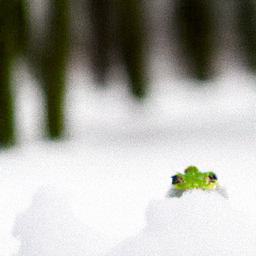}  
\includegraphics[width=0.092\linewidth]{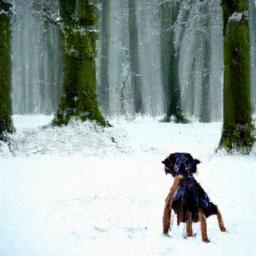} 
\hspace{2mm}
  \includegraphics[width=0.092\linewidth]{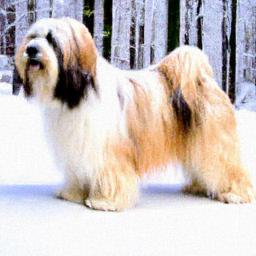}
 \includegraphics[width=0.092\linewidth]{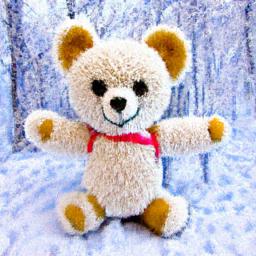}  \includegraphics[width=0.092\linewidth]{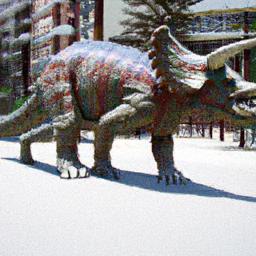}   
\includegraphics[width=0.092\linewidth]{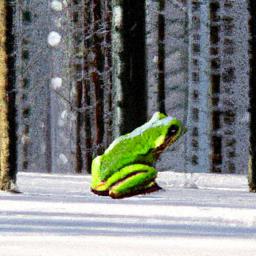}  
\includegraphics[width=0.092\linewidth]{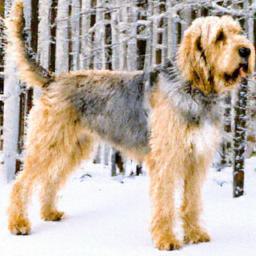} 
    \tabularnewline
{\scriptsize\hspace{10pt}Tibetan terrier\hskip10pt Teddy Bear \hskip10pt   Triceratops \hskip20pt Tree frog \hskip15pt Otterhound \hspace{25pt}Tibetan terrier\hskip10pt Teddy Bear \hskip10pt   Triceratops \hskip20pt Tree frog \hskip15pt Otterhound }\\
\tabularnewline
    {\hspace{-2mm}GLIDE\cite{nichol2021glide} \hspace{215pt}OURS}\\
    \tabularnewline
        {\hspace{30pt}(b) \{\textit{(Face, hair)} semantic labels, Text\}$\longrightarrow$ Facial image \hspace{120pt}(c)
        \{Sketch, Text\}$\longrightarrow$ Facial image}\\
    \tabularnewline
\includegraphics[width=0.11\linewidth]{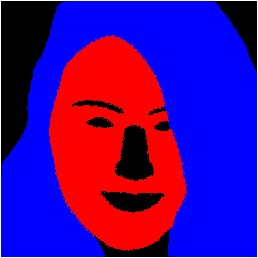}
\hspace{2mm}
\raisebox{0.1in}{\rotatebox{90}{\footnotesize TediGAN\cite{xia2021tedigan}}}
\includegraphics[width=0.11\linewidth]{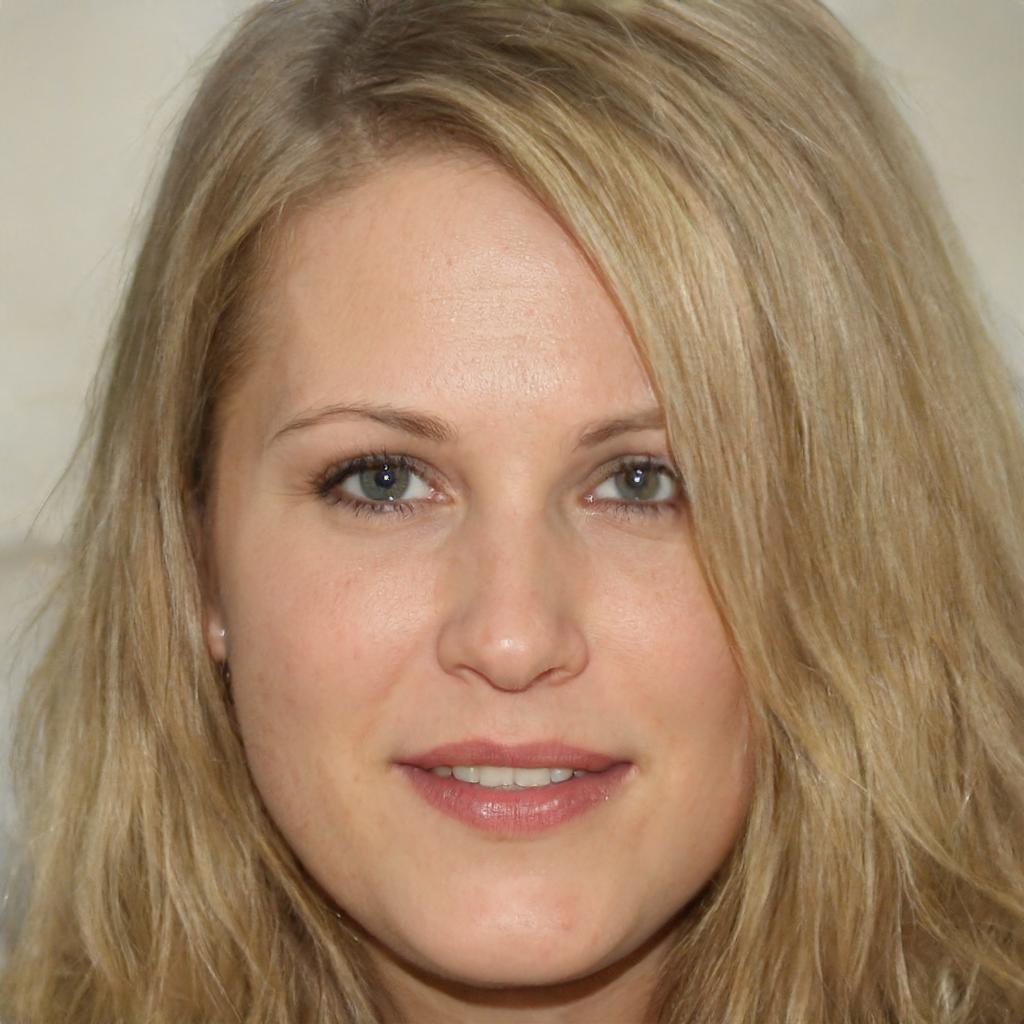}
\includegraphics[width=0.11\linewidth]{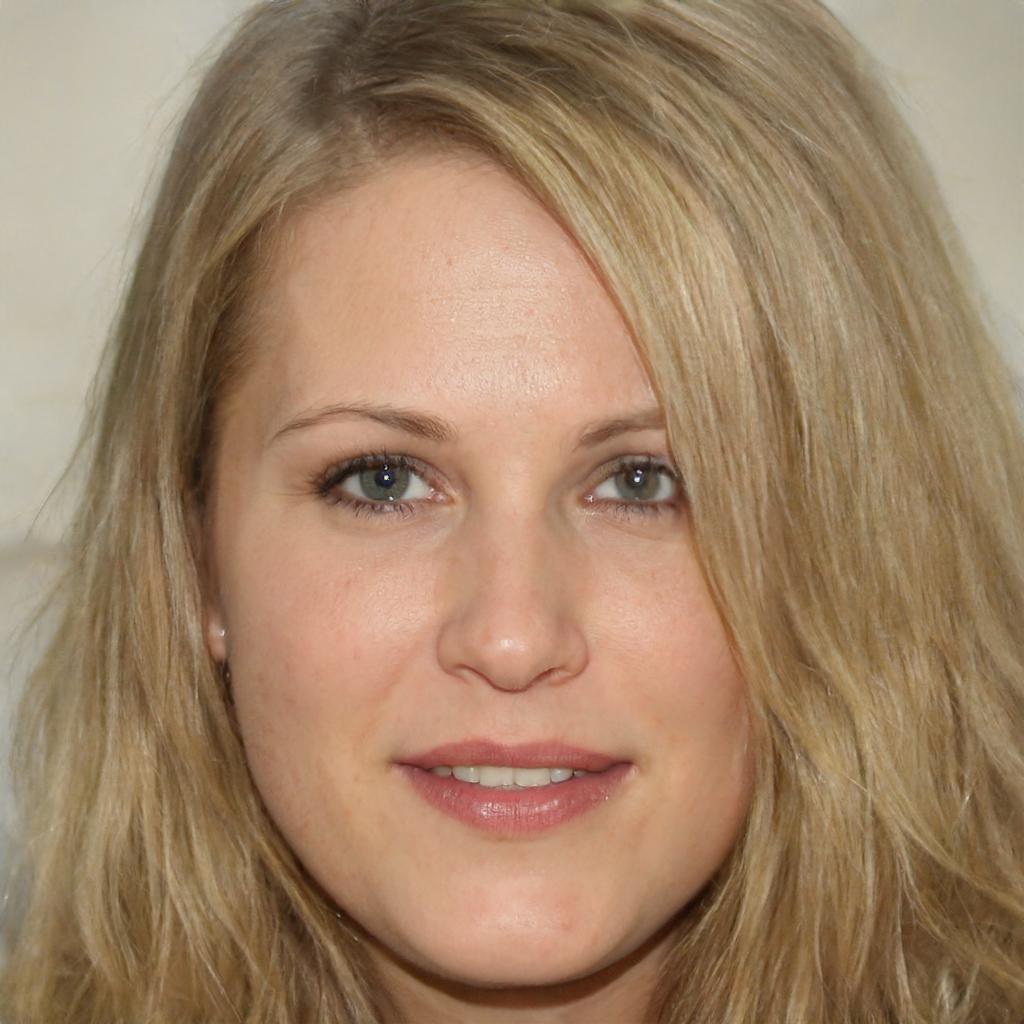}
\hspace{1mm}
\includegraphics[width=0.11\linewidth]{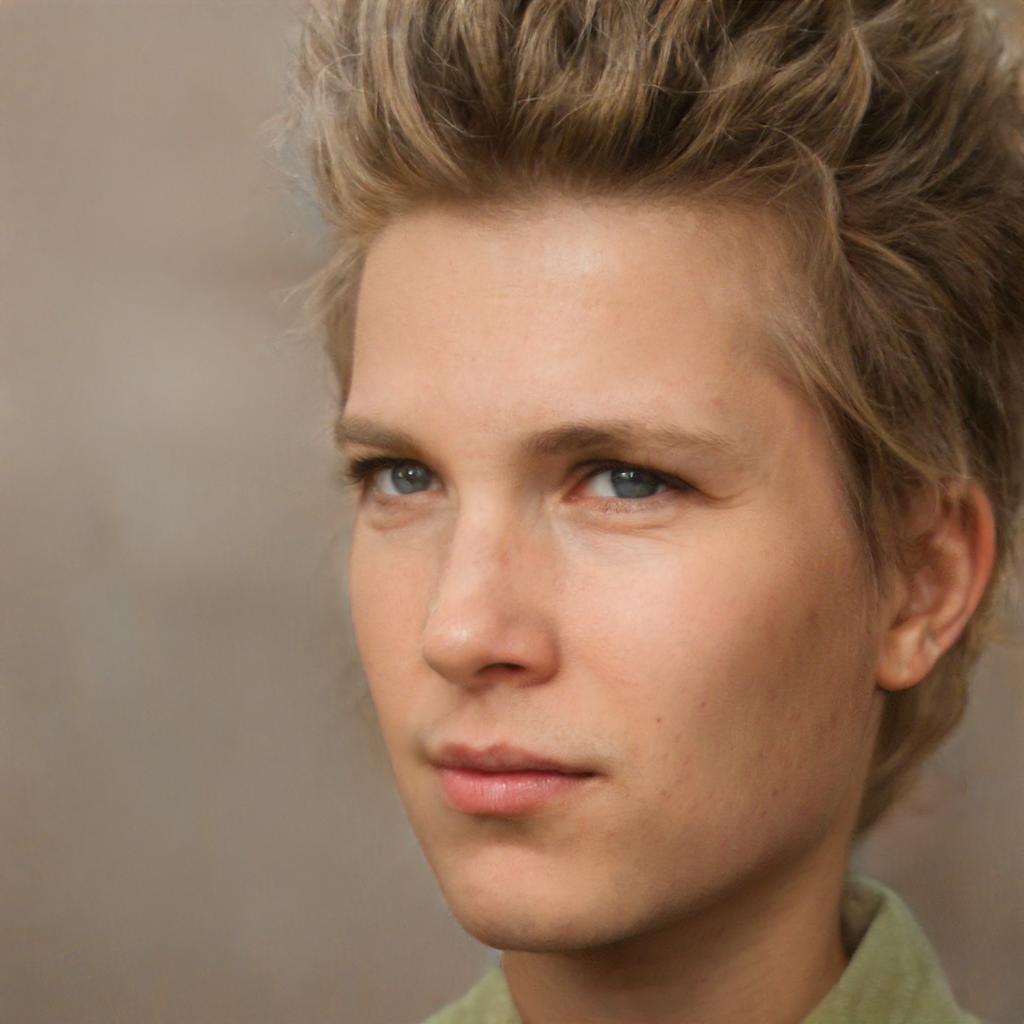}
\includegraphics[width=0.11\linewidth]{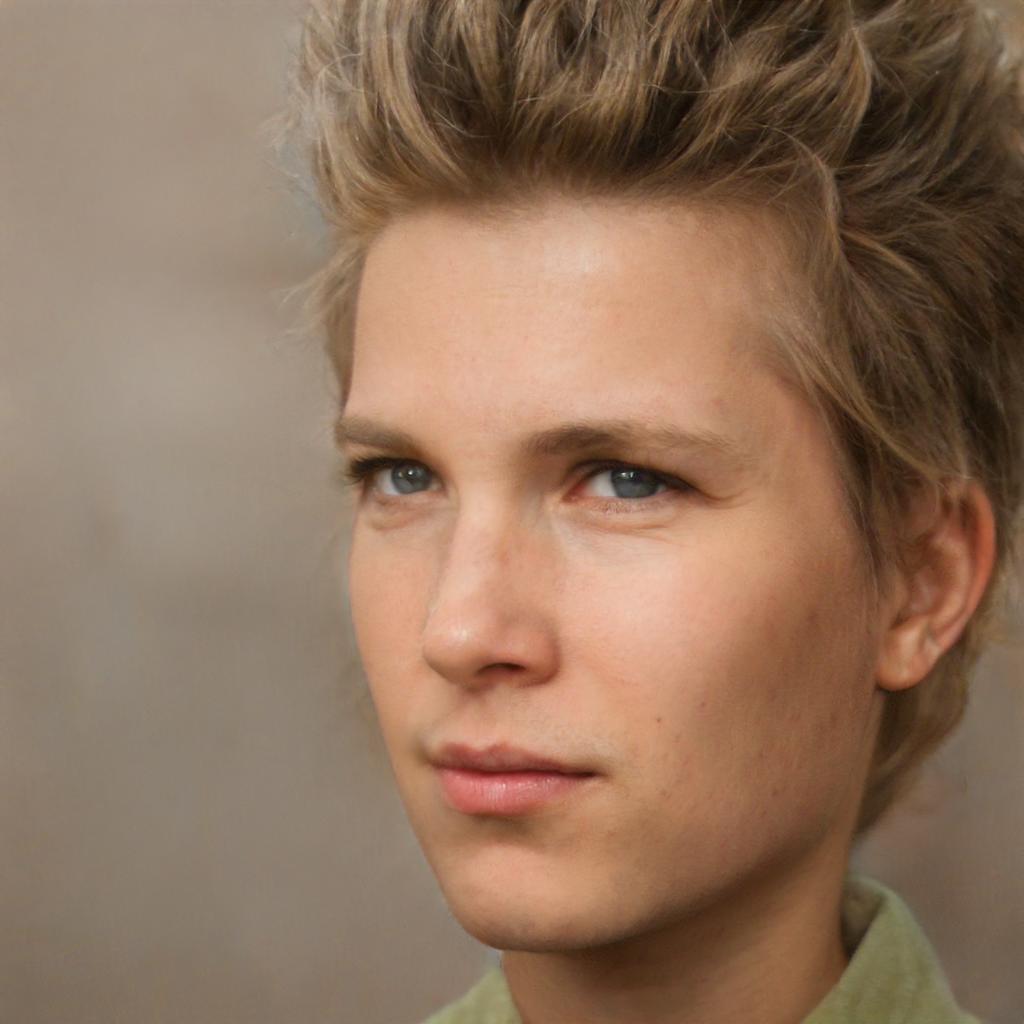}
\hspace{4mm}
\includegraphics[width=0.11\linewidth]{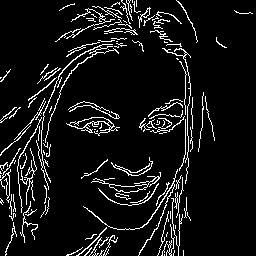}
\includegraphics[width=0.11\linewidth]{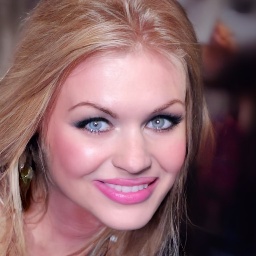}
\includegraphics[width=0.11\linewidth]{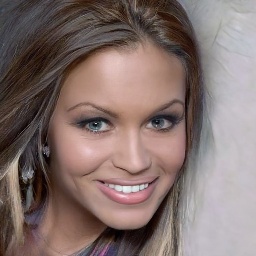}
\tabularnewline
\includegraphics[width=0.11\linewidth]{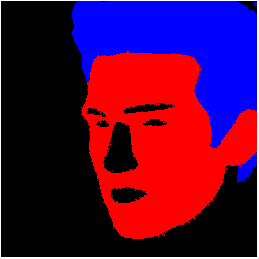}
\hspace{2mm}
\raisebox{0.15in}{\rotatebox{90}{\footnotesize OURS}}
\includegraphics[width=0.11\linewidth]{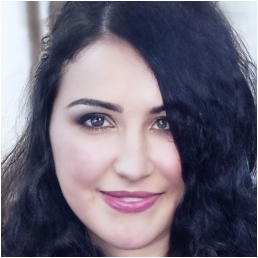}
\includegraphics[width=0.11\linewidth]{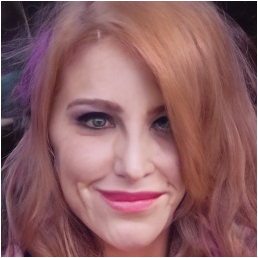}
\hspace{1.5mm}
\includegraphics[width=0.11\linewidth]{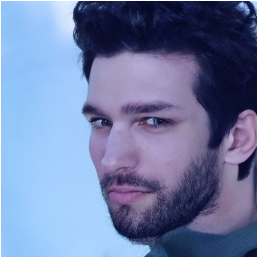}
\includegraphics[width=0.11\linewidth]{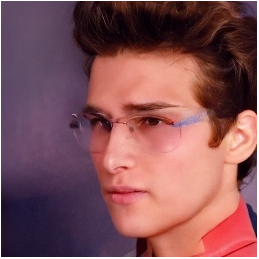}
\hspace{4mm}
\includegraphics[width=0.11\linewidth]{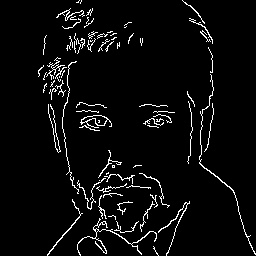}
\includegraphics[width=0.11\linewidth]{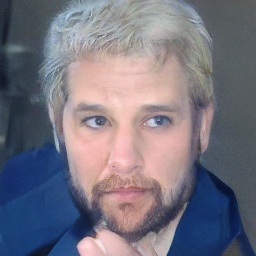}
\includegraphics[width=0.11\linewidth]{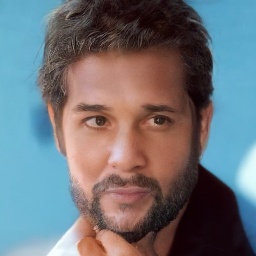}
\tabularnewline
\begin{minipage}{0.11\linewidth}
\vspace{-20pt}
  \hspace{-2mm}\textit{Semantic Label}
\end{minipage}
\hspace{3mm}
\begin{minipage}{0.11\linewidth}
  \textit{\color{violet}This person is chubby and has wavy black hair}
\end{minipage}
\begin{minipage}{0.11\linewidth}
\vspace{-10pt}
 \textit{\color{orange}An old person with brown hair}
\end{minipage}
\hspace{0.1mm}
\begin{minipage}{0.11\linewidth}
\vspace{0pt}
\textit{\color{violet}This person has black hair and wears beard}
\end{minipage}
\begin{minipage}{0.11\linewidth}
\vspace{0pt}
\textit{\color{orange}This person has brown hair and wears eyeglasses}
\end{minipage}
\hspace{3mm}
\begin{minipage}{0.11\linewidth}
\vspace{-20pt}
\hspace{15pt}\textit{Sketch}
\end{minipage}
\begin{minipage}{0.11\linewidth}
\textit{\color{violet}This person has blonde hair and black eyebrows}
\end{minipage}
\begin{minipage}{0.11\linewidth}

\textit{\color{orange}This person has brown hair and dark skin tone}
\end{minipage}
\tabularnewline
\end{tabular}}
\vspace{-0.5\baselineskip}
\hspace{20pt}\captionof{figure}{\textbf{Applications of our method:} \textbf{(a)} Cross-dataset multimodal generation (ImageNet and GLIDE\cite{nichol2021glide}) using two off-the-shelf diffusion models. We bring in new novel classes to a predefined background. We consider five rare classes from the ImageNet classes, and three different text prompts as shown in the figure. For GLIDE\cite{nichol2021glide}, we create a new text prompt by adding an extra sentence utilizing \textit{``and a \{\} in the photo"}. For GLIDE\cite{nichol2021glide}, we sample five images and show the results having the maximum value of CLIP\cite{radford2021learning} correlation value to the text prompt. (b) Multimodal Face generation with three modalities (hair segmentation map, skin segmentation map, and text). (c) Multimodal Generation with two modalities. The text prompt for (b) and (c) is a user-defined text prompt with multiple attributes.}
\label{fig:introfig}
\vspace{-2mm}
\end{center}%
}]
\thispagestyle{empty}

\begin{abstract}
\vspace{-1mm}
Generating photos satisfying multiple constraints finds broad utility in the content creation industry. A key hurdle to accomplishing this task is the need for paired data consisting of all modalities (i.e., constraints) and their corresponding output. Moreover, existing methods need retraining using paired data across all modalities to introduce a new condition. This paper proposes a solution to this problem based on denoising diffusion probabilistic models (DDPMs). Our motivation for choosing diffusion models over other generative models comes from the flexible internal structure of diffusion models. Since each sampling step in the DDPM follows a Gaussian distribution, we show that there exists a closed-form solution for generating an image given various constraints. Our method can \underline{unite} multiple diffusion models trained on multiple sub-tasks and \underline{conquer}  the combined task through our proposed sampling strategy. We also introduce a novel reliability parameter that allows using different off-the-shelf diffusion models trained across various datasets  during sampling time alone to guide it to the desired outcome satisfying multiple constraints.  We perform experiments on various standard multimodal tasks to demonstrate the effectiveness of our approach. More details can be found in \href{https://nithin-gk.github.io/projectpages/Multidiff/index.html}{Multimodal-diff.io}
\end{abstract}

\vspace{-2mm}
\section{Introduction}
Today's entertainment industry is rapidly investing
in content creation tasks~\cite{park2019SPADE,huang2022multimodal}. Studios and companies working on games or animated movies find various applications of photos/videos satisfying multiple characteristics (or constraints) simultaneously. However, creating such photos is time-consuming and requires a lot of manual labor.
This era of content creation has led to some exciting and valuable works like Stable Diffusion \cite{rombach2022high}, Dall.E-2 \cite{ramesh2022hierarchical}, Imagen \cite{saharia2022photorealistic} and multiple other works that can create photorealistic images using text prompts. All of these methods belong to the broad field of conditional image generation \cite{sohl2015deep,radford2015unsupervised}. 
This process is equivalent to sampling a point from the  multi-dimensional space $P(z|x)$ and can be mathematically expressed as:
\setlength{\belowdisplayskip}{0pt} \setlength{\belowdisplayshortskip}{0pt}
\setlength{\abovedisplayskip}{0pt} \setlength{\abovedisplayshortskip}{0pt}
\begin{equation}
\hat{z} \sim P(z|x),
\end{equation}
where $\hat{z}$ denotes the image to be generated based on a condition $x$. 
The task of image synthesis becomes more restricted when the number of conditions increases, but it also happens according to the user's expectations. Several previous works have attempted to solve the conditional generation problem using generative models, such as VAEs \cite{kingma2013auto,radford2015unsupervised} and Generative Adversarial Networks (GANs)~\cite{goodfellow2020generative,sohn2015learning}. However, most of these methods use only one constraint.
In terms of image generation quality, the GAN-based methods outperform VAE-based counterparts. Furthermore, different strategies for conditioning GANs have been proposed in the literature. Among them, the text conditional GANs \cite{tang2021attentiongan,ramesh2021zero, chen2020dmgan,xia2021tedigan} embed conditional feature into the features from the initial layer through adaptive normalization scheme. For the case of image-level conditions such as a sketches or semantic labels, the conditional image is also the input to the discriminator and is embedded with an adaptive normalization scheme~\cite{park2019SPADE,tan2021diverse,wang2018pix2pixHD,oasis}. Hence, a GAN-based method for multimodal generation has multiple architectural constraints \cite{huang2021multimodal}

\begin{figure}[tb]
    \centering
    \includegraphics[width=\linewidth]{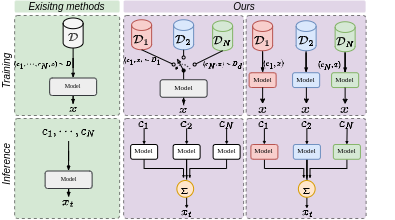}
  \caption{\textbf{An illustration of the difference between the existing multimodal generation approaches~\cite{xia2021tedigan} and the proposed approach.} Existing multimodal methods require training on paired data across all modalities. In contrast, we present two ways that can be used for training: (1) Train with data pairs belonging to different modalities one at a time, and (2) Train only for the additional modalities using a separate diffusion model in case existing models are available for the remaining modalities. During sampling, we forward pass for each conditioning strategy independently and combine their corresponding outputs, hence preserving the different conditions.}
    \label{fig:intro2}
    \vspace{-5mm}
\end{figure}

\begin{figure*}[tbh!]
    \centering
    \includegraphics[width=.95\linewidth]{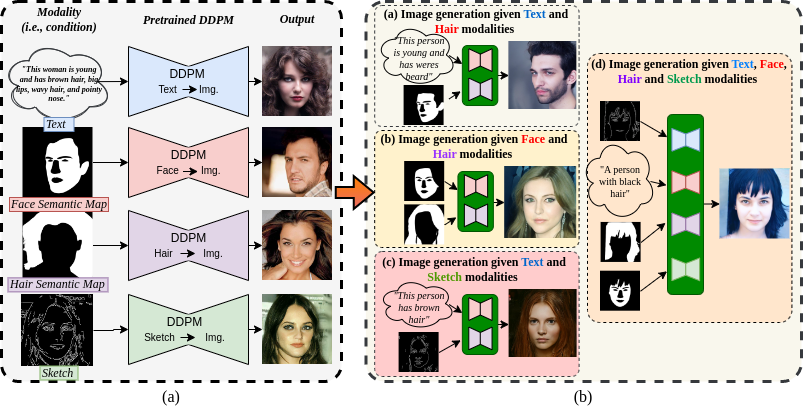}
    \vskip-12pt
    \caption{\textbf{An illustration of our proposed approach.} \textit{During training,} we use diffusion models trained across multiple datasets (we can either train a single model that supports multiple different conditional strategies one at a time or multiple models). \textit{During Inference,} we sample using the proposed approach and condition them using different modalities at the same time.}
    \label{fig:main_fig}
    \label{fig:main}
    \vspace{0mm}
\end{figure*}

A major challenge in training generative models for multimodal image synthesis is the need for paired data containing multiple modalities~\cite{wu2018multimodal,shi2019variational,huang2022multimodal}. This is one of the main reasons why most existing models restrict themselves to one or two modalities~\cite{wu2018multimodal,shi2019variational}. Few works use more than two domain variant modalities for multimodal generation~\cite{huang2021multimodal,xia2021tedigan}. These methods can perform high-resolution image synthesis and require training with paired data across different domains to achieve good results. But to increase the number of modalities, the models need to be retrained; thus they do not scale easily.  Recently, Shi \etal \cite{shi2019variational} proposed a weakly supervised VAE-based multimodal generation method without paired data from all modalities. The model performs well when trained with sparse data. However, if we need to increase the number of modalities, the model needs to be retrained; therefore, it is not scalable. Scalable multimodal generation is an area that has not been properly explored because of the difficulty in obtaining the large amounts of data needed to train models for the generative process.


Recently diffusion models have outperformed other generative models in the task of image generation~\cite{dhariwal2021diffusion,ho2020denoising}. This is due to the ability of diffusion models to perform exact sampling from very complex distributions~\cite{sohl2015deep}. A unique quality of the diffusion models compared to other generative processes is that the model performs generation through a tractable Markovian process, which happens over many time steps. The output at each timestep is easily accessible. Therefore, the model is more flexible than other generative models, and this form of generation allows manipulation of images by adjusting latents~\cite{avrahami2022blended,dhariwal2021diffusion,preechakul2022diffusion}. Various techniques have used this interesting property of diffusion models for low-level vision tasks such as image editing~\cite{avrahami2022blended,kawar2022imagic}, image inpainting~\cite{lugmayr2022repaint}, image super-resolution~\cite{choi2021ilvr}, and image restoration problems~\cite{kawar2022denoising}. 

In this paper, we exploit this flexible property of the denoising diffusion probabilistic models and use it to design a solution to multimodal image generation problems without explicitly retraining the network with paired data across all modalities. Figure \ref{fig:intro2} depicts the comparison between existing methods and our proposed method. Current approaches face a major challenge: the inability to combine models trained across different datasets during inference time \cite{liu2022compositional,ho2020denoising}. In contrast, our work allows users flexibility during training and can also use off-the-shelf models for multi-conditioning, providing greater flexibility when using diffusion models for multimodal synthesis task.
Figure \ref{fig:introfig} visualizes some applications of our proposed approach. As shown in Figure  \ref{fig:introfig}-(a), we use two open-source models~\cite{nichol2021glide,dhariwal2021diffusion} for generic scene creation.  Using these two models, we can bring new novel categories into an image (e.g. Otterhound: the rarest breed of dog). We also illustrate the results showing multimodal face generation, where we use a model trained to utilize different modalities from different datasets. As it can be seen in \ref{fig:introfig}-(b) and (c), our work can leverage models trained across different datasets and combine them for multi-conditional synthesis during sampling. We  evaluate the performance of our method for the task of multimodal synthesis using the only existing multimodal dataset~\cite{xia2021tedigan} for face generation where we condition based on semantic labels and text attributes. We also evaluate our method based on the quality of generic scene generation.


The main contributions of this paper are summarized as follows:


\begin{itemize}[noitemsep]
    \item  We propose a diffusion-based solution for image generation under the presence of multimodal priors.
    \item We tackle the problem of need for paired data for multimodal synthesis by deriving upon the flexible property of diffusion models.
    \item Unlike existing methods, our method is easily scalable and can be incorporated with off-the-shelf models to add additional constraints.
\end{itemize}

\section{Related Work}
 In this section, we describe the existing works on Conditional Image generation using GANs and multimodal image generation using GANs
 
 \subsection{Conditional Image Generation}
 The earliest methods for conditional image generation are based on non-parametric models\cite{zeng2021np}. However, these models often result in unrealistic images. On the other hand, deep learning-based generative models produce faster and better-quality images. Within deep learning based techniques, multiple methods have been proposed in literature for performing conditional image generation,  where the images are generated conditioned on different kinds of input data. have proposed methods where images are generated based on a text prompts\cite{ramesh2021zero}.  \cite{park2019SPADE} et. al proposed a method for conditioning semantic label inputs based on a spatially-adaptive normalization scheme. \cite{tan2021diverse,oasis} Multiple conditional GAN-based techniques also tackle the problem where the conditioning happens from image-level semantics like thermal image \cite{isola2017image,wang2018pix2pixHD}.
 
\subsection{Multimodal Image generation}
Recently multimodal image synthesis has gained significant attention~\cite{huang2021multimodal,shi2019variational,sutter2021generalized,suzuki2016joint,wu2018multimodal,xia2021tedigan,zhang2021m6}, where these method attempts to learn the posterior distribution of an image when conditioned on the prior joint distribution of all the different modalities. The approaches \cite{shi2019variational,suzuki2016joint,wu2018multimodal,zhang2021m6,sutter2021generalized} follow a variational auto-encoder-based solution, where all the input modalities are first processed through their respective encoders to obtain the mean and variance of the underlying Gaussian distributions which are combined using the product of experts in the latent space. Finally, the image is generated by sampling using the new posterior mean and variance. Some methods using GANs for multimodal image synthesis have also gained recent attention. Huang et al\cite{huang2021multimodal} perform multimodal image synthesis using introduces a new Local-Global Adaptive Instance normalization to combine the modalities. Huang et al \cite{huang2021multimodal} also use the product of experts theory to combine encoded feature vectors and use a feature decoder to obtain the final output. TediGAN\cite{xia2021tedigan} uses a StyleGAN \cite{karras2019style} based framework where the different visual-linguistic modalities are combined in the feature space and decoded to obtain the final output. TediGAN allows user-defined image manipulation according to the input of other modalities.

\section{Proposed Method}

\subsection{Langevian score based sampling from a diffusion process}
\label{sec:score}
An alternate interpretation of denoising diffusion probabilistic models is denoising score-based approach \cite{song2019generative}, where the sampling during inference is performed using stochastic gradient langevian dynamics \cite{welling2011bayesian}. Here a network is used to compute the score representing the gradient of likelihood of the data. The sampling during the reverse timestep  can be represented by \cite{song2019generative}:
\begin{multline}
    z_{t-1} \leftarrow{} \frac{1}{\sqrt{1-\beta_t}} \left( z_t -  \beta_t s_{\theta}(z_t, x, t) \right) + \sigma_t^2 \boldsymbol{\eta},
    \label{eq:score}
\end{multline}
where $\boldsymbol{\eta} = \mathcal{N}(\boldsymbol{0}, \boldsymbol{I})$, $z_t$ is the sample at timestep $t$ and $x$ is the condition. The score value  $s_{\theta}(\cdot)$ is  given by,
\begin{equation}
    s_{\theta}(z_t,t) =\nabla_x \log P(z_t|x) =\frac{\epsilon_{\theta}(z_t,x,t)}{\sqrt{1-\bar{\alpha}_t}},
\end{equation}
 where $\epsilon_{\theta}$ is the output from the denoising network.
 

\subsection{Multimodal conditioning using diffusion models}
In regular conditional denoising diffusion models~\cite{saharia2021image}, the input image (i.e., the condition) is concatenated  with the sampled noise when passing through the network. When multiple modalities are present, the trivial solution of finding the conditional distribution is by concatenating all the $N$ modalities with the noisy image. However, if we want to improve the functionality of the trained network by by adding a new modality, the whole model needs to be retrained with all $N+1$ modalities. Instead, we propose an alternative way to achieve this goal. Let $(z, x_i)$ denote a point in the space of the images of a particular domain and $p(z|x_i)$ denote the distribution of the predicted image $z$ based on the modality $x_i$.  Let ${\bf{X}}=\{x_1,,x_2,..x_N\}$. Let the distribution of the image conditioned on all modalities be denoted by $P(z|{\bf{X}})$
and the distribution of the image conditioning on the individual modalities be $P(z|x_i)$.  Assuming that all the modalities are statistically independent,
\begin{multline}
    P(z|{\bf{X}})=\frac{P(z)}{P({\bf{X}})}\prod_{i=1}^{N}P(x_i|z) 
    = KP(z) \frac{\prod_{i=1}^{N}P(z|x_i)}{\prod_{i=1}^{N}P(z)},
    \label{eq:POE1}
\end{multline}
where $K$ is a term which is independent from $z_t$. Assuming the individual distributions $P(z|x_i)$ and $P(z)$ follow a Gaussian distribution, the distribution $P(z|{\bf{X}})$ will also follow a Gaussian distribution.
Now, let's assume that $N$ diffusion models are trained to generate samples from the distributions $P(z|x_i)$ conditioned on each modality $x_i$ separately. 
 We have $N$ modalities from where the unconditional distribution could be computed.  However, how good each model can model the unconditional distribution is not certain, hence we utilize the generalized product of experts rule \cite{cao2014generalized} to compute the effective unconditional density as
 \begin{equation}
     P(z) =\prod_{i=1}^{N}P^{a_i}_{\delta_i}(z|\phi),
 \end{equation}
where $a_i$ is the confidence factor of each individual distribution with a null condition to modelling the overall unconditional density. To preserve the effective variance so that the reverse diffusion process still holds,  we set the constraint $  \sum_{i=1}^N a_i =1$ .
  As mentioned in Section \ref{sec:score}, we can use stochastic gradient Langvein sampling based sampling to sample from the conditional distribution $P(z|{\bf{X}})$ . Please note that we are imposing an assumption that the diffusion process for each of the individual modalities have the same variance schedule. Hence the score-based derivations are valid and the effective diffusion process has the same variance schedule
 \begin{equation}
\begin{split}
 \nabla_{z_t}\log  (z_t|{\bf{X}}) =   \\
\nabla_{z_t}\log  \biggl((\prod_{i=1}^{N}P_{\delta_i}^{a_i}(z_t|\phi))\frac{\prod_{i=1}^{N}P_{\delta_i}(z_t|x_i)}{\prod_{j=1}^{N}\prod_{i=1}^{N}P_{\delta_i}^{a_i}(z_t|\phi)} \biggr)=\\
     \sum_{i=1}^N \biggl( \nabla_{z_t}\text{log} P_{\delta_i}(z_t|x_i) - \sum_{j\neq i}a_j \nabla_{z_t}\text{log}  P_{\delta_j}(z_t|\phi)\biggr),
\end{split}
\end{equation}
where $\delta_i$ denotes the parameters of the individual distribution densities and $\phi$ denotes the null condition.  $b_{ij}$ denotes the confidence of a  paramteric model to estimate the unconditional density of another model. Hence the effective score when conditioned on all the modalities can be represented in terms of scores of the individual conditional distribution as well as the score of the unconditional model. Hence the effective score $s_{c}$ is given by:
\begin{align}
    s_{c} = \frac{\epsilon_c}{\sqrt{1-\bar{\alpha}_t}},
\end{align}
\begin{multline}
   \epsilon_{c}= \epsilon_{\theta}(z_t,{\bf{X}} ,t) =\sum_{i=1}^N a_i \epsilon_{i}(z_t,\phi,t) +\\
   \sum_{i=1}^N\biggl( \epsilon_{i}(z_t,x_i,t)-
   \sum_{j=i}^N a_{j}\epsilon_j (z_t,\phi,t)\biggr),
   \label{eq:multimodal1}
\end{multline}
where $\epsilon_{\theta}(z_t,x_i,t)$ denotes the output prediction of the individual conditional networks and $\epsilon_{\theta}(z_t,t)$ is the prediction of the unconditional network. After computing the effective score, sampling could be performed using equation by,
\begin{multline}
    z_{t-1} \leftarrow{} \frac{1}{\sqrt{1-\beta_t}} \left( z_t - \frac{ \beta_t}{\sqrt{1-\bar{\alpha}_t}}\epsilon_c \right) + \sigma_t^2 \boldsymbol{\eta},
\end{multline}
where $\boldsymbol{\eta} \sim\mathcal{N}(\boldsymbol{0}, \boldsymbol{I})$. An additional parameter for stringer conditions can be incorporated to  \eqref{eq:multimodal1} using Generalized product of experts \cite{cao2014generalized} (proof in supplementary). Hence giving importance to some modalities over the others by partial weighting to the scores estimated by each modality as follows:
\begin{multline}
   \epsilon_{c}=
   \sum_{i=1}^N w_i \epsilon_{i}(z_t,x_i,t)- 
  (\sum_{i=1}^N w_i-1) \sum_{j=1}^N a_j \epsilon_{j}(z_t,\phi, t).\\
 ,   w_i\geq 1
   \label{eq:multimodal}
\end{multline}

Recently Ho \textit{et al.}\cite{ho2021classifier} proposed a method for sampling from a particular conditional distribution without the need of an explicit classifier as used by previous methods\cite{dhariwal2021diffusion}. According to this method, given a model trained for modeling a conditional distribution $p(z|c)$, and with the same model trained for an modelling an unconditional distribution $p(z)$, The effective score for generating samples could be obtained using,
\begin{equation}
    \hat{\epsilon}(z_t,c,t)=(1+w) \cdot \epsilon(z_t,c,t)-w \cdot \epsilon(z_t,t),
    \label{eq:classifierfree}
\end{equation}
where $w$ is a scalar. 
By comparing equations \eqref{eq:multimodal} and \eqref{eq:classifierfree}, we can see that equation \eqref{eq:classifierfree}, is the case of MCDM with all modalities being the same. 
\section{Experiments}

In this section we describe in detail the experiments performed  and reason out the choice on experimental setup. We consider different multimodal settings for our network and  evaluate the performance quantitatively for  and multimodal image generation on the CelebA and FFHQ datasets.
For multimodal image semantics to face generation, we choose networks that can perform semantic labels to face generation retrain them for scratch for the different scenarios considered. As detailed in earlier sections, one major challenge of multimodal image generation is the lack of paired data across all modalities. To extend existing approaches for the case of multimodal generation, two approaches could be followed. The first is to take a model trained for a particular conditioning modality and finetune it for another modality. In the second approach, the training corpus becomes the combination of the datasets with individual modalities and different iterations could model training sees a different modality and its corresponding image.  
\begin{figure*}[htb!]
    \centering
    \setlength{\tabcolsep}{0.5pt}
    {\small
    \renewcommand{\arraystretch}{0.5} 
    \begin{tabular}{c c c c c c c c c c c c}
    \captionsetup{type=figure, font=scriptsize}
    \hspace{-2mm}
    \raisebox{0.15in}{\rotatebox{90}{\scriptsize \emph{A yellow}
 }}
\raisebox{0.1in}{\rotatebox{90}{\footnotesize \emph{flower field}}}
 \includegraphics[width=0.093\linewidth]{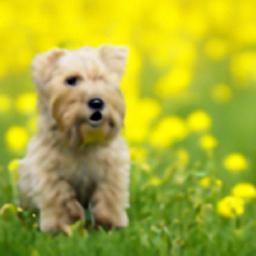}
 \includegraphics[width=0.093\linewidth]{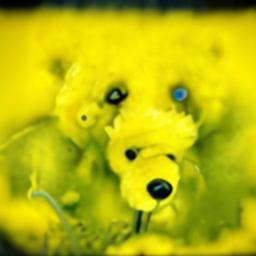} \includegraphics[width=0.093\linewidth]{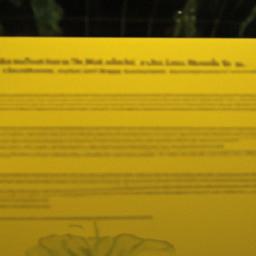}
\includegraphics[width=0.093\linewidth]{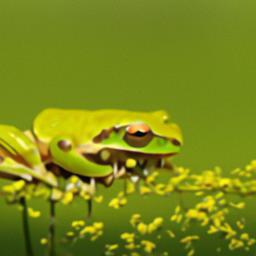}
\includegraphics[width=0.093\linewidth]{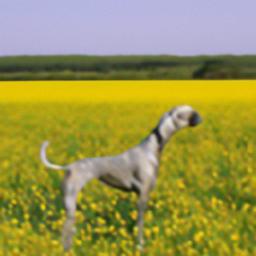}
\hspace{2mm}
 \includegraphics[width=0.093\linewidth]{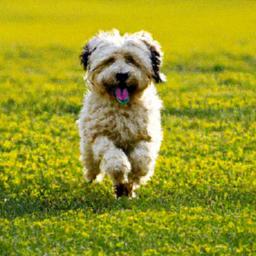}
 \includegraphics[width=0.093\linewidth]{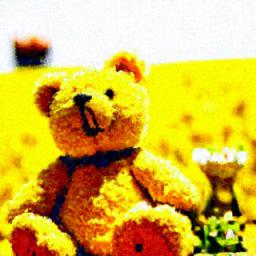}  \includegraphics[width=0.093\linewidth]{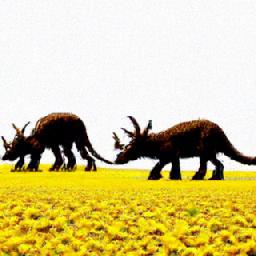}
\includegraphics[width=0.093\linewidth]{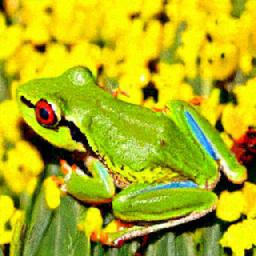}  
\includegraphics[width=0.093\linewidth]{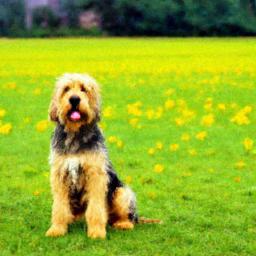}
    \tabularnewline
    \hspace{-1.5mm}
        \raisebox{0.15in}{\rotatebox{90}{\scriptsize  \emph{Photo of a}
 }}
\raisebox{0.20in}{\rotatebox{90}{\footnotesize \emph{beach}}}
 \includegraphics[width=0.093\linewidth]{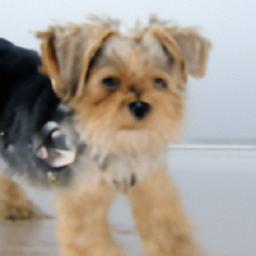}
 \includegraphics[width=0.093\linewidth]{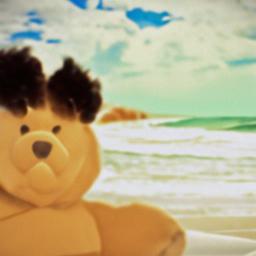} \includegraphics[width=0.093\linewidth]{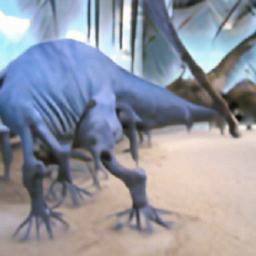} 
\includegraphics[width=0.093\linewidth]{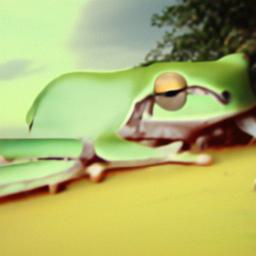}
\includegraphics[width=0.093\linewidth]{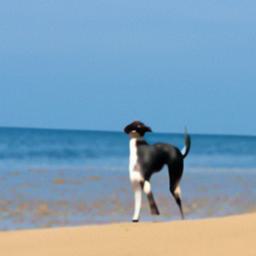}
\hspace{2mm}
  \includegraphics[width=0.093\linewidth]{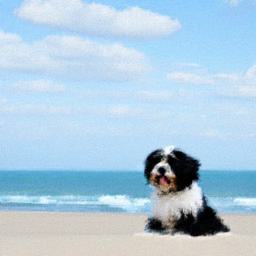}
 \includegraphics[width=0.093\linewidth]{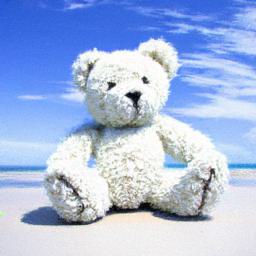}  \includegraphics[width=0.093\linewidth]{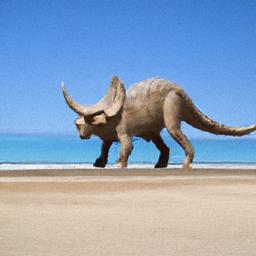}   
\includegraphics[width=0.093\linewidth]{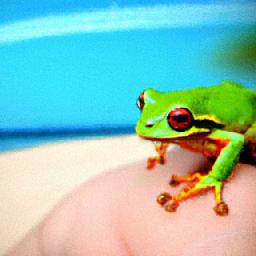}  
\includegraphics[width=0.093\linewidth]{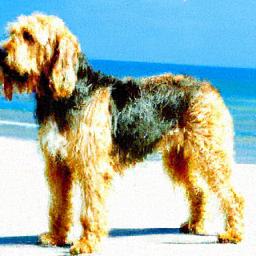} 
    \tabularnewline
{\scriptsize\hspace{10pt}Tibetan terrier\hskip10pt Teddy Bear \hskip10pt   Triceratops \hskip20pt Tree frog \hskip15pt Otterhound \hspace{25pt}Tibetan terrier\hskip10pt Teddy Bear \hskip10pt   Triceratops \hskip20pt Tree frog \hskip15pt Otterhound }\\\\
\tabularnewline
\vspace{2mm}
\begin{minipage}{0.5\linewidth}
\hspace{3cm}
CompGen (Uni-modal)\cite{liu2022compositional}
\end{minipage}
\begin{minipage}{0.5\linewidth}
\hspace{4cm}
Ours (Multi-Modal)
\end{minipage}
\end{tabular}}
\vspace{-0.4cm}
\hspace{20pt}\captionof{figure}{ \textbf{Qualitative results for cross dataset multimodal generation} (ImageNet and CompGen\cite{liu2022compositional}) using two off-the-shelf diffusion models.}
\label{fig:facelrfid}
\vspace{-2mm}
\end{figure*}%

\begin{figure*}[t!]
    \centering
    \begin{subfigure}[t]{0.137\linewidth}
      \captionsetup{justification=centering, labelformat=empty, font=scriptsize}
      \includegraphics[width=1\linewidth]{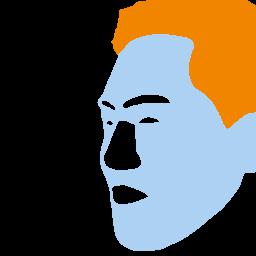}

      \includegraphics[width=1\linewidth]{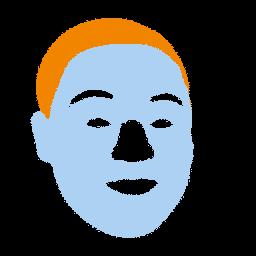}
      \caption{Semantic label}
    \end{subfigure}
    \begin{subfigure}[t]{0.137\linewidth}
      \captionsetup{justification=centering, labelformat=empty, font=scriptsize}
      \includegraphics[width=1\linewidth]{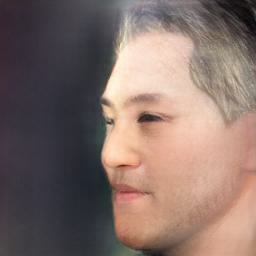}
      \includegraphics[width=1\linewidth]{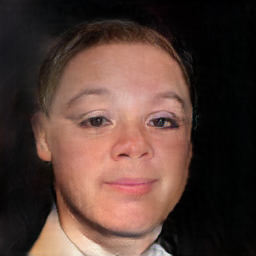}
      \caption{SPADE\cite{park2019SPADE}}
    \end{subfigure}
    \begin{subfigure}[t]{0.137\linewidth}
      \captionsetup{justification=centering, labelformat=empty, font=scriptsize}
      \includegraphics[width=1\linewidth]{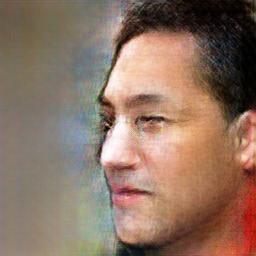}
      \includegraphics[width=1\linewidth]{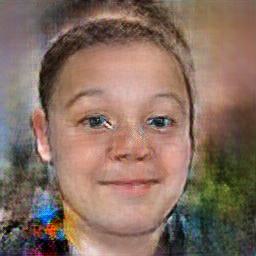}
      \caption{PIX2PIXHD\cite{wang2018pix2pixHD}}
    \end{subfigure}
    \begin{subfigure}[t]{0.137\linewidth}
      \captionsetup{justification=centering, labelformat=empty, font=scriptsize}
      \includegraphics[width=1\linewidth]{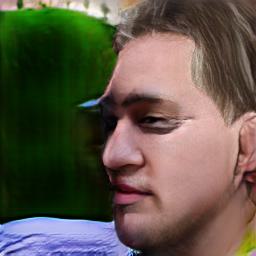}
      \includegraphics[width=1\linewidth]{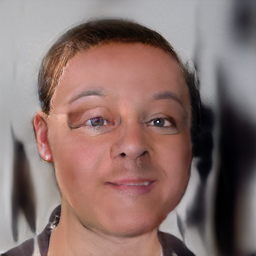}
      \caption{INADE\cite{tan2021diverse}}
    \end{subfigure}
    \begin{subfigure}[t]{0.137\linewidth}
      \captionsetup{justification=centering, labelformat=empty, font=scriptsize}
      \includegraphics[width=1\linewidth]{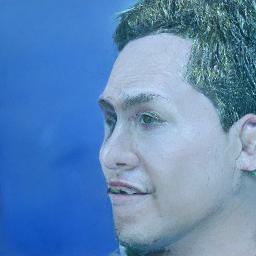}
      \includegraphics[width=1\linewidth]{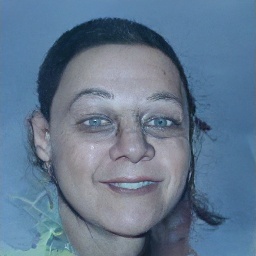}
      \caption{DDPM\cite{ho2020denoising}}
    \end{subfigure}
    \begin{subfigure}[t]{0.137\linewidth}
      \captionsetup{justification=centering, labelformat=empty, font=scriptsize}
      \includegraphics[width=1\linewidth]{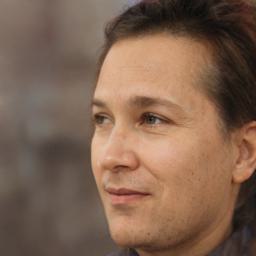}
      \includegraphics[width=1\linewidth]{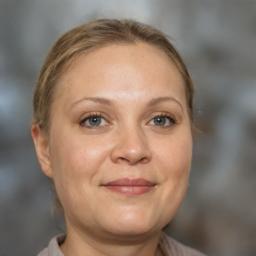}
      \caption{TediGAN\cite{xia2021tedigan}}
    \end{subfigure}
    \begin{subfigure}[t]{0.137\linewidth}
      \captionsetup{justification=centering, labelformat=empty, font=scriptsize}
      \includegraphics[width=1\linewidth]{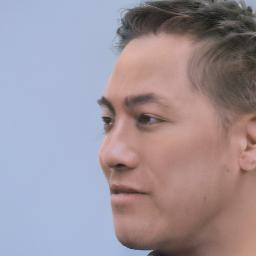}
      \includegraphics[width=1\linewidth]{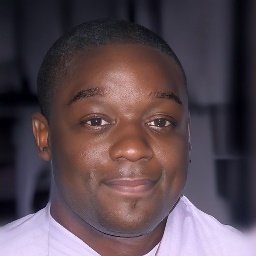}
      \caption{OURS}
    \end{subfigure}
    \vspace{-3mm}    \caption{\textbf{Qualitative comparisons for semantic to face generation.} In this case, a single model is trained by alternating different input datasets across different iterations. During Inference time all the modalities are taken from a single dataset and the proposed sampling technique is used.}
    \label{fig:facesematic}
  \end{figure*}

\subsection{Multimodal Face Generation} To create a criteria to quantitatively evaluate the existing works and our method for multimodal generation, we follow the works \cite{wu2018multimodal} where complementary information comes from semantic labels of different portions of the same image. To make the problem even more generic, during training we choose the  semantic labels for different regions from different datasets. For multimodal semantics to face generation, we use the CelebA-HQ dataset \cite{karras2017progressive} and the FFHQ dataset \cite{karras2019style}. We choose the semantic labels of hair from the CelebA dataset and the semantic labels of skin from the FFHQ dataset for the training process. The semantic labels for the skin and the hair are obtained using the face parser released by Zheng et al. \cite{zheng2021farl}. For training our dataset, we choose 27,000 images from the CelebA-HQ dataset and 27,000 images from the FFHQ dataset, and train the corresponding individual diffusion models with these semantic labels as well as attributes. We test our method using 3,000 images chosen from the CelebA-HQ dataset and 3,000  images from the FFHQ dataset. During evaluations, we utilize the semantic labels of hair as well as skin from the same image to illustrate how well our method is able to generalize. As for the evaluations metrics we choose the FID score \cite{heusel2017gans} and the LPIPS score \cite{zhang2018perceptual}  to evaluate the sample diversity and quality of facial images generated. To evaluate the structural similarity of the produced result with the actual image we use the SSIM score. Finally to see how close the input semantic labels are to the ones produced by our method, we obtain the parsed masks of the generated images and compute the  mean intersection over union over all classes (mIoU) and F1 precision scores of the semantic images obtained from the reconstructed image and the input semantic image and report the mean for all testing images. We set the value of $a_i=1/N$ and $w_i=1$ for all the experiments. Here N denotes the number of modalities used. For training the sketch to face model, we utilized Canny egde detector\cite{ding2001canny} to extract the edges. For text to face generation, we utilize the FFHQ dataset and trained a model based on the extracted face embeddings using FARL \cite{zheng2021farl}. The reader is referred to the supplementary material for more details on the implementation of the individual multimodal face generation sub-parts.

\begin{table}[!t]
\begin{center}
\caption{\textbf{Quantitative results for multimodal semantic labels to face generation on CelebA dataset.} The combined training based strategy and fine-tuning based training strategy are shown in the corresponding sections. $(\uparrow)$/  $(\downarrow)$ represents higher/ lower the metric lower the metrics, the better respectively.
}
\label{table:celeba_semantics}
\vspace{-2mm}
\scalebox{0.7}{
\begin{tabular}{|c| c | c c c c c|}
\toprule[0.15em]
   \textbf{Type}&\textbf{Method} & FID$\textcolor{black}{\downarrow}$ & LPIPS$\textcolor{black}{\downarrow}$& SSIM$\textcolor{black}{\uparrow}$ & mIoU$\textcolor{black}{\uparrow}$  & F1~$\textcolor{black}{\uparrow}$      \\
\midrule[0.15em]
\multirow{5}{*}{Fine-Tuning} &SPADE\cite{park2019SPADE} &131.91 & 0.638 & 0.316 & 0.605&0.694 \\
&OASIS\cite{oasis}  & 118.67 & 0.624 & 0.318 & 0.579&0.663  \\
&PIX2PIXHD\cite{wang2018pix2pixHD}& 153.19 & 0.611 & 0.282 & 0.716&0.819   \\
&INADE\cite{tan2021diverse} &125.43& 0.632 & 0.279 & 0.881&0.932 \\
\midrule[0.15em]
\multirow{5}{*}{Combined} &SPADE\cite{park2019SPADE} & 89.29 & \textbf{0.523} & 0.376 & 0.890&0.936 \\
&OASIS\cite{oasis}  & 71.20 & 0.571 & 0.323 & 0.792&0.870  \\
&PIX2PIXHD\cite{wang2018pix2pixHD}& 73.32 & 0.512 & 0.373 & 0.872&0.925   \\
&INADE\cite{tan2021diverse} &54.27& 0.552  & 0.332& 0.887&0.933 \\
\midrule[0.15em]

&TediGAN\cite{xia2021tedigan} &69.51 & 0.4823 & 0.417 & 0.834&0.905 \\
&OURS &\textbf{26.09} & 0.519 & \textbf{0.416} & \textbf{0.911}&\textbf{0.948}  \\

\bottomrule[0.1em]
\end{tabular}}
\end{center}
\vspace{-1.5em}
\end{table}
\subsection{Generic Scenes Generation}
\label{sec:generic}
To show that our method is generalized and could combine existing works to make their generation process more powerful, we use the text-to-image generation-based diffusion model released by GLIDE and an ImageNet class conditional generation model. We perform a combined multimodal generation task where we choose the GLIDE model to decide the background in the image and use the Imagenet model to bring specific objects to the image. During evaluations, we generate images using text prompts that contain a scene information as well as an ImageNet object using an and conditioning. We generate 500 such images for all methods and evaluate how close the images are to the text prompts using the average clip distance between the image embeddings and the embeddings of the input text prompt. We make use of nonreference quality metrics NIQE to evaluate the method. We also present the accuracy using the state-of-the-art Imagenet classifier\cite{tan2019efficientnet} to detect whether the class is present in the image. We set  $w_i=5$ for all the experiments. 

 \begin{figure}[tb!]
    \centering
    \setlength{\tabcolsep}{0.5pt}
    {\small
    \renewcommand{\arraystretch}{0.5} 
    \begin{tabular}{c c c c c c c c c c}
    \captionsetup{type=figure, font=scriptsize}
    \raisebox{0.1in}{\rotatebox{90}{\small \emph{$0.0$}
 }}
  \includegraphics[width=0.135\linewidth]{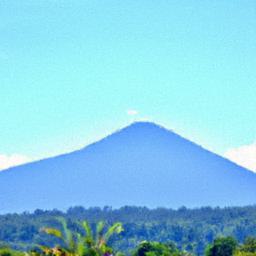}
  \includegraphics[width=0.135\linewidth]{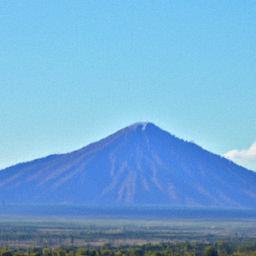}
  \includegraphics[width=0.135\linewidth]{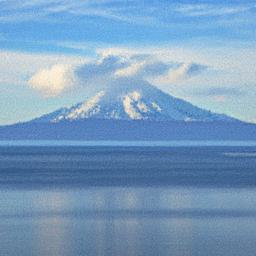}
  \includegraphics[width=0.135\linewidth]{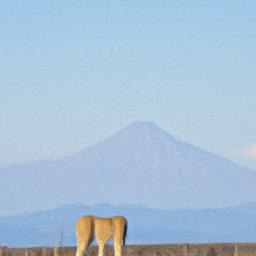}
  \includegraphics[width=0.135\linewidth]{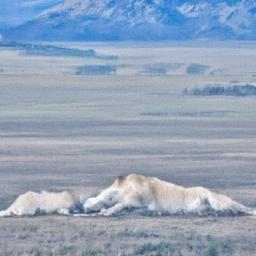}
  \includegraphics[width=0.135\linewidth]{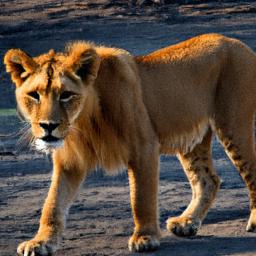}
  \includegraphics[width=0.135\linewidth]{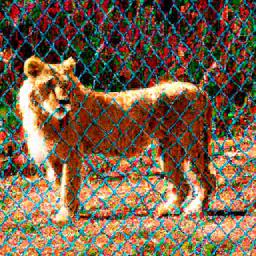}
 \tabularnewline
     \raisebox{0.1in}{\rotatebox{90}{\small \emph{$0.2$}
 }}
  \includegraphics[width=0.135\linewidth]{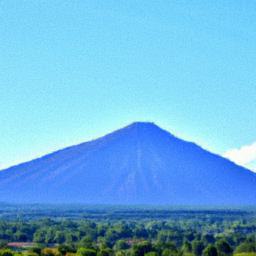}
  \includegraphics[width=0.135\linewidth]{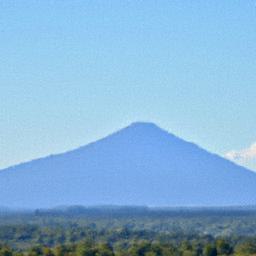}
  \includegraphics[width=0.135\linewidth]{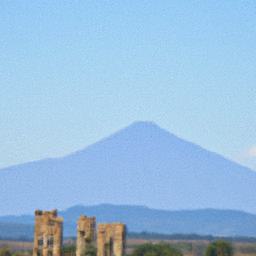}
  \includegraphics[width=0.135\linewidth]{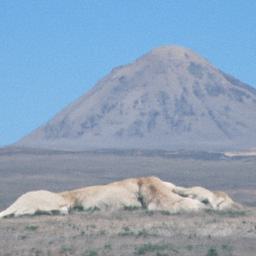}
  \includegraphics[width=0.135\linewidth]{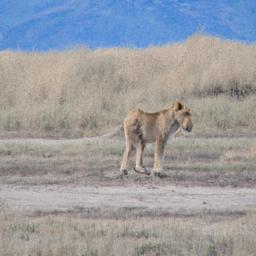}
  \includegraphics[width=0.135\linewidth]{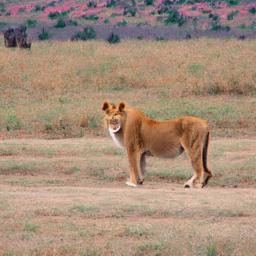}
  \includegraphics[width=0.135\linewidth]{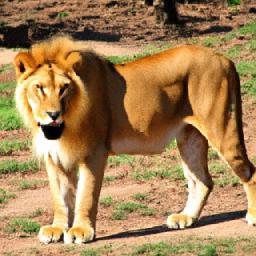}
\tabularnewline
    \raisebox{0.1in}{\rotatebox{90}{\small \emph{$0.4$}
 }}
  \includegraphics[width=0.135\linewidth]{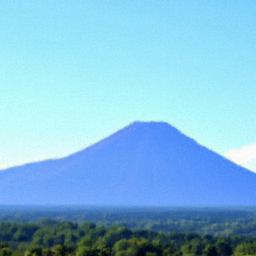}
  \includegraphics[width=0.135\linewidth]{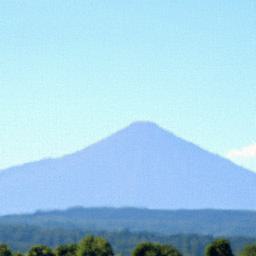}
  \includegraphics[width=0.135\linewidth]{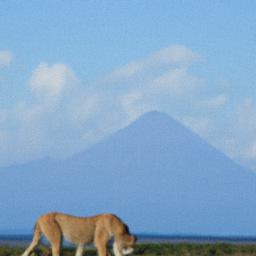}
  \includegraphics[width=0.135\linewidth]{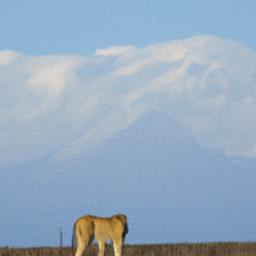}
  \includegraphics[width=0.135\linewidth]{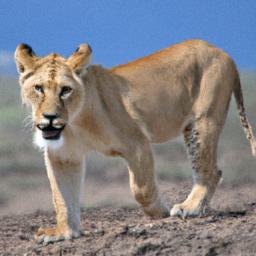}
  \includegraphics[width=0.135\linewidth]{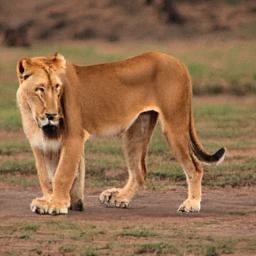}
  \includegraphics[width=0.135\linewidth]{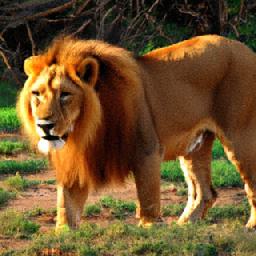}
\tabularnewline
    \raisebox{0.1in}{\rotatebox{90}{\small \emph{$0.5$}
 }}
  \includegraphics[width=0.135\linewidth]{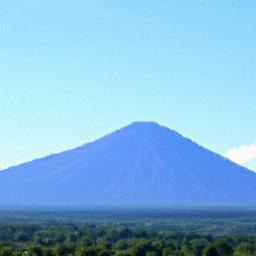}
  \includegraphics[width=0.135\linewidth]{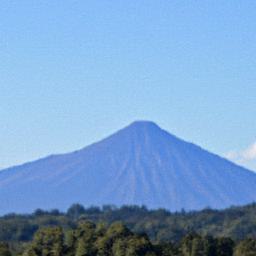}
  \includegraphics[width=0.135\linewidth]{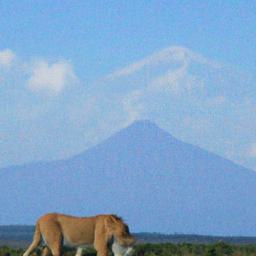}
  \includegraphics[width=0.135\linewidth]{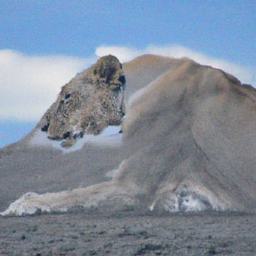}
  \includegraphics[width=0.135\linewidth]{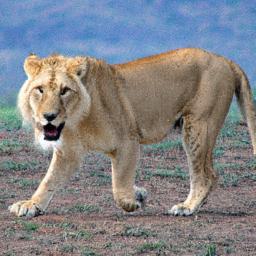}
  \includegraphics[width=0.135\linewidth]{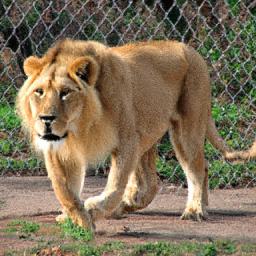}
  \includegraphics[width=0.135\linewidth]{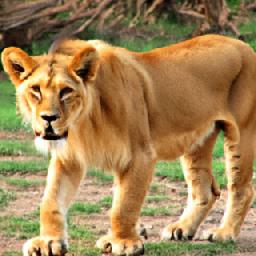}
\tabularnewline
    \raisebox{0.1in}{\rotatebox{90}{\small \emph{$0.6$}
 }}
  \includegraphics[width=0.135\linewidth]{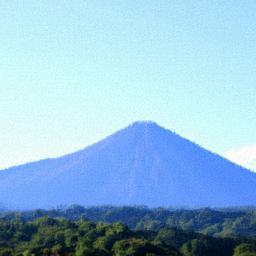}
  \includegraphics[width=0.135\linewidth]{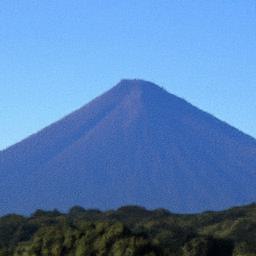}
  \includegraphics[width=0.135\linewidth]{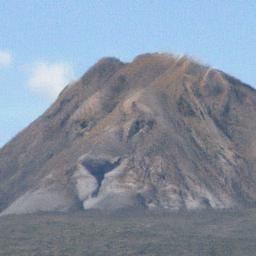}
  \includegraphics[width=0.135\linewidth]{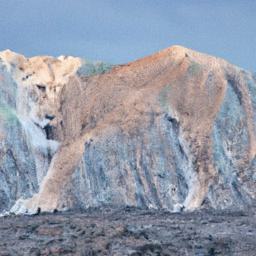}
  \includegraphics[width=0.135\linewidth]{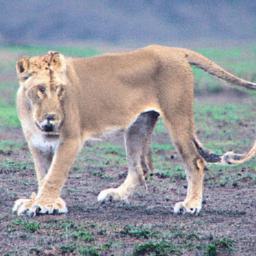}
  \includegraphics[width=0.135\linewidth]{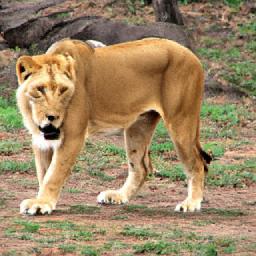}
  \includegraphics[width=0.135\linewidth]{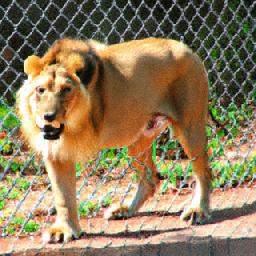}
\tabularnewline
    \raisebox{0.1in}{\rotatebox{90}{\small \emph{$0.8$}
 }}
  \includegraphics[width=0.135\linewidth]{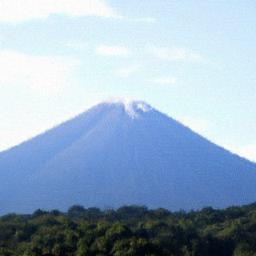}
  \includegraphics[width=0.135\linewidth]{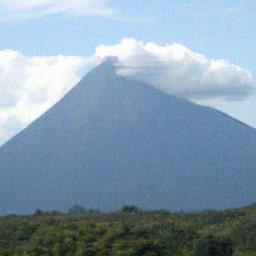}
   \includegraphics[width=0.135\linewidth]{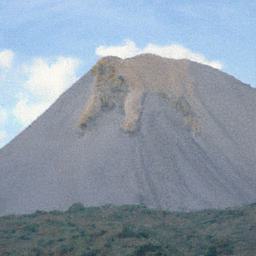}
  \includegraphics[width=0.135\linewidth]{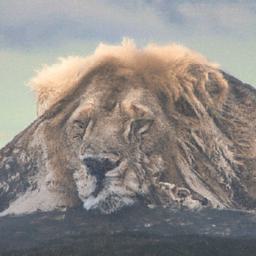}
  \includegraphics[width=0.135\linewidth]{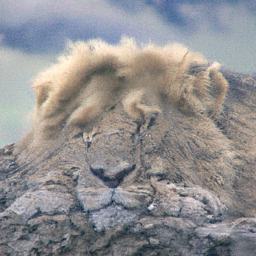}
  \includegraphics[width=0.135\linewidth]{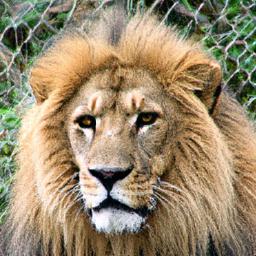}
  \includegraphics[width=0.135\linewidth]{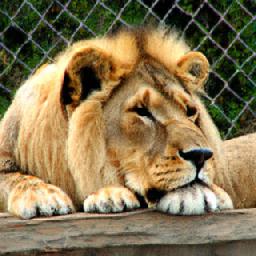}
\tabularnewline
    \raisebox{0.1in}{\rotatebox{90}{\small \emph{$1.0$}
 }}
  \includegraphics[width=0.135\linewidth]{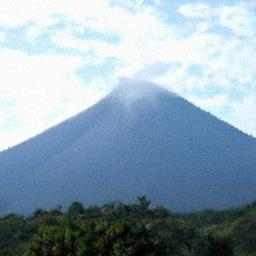}
  \includegraphics[width=0.135\linewidth]{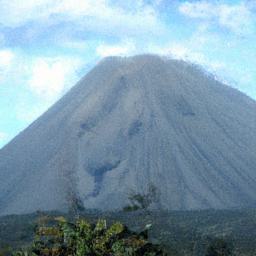}
   \includegraphics[width=0.135\linewidth]{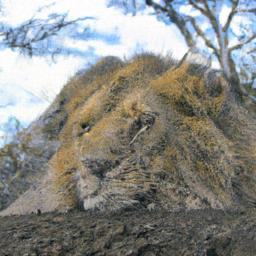}
  \includegraphics[width=0.135\linewidth]{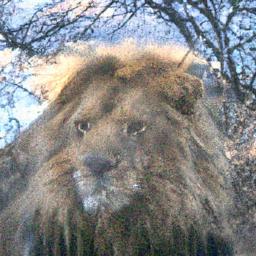}
  \includegraphics[width=0.135\linewidth]{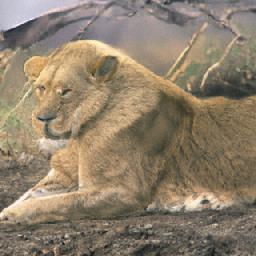}
  \includegraphics[width=0.135\linewidth]{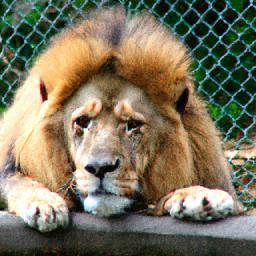}
  \includegraphics[width=0.135\linewidth]{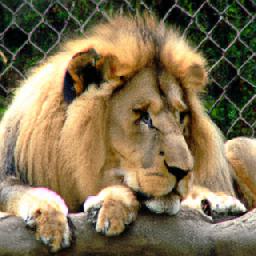}
    \tabularnewline
        \raisebox{0.1in}{\rotatebox{90}{
 }}
 \hspace{0.5mm}
   $\scriptsize{Text}\xrightarrow[\text{weighing factor}]{\hspace*{3cm}}\scriptsize{Class}$
  \tabularnewline
\vspace{2mm}
\vspace{-2\baselineskip}
\end{tabular}}
\vspace{-0.8cm}
\hspace{20pt}\captionof{figure}{\textbf{The role of reliability factor for multimodal generation.} Here, the vertical row going down denotes increasing reliability to the text model. Text is "A mountain". ImageNet class is "Lion"}
\label{fig:glideablation1}
\vspace{-2mm}
\end{figure}%

\begin{table}[!t]
\begin{center}

\vspace{-2mm}
\scalebox{0.7}{
\begin{tabular}{|c |c | c c c c c|}
\toprule[0.15em]
   \textbf{Type}&\textbf{Method} & FID$\textcolor{black}{\downarrow}$ & LPIPS$\textcolor{black}{\downarrow}$& SSIM$\textcolor{black}{\uparrow}$ & MIoU$\textcolor{black}{\uparrow}$  & F1~$\textcolor{black}{\uparrow}$      \\
\midrule[0.15em]
\multirow{5}{*}{Fine-Tuning} &SPADE & 91.77& 0.624 & 0.340 & 0.649&0.728\\
&OASIS& 99.94 &0.634 & 0.320 & 0.625&0.704 \\
&PIX2PIXHD& 223.85& 0.670 & 0.265 & 0.666&0.775  \\
&INADE & 134.11 & 0.656& 0.274 & 0.869&0.922 \\
\midrule[0.15em]
\multirow{5}{*}{Combined} &SPADE & 75.73& \textbf{0.545} & 0.373 & 0.876&0.921\\
&OASIS  &75.57 &0.593 & 0.331 & 0.786&0.863 \\
&PIX2PIXHD& 138.30& 0.560 & 0.363 & 0.840&0.898  \\
&INADE  &47.40& 0.574 & 0.334  & 0.862&0.910\\
\midrule[0.15em]
&TediGAN\cite{xia2021tedigan} & 125.27& 0.545 & 0.409&0.813 & 0.887\\
&OURS &\textbf{33.60}& 0.542 & \textbf{0.421}&\textbf{0.919} & \textbf{0.950} \\

\bottomrule[0.1em]
\end{tabular}}
\caption{Quantitative results for multimodal semantic labels to face generation on FFHQ dataset
}
\label{table:ffhq_semantics}
\end{center}
\vspace{-3em}
\end{table}

\subsection{Analysis and Discussion}

\noindent\textbf{Semantic label to face generation.} The quantitative results for semantic face generation on the FFHQ and CelebA datasets can be found in Tables \ref{table:ffhq_semantics} and \ref{table:celeba_semantics}, respectively. 
As the choice of comparison methods, we use the current state-of-the-art method for semantic face generation TediGAN \cite{xia2021tedigan} and several recently introduced semantic to face generation methods \cite{park2019SPADE,wang2018pix2pixHD,oasis,tan2021diverse}.
The fine-tuning-based multimodal generation and the combined unpaired  
training-based results are shown separately.  TediGAN \cite{xia2021tedigan} has been trained with paired data across all modalities since it supports such a provision. From Tables~ \ref{table:ffhq_semantics}, and \ref{table:celeba_semantics} we can see that all the methods fail to produce reasonable results when trained in the finetuning-based strategies. This can be seen in the high FID scores, low SSIM, and parsed mask accuracy metrics. The alternating training strategy produces reasonably good results, and improve all the evaluation metrics improve. But training this way introduces dataset-specific bias because of which the quality of the existing semantic-to-face generation techniques deteriorates when used on an independent test set during testing. 

   \begin{figure}[tb!]
    \centering
    \setlength{\tabcolsep}{0.5pt}
    {\small
    \renewcommand{\arraystretch}{0.5} 
    \begin{tabular}{c c c c c c c c c c  }
    \captionsetup{type=figure, font=scriptsize}
    \raisebox{0.1in}{\rotatebox{90}{\small \emph{Text
} }}
  \includegraphics[width=0.135\linewidth]{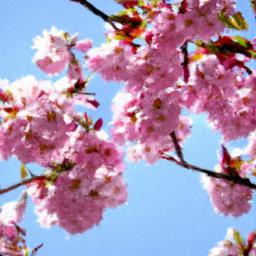}
\includegraphics[width=0.135\linewidth]{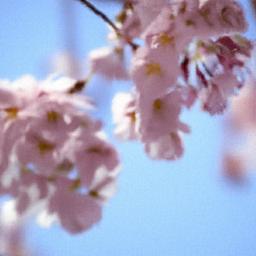}
\includegraphics[width=0.135\linewidth]{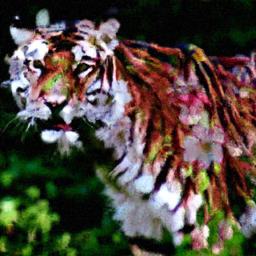}
\includegraphics[width=0.135\linewidth]{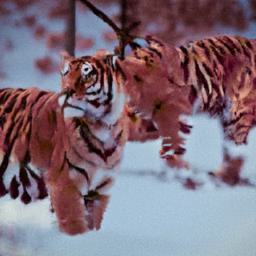}
\includegraphics[width=0.135\linewidth]{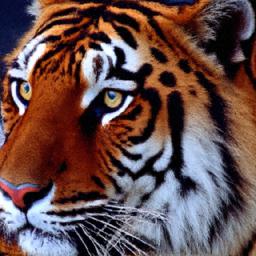}
\includegraphics[width=0.135\linewidth]{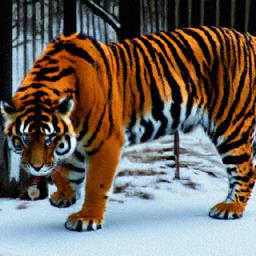}
\includegraphics[width=0.135\linewidth]{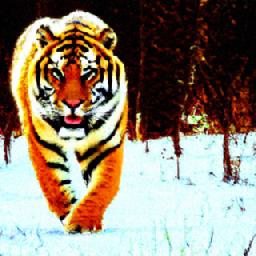}
\tabularnewline
    \raisebox{0.1in}{\rotatebox{90}{\small \emph{Class}
}}
  \includegraphics[width=0.135\linewidth]{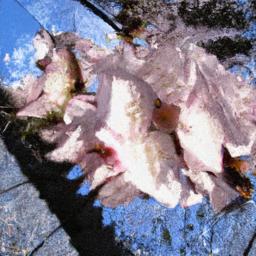}
\includegraphics[width=0.135\linewidth]{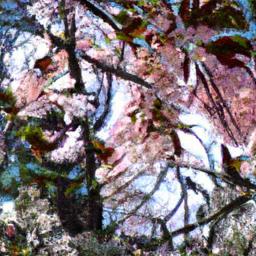}
\includegraphics[width=0.135\linewidth]{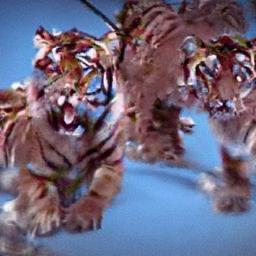}
\includegraphics[width=0.135\linewidth]{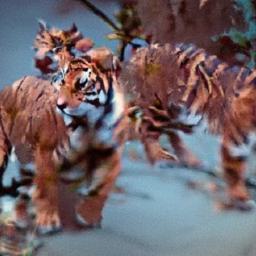}
\includegraphics[width=0.135\linewidth]{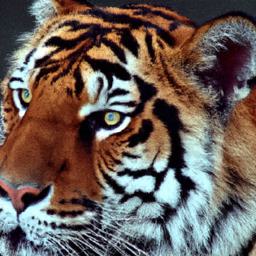}
\includegraphics[width=0.135\linewidth]{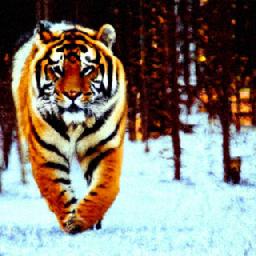}
\includegraphics[width=0.135\linewidth]{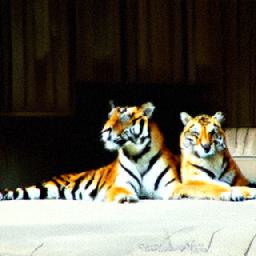}
\tabularnewline
    \raisebox{0.1in}{\rotatebox{90}{\small \emph{Mix
}}}
  \includegraphics[width=0.135\linewidth]{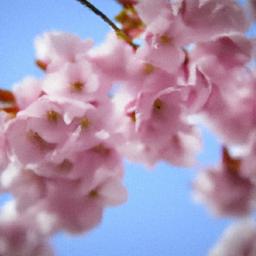}
  \includegraphics[width=0.135\linewidth]{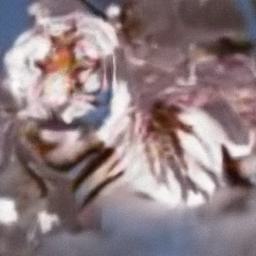}
  \includegraphics[width=0.135\linewidth]{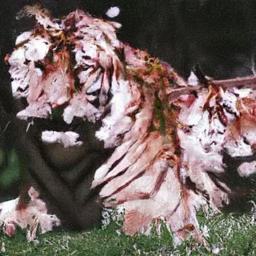}
  \includegraphics[width=0.135\linewidth]{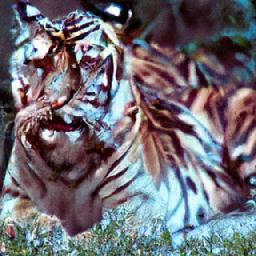}
  \includegraphics[width=0.135\linewidth]{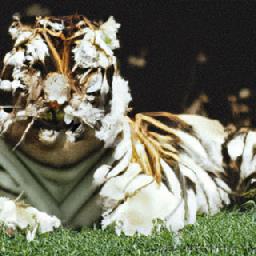}
  \includegraphics[width=0.135\linewidth]{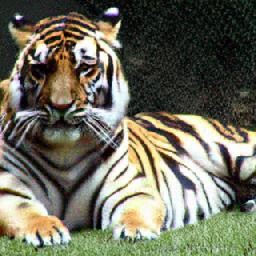}
  \includegraphics[width=0.135\linewidth]{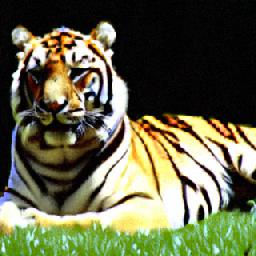}
\tabularnewline
    \raisebox{0.1in}{\rotatebox{90}{\small \emph{Text
} }}
  \includegraphics[width=0.135\linewidth]{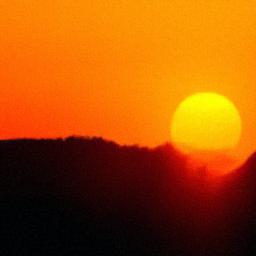}
\includegraphics[width=0.135\linewidth]{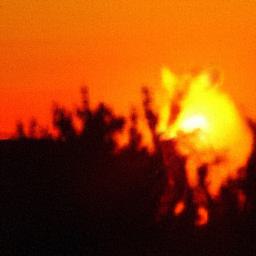}
\includegraphics[width=0.135\linewidth]{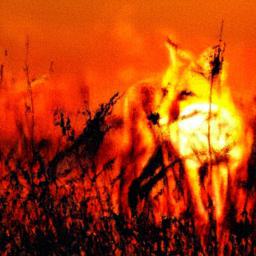}
\includegraphics[width=0.135\linewidth]{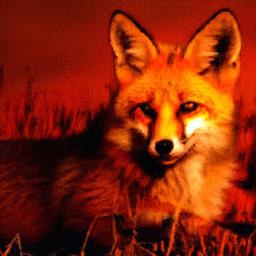}
\includegraphics[width=0.135\linewidth]{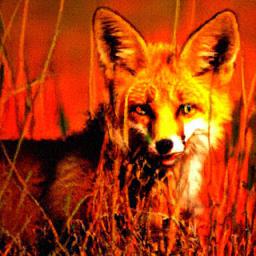}
\includegraphics[width=0.135\linewidth]{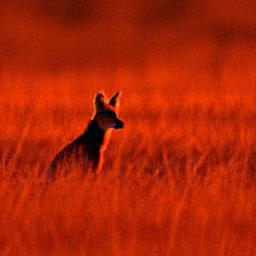}
\includegraphics[width=0.135\linewidth]{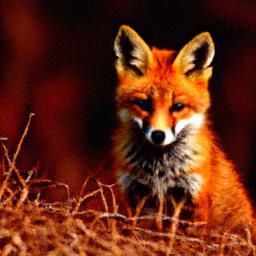}
\tabularnewline
    \raisebox{0.1in}{\rotatebox{90}{\small \emph{Class}
 }}
 \includegraphics[width=0.135\linewidth]{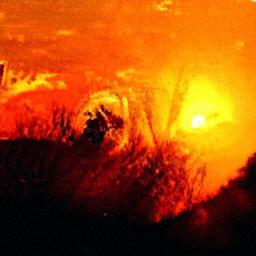}
\includegraphics[width=0.135\linewidth]{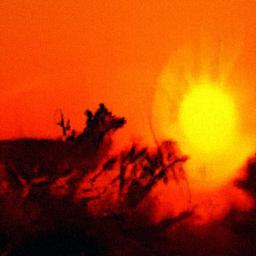}
\includegraphics[width=0.135\linewidth]{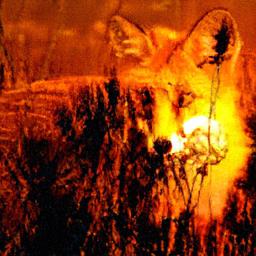}
\includegraphics[width=0.135\linewidth]{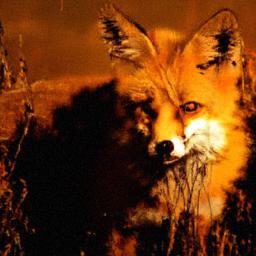}
\includegraphics[width=0.135\linewidth]{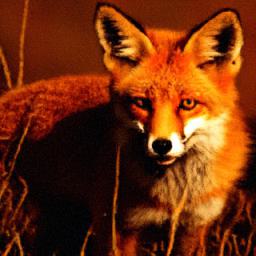}
\includegraphics[width=0.135\linewidth]{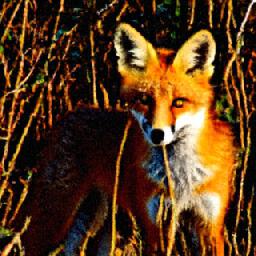}
\includegraphics[width=0.135\linewidth]{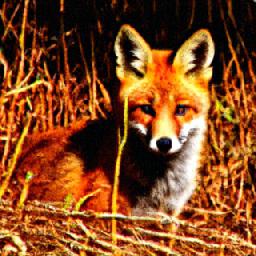}
\tabularnewline
    \raisebox{0.1in}{\rotatebox{90}{\small \emph{Mix
} }}
  \includegraphics[width=0.135\linewidth]{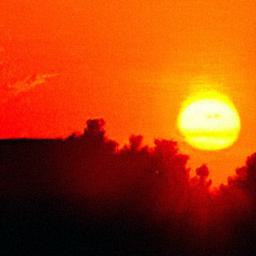}
  \includegraphics[width=0.135\linewidth]{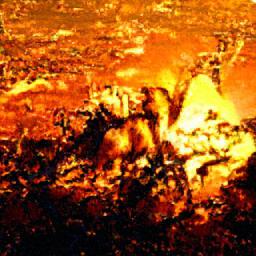}
  \includegraphics[width=0.135\linewidth]{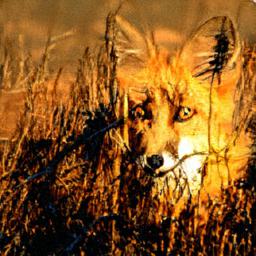}
  \includegraphics[width=0.135\linewidth]{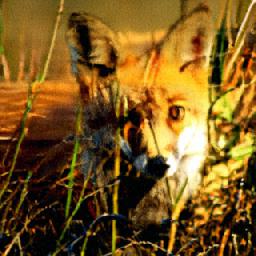}
  \includegraphics[width=0.135\linewidth]{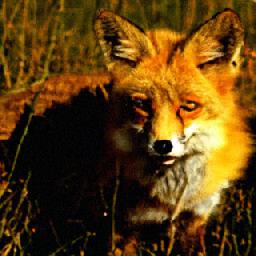}
  \includegraphics[width=0.135\linewidth]{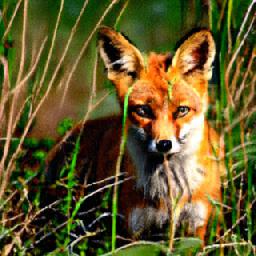}
  \includegraphics[width=0.135\linewidth]{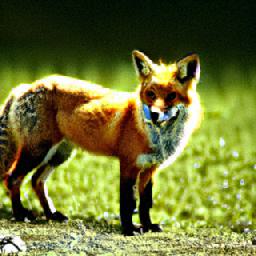}

\tabularnewline

    \tabularnewline
  $\scriptsize{Text}\xrightarrow[\text{weighing factor}]{\hspace*{2cm}}\scriptsize{Class}$
\vspace{2mm}
\vspace{-2\baselineskip}
\end{tabular}}
\vspace{-0.4cm}
\hspace{20pt}\captionof{figure}{\textbf{Interpolation across multiple modalities.} Here, \textit{\{class $\rightarrow$Text\}} denotes to where the unconditional model comes from.}
\label{fig:glideablation2}
\vspace{-2mm}
\end{figure}%

 \begin{table}[t]
\centering

\scalebox{0.8}{
\begin{tabular}{c|c|c c c}
\toprule
Method&Modality&NIQE$(\downarrow)$&Clip$(\uparrow)$&Acc$(\uparrow)$\\
\midrule
GLIDE\cite{nichol2021glide}&\emph{Uni}  &6.64&\textbf{0.317} &0.3  \\
CompGen\cite{liu2022compositional}&\emph{Uni}  &5.71 &0.282 &22.4 \\
Ours&\emph{Multimodal}  &\textbf{5.34}&0.286 &\textbf{87.0}  \\
\bottomrule
\end{tabular}
}
\caption{\textbf{Comparison for generic scenes creation.} Here we consider three different reliability values $\{0,0.5,1\}$ and report the best possible value.Acc denotes ImageNet classification  accuracy}
\label{tab:compgen}
\vspace{-3mm}
\end{table}

 \noindent\textbf{Generic Scenes Generation.} Table \ref{tab:compgen} shows the quantitative comparison for the proposed method. As we can see, the accuracy of the object being in the image is low for GLIDE \cite{nichol2021glide}  because it focuses on regions on the text which are more easier to generate.  One could always make the case that the text model fails because of its weak robustness to doctored text prompts. Hence we perform a comparison with compositional unimodal generation \cite{liu2022compositional}  against a multimodal scenario where we utilize information across datasets. This analysis can be seen in Table \ref{tab:compgen}. we evaluate the performances on 500 generated images using the CLIP score, NIQE score and ImageNet classification accuracy. As can be seen multimodal generation has its advantages that it can introduce new novel classes to the image hence being more accurate and can generate realistic images leading to better metrics.

\noindent\textbf{How to choose a good reliability factor?}
 When we have multiple models trained on different datasets, and conditioning needs to be applied based on both of these models. There exist two scenarios. In the case of independent attributes like a face semantic mask  and hair semantic map, the reliability factors doesn't affect the composite image since each attribute could be added without affecting the others performance. The next possible case is of non independent attributes, where a blend of both images is a possible solution like in \ref{fig:glideablation1}, here the reliability factor depends on how much of each image is desired by the user. For an illustration, we refer the reader to Figure \ref{fig:glideablation1}. Here we utilize the GLIDE model and the ImageNet generation model \cite{dhariwal2021diffusion}. We utilize $100$ steps of deterministic DDIM \cite{song2020denoising} with same random initial noise for all settings.  The vertical values to the left denotes increasing reliability factor for the text model. When reliability factor equals to zero, the top most row, the class model is used as the unconditional model and bottom row shows the case when the text model is used as the unconditional model. Since the text model has more generation capability, we can see that it creates much better naturalistic series of images compared to the ImageNet generation model \cite{dhariwal2021diffusion} which cannot to create artistic scenes.
 \newline
 \noindent\textbf{Why is a reliable score better than a direct solution?} One straightforward solution to the case of multi-modal generation is to use equation \eqref{eq:classifierfree} and  consider the most powerful model as the unconditional model. But this scenario is  a subset of our reliable mean solution. As one can imagine, this formulation puts a strict bias towards the space of data points of one of the models over the other. 
 Whereas using a reliable mean, this specific bias can be negated  and more user defined control is possible. For example, as seen in Fig \ref{fig:glideablation1}, one can obtain a user defined mix of how much each modality should be mixed using the reiable mean. Moreover, In Fig  \ref{fig:glideablation2},
 row 2, we can see that distorted cherry blossoms are created when the ImageNet class conditional model is treated as the stronger model. The text model used is a very powerful model\cite{nichol2021glide}. Hence it is able to model class specific points accurately and a mix of the unconditional densities of both the models can perform valid interpoaltions between both modalities.
 

\section{Ablation studies}
\begin{figure}[t!]
    \centering
    \begin{subfigure}[t]{0.250\linewidth}
      \captionsetup{justification=centering, labelformat=empty, font=scriptsize}
      \includegraphics[width=1\linewidth]{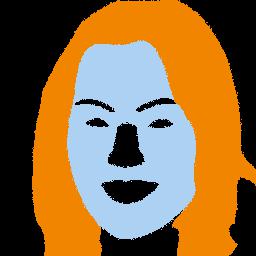}
      \includegraphics[width=1\linewidth]{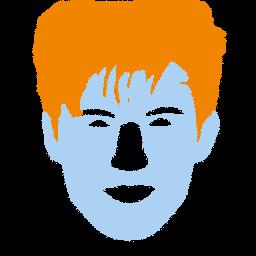}
      \caption{Label}
    \end{subfigure}
    \begin{subfigure}[t]{0.250\linewidth}
      \captionsetup{justification=centering, labelformat=empty, font=scriptsize}
      \includegraphics[width=1\linewidth]{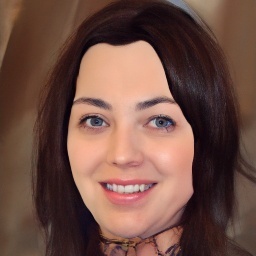}
      \includegraphics[width=1\linewidth]{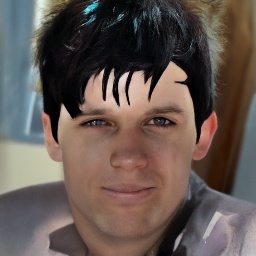}
      \caption{Combined}
    \end{subfigure}
    \begin{subfigure}[t]{0.250\linewidth}
      \captionsetup{justification=centering, labelformat=empty, font=scriptsize}
      \includegraphics[width=1\linewidth]{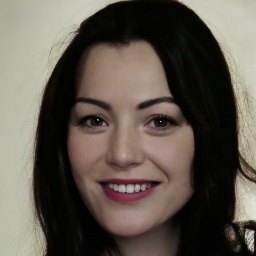}
      \includegraphics[width=1\linewidth]{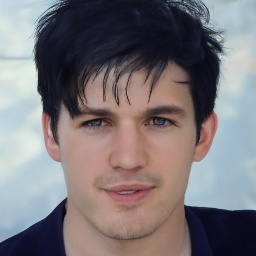}
      \caption{OURS}
    \end{subfigure}
    \vspace{-3mm}    \caption{A visualization of generated images with combined conditional generation (uni-modal) with ours (multi-modal).  }
    \label{fig:ablation}
    \vspace{-2mm}
  \end{figure}

\begin{table}[t!]
\centering
\scalebox{0.8}{
\begin{tabular}{c|c|cc|c c}
\toprule
Semantics&Modality&FID&SSIM&mIoU&F1\\
\midrule
Face&\emph{Uni} &27.66 &0.373&0.842 &0.907  \\
Face+Hair&\emph{Uni}&61.62 & 0.376 &0.872 &0.922  \\
Ours&\emph{Multi}  &26.09 &0.416 &0.911 &0.948 \\
\bottomrule
\end{tabular}
}
\caption{Analyzing advantage of the proposed sampling method over normal training and regular sampling techniques.}
\label{tab:abl_gen}
\vspace{-8mm}
\end{table}

To show the effectiveness of the proposed multimodal strategy over a normal method trained unimodally, we retrain a diffusion model that can take one or more conditioning simultaneously. We analyze the FID score and the SSIM scores on the images and their corresponding ground truths,  and the F1 score and mIoU score between the original parsed maps vs the reconstructed parsed maps. We perform ablation over three different scenarios: with using one modality when both conditioning modality is given at a time and our proposed sampling strategy. As we can see from Table \ref{tab:abl_gen}, there is a significant boost to the performance using the specified sampling strategy. This shows that the multimodal sampling strategy enforces stronger conditioning and produces better-quality results. Figure \ref{fig:ablation} shows the results corresponding to the combined conditioning strategy and our proposed method. As can be seen from this figure\ref{fig:ablation}, when regular diffusion-based inference time sampling is performed from a model trained with different modalities across different datasets, the sampling procedure generates unrealistic results. In contrast, we are able to generate much more realistic results.

\section{Limitations and Future scope}

One limitation of our method is that the dimension of the latent space modelled by the diffusion models is required to be the same. If the number of channels differ for the latent variable $z_t$, then this method cannot be utilized. This is the reason why we could not utilize stable diffusion model \cite{rombach2022high} for our experiments. Another scenario when our model can fail is when one model is asked to create a sample that could be easily generated and the other model is asked for a harder class. Thirdly, if contradictory information is given as input across modalities, our method fails to produce the desired output. More visual results corresponding to this condition are given in the supplementary document. This problem can be easily alleviated by using different reliability weights to the different conditioning modalities and giving more weight to the most desired conditioning modality as can be seen from Fig.  \ref{fig:glideablation1}.

\section{Conclusion}

In this paper, we propose one of the first methods that can perform multimodal generation using individual models trained for multiple sub-tasks. The multimodal generation is enabled by a newly proposed formulation utilizing a generalized product of experts. We introduce a new reliability parameter that allows user-defined control while performing multimodal mixing of correlated modalities. We briefly discuss the design choices of the reliability parameter for different applications. The proposed sampling significantly boosts the performance of multimodal modal generation using diffusion models  compared to the sampling using a unimodal network. We show results on various multimodal tasks with trained as well as publically available off-the-shelf models to show the effectiveness of our method.
{\small
\bibliographystyle{ieee_fullname}
\bibliography{paper}

\begin{thebibliography}{10}\itemsep=-1pt

\bibitem{avrahami2022blended}
Omri Avrahami, Dani Lischinski, and Ohad Fried.
\newblock Blended diffusion for text-driven editing of natural images.
\newblock In {\em Proceedings of the IEEE/CVF Conference on Computer Vision and
  Pattern Recognition}, pages 18208--18218, 2022.

\bibitem{cao2014generalized}
Yanshuai Cao and David~J Fleet.
\newblock Generalized product of experts for automatic and principled fusion of
  gaussian process predictions.
\newblock {\em arXiv preprint arXiv:1410.7827}, 2014.

\bibitem{chen2020dmgan}
Zhangling Chen, Ce Wang, Huaming Wu, Kun Shang, and Jun Wang.
\newblock Dmgan: Discriminative metric-based generative adversarial networks.
\newblock {\em Knowledge-Based Systems}, 192:105370, 2020.

\bibitem{choi2021ilvr}
Jooyoung Choi, Sungwon Kim, Yonghyun Jeong, Youngjune Gwon, and Sungroh Yoon.
\newblock Ilvr: Conditioning method for denoising diffusion probabilistic
  models.
\newblock {\em arXiv preprint arXiv:2108.02938}, 2021.

\bibitem{dhariwal2021diffusion}
Prafulla Dhariwal and Alexander Nichol.
\newblock Diffusion models beat gans on image synthesis.
\newblock {\em Advances in Neural Information Processing Systems}, 34, 2021.

\bibitem{ding2001canny}
Lijun Ding and Ardeshir Goshtasby.
\newblock On the canny edge detector.
\newblock {\em Pattern recognition}, 34(3):721--725, 2001.

\bibitem{goodfellow2020generative}
Ian Goodfellow, Jean Pouget-Abadie, Mehdi Mirza, Bing Xu, David Warde-Farley,
  Sherjil Ozair, Aaron Courville, and Yoshua Bengio.
\newblock Generative adversarial networks.
\newblock {\em Communications of the ACM}, 63(11):139--144, 2020.

\bibitem{heusel2017gans}
Martin Heusel, Hubert Ramsauer, Thomas Unterthiner, Bernhard Nessler, and Sepp
  Hochreiter.
\newblock Gans trained by a two time-scale update rule converge to a local nash
  equilibrium.
\newblock {\em Advances in neural information processing systems}, 30, 2017.

\bibitem{ho2020denoising}
Jonathan Ho, Ajay Jain, and Pieter Abbeel.
\newblock Denoising diffusion probabilistic models.
\newblock {\em Advances in Neural Information Processing Systems},
  33:6840--6851, 2020.

\bibitem{ho2021classifier}
Jonathan Ho and Tim Salimans.
\newblock Classifier-free diffusion guidance.
\newblock In {\em NeurIPS 2021 Workshop on Deep Generative Models and
  Downstream Applications}, 2021.

\bibitem{huang2021multimodal}
Xun Huang, Arun Mallya, Ting-Chun Wang, and Ming-Yu Liu.
\newblock Multimodal conditional image synthesis with product-of-experts gans.
\newblock {\em arXiv preprint arXiv:2112.05130}, 2021.

\bibitem{huang2022multimodal}
Xun Huang, Arun Mallya, Ting-Chun Wang, and Ming-Yu Liu.
\newblock Multimodal conditional image synthesis with product-of-experts gans.
\newblock In {\em European Conference on Computer Vision}, pages 91--109.
  Springer, 2022.

\bibitem{isola2017image}
Phillip Isola, Jun-Yan Zhu, Tinghui Zhou, and Alexei~A Efros.
\newblock Image-to-image translation with conditional adversarial networks.
\newblock In {\em Proceedings of the IEEE conference on computer vision and
  pattern recognition}, pages 1125--1134, 2017.

\bibitem{karras2017progressive}
Tero Karras, Timo Aila, Samuli Laine, and Jaakko Lehtinen.
\newblock Progressive growing of gans for improved quality, stability, and
  variation.
\newblock {\em arXiv preprint arXiv:1710.10196}, 2017.

\bibitem{karras2019style}
Tero Karras, Samuli Laine, and Timo Aila.
\newblock A style-based generator architecture for generative adversarial
  networks.
\newblock In {\em Proceedings of the IEEE/CVF conference on computer vision and
  pattern recognition}, pages 4401--4410, 2019.

\bibitem{kawar2022denoising}
Bahjat Kawar, Michael Elad, Stefano Ermon, and Jiaming Song.
\newblock Denoising diffusion restoration models.
\newblock {\em arXiv preprint arXiv:2201.11793}, 2022.

\bibitem{kawar2022imagic}
Bahjat Kawar, Shiran Zada, Oran Lang, Omer Tov, Huiwen Chang, Tali Dekel, Inbar
  Mosseri, and Michal Irani.
\newblock Imagic: Text-based real image editing with diffusion models.
\newblock {\em arXiv preprint arXiv:2210.09276}, 2022.

\bibitem{kingma2013auto}
Diederik~P Kingma and Max Welling.
\newblock Auto-encoding variational bayes.
\newblock {\em arXiv preprint arXiv:1312.6114}, 2013.

\bibitem{liu2022compositional}
Nan Liu, Shuang Li, Yilun Du, Antonio Torralba, and Joshua~B Tenenbaum.
\newblock Compositional visual generation with composable diffusion models.
\newblock {\em arXiv preprint arXiv:2206.01714}, 2022.

\bibitem{lugmayr2022repaint}
Andreas Lugmayr, Martin Danelljan, Andres Romero, Fisher Yu, Radu Timofte, and
  Luc Van~Gool.
\newblock Repaint: Inpainting using denoising diffusion probabilistic models.
\newblock In {\em Proceedings of the IEEE/CVF Conference on Computer Vision and
  Pattern Recognition}, pages 11461--11471, 2022.

\bibitem{nichol2021glide}
Alex Nichol, Prafulla Dhariwal, Aditya Ramesh, Pranav Shyam, Pamela Mishkin,
  Bob McGrew, Ilya Sutskever, and Mark Chen.
\newblock Glide: Towards photorealistic image generation and editing with
  text-guided diffusion models.
\newblock {\em arXiv preprint arXiv:2112.10741}, 2021.

\bibitem{park2019SPADE}
Taesung Park, Ming-Yu Liu, Ting-Chun Wang, and Jun-Yan Zhu.
\newblock Semantic image synthesis with spatially-adaptive normalization.
\newblock In {\em Proceedings of the IEEE Conference on Computer Vision and
  Pattern Recognition}, 2019.

\bibitem{preechakul2022diffusion}
Konpat Preechakul, Nattanat Chatthee, Suttisak Wizadwongsa, and Supasorn
  Suwajanakorn.
\newblock Diffusion autoencoders: Toward a meaningful and decodable
  representation.
\newblock In {\em Proceedings of the IEEE/CVF Conference on Computer Vision and
  Pattern Recognition}, pages 10619--10629, 2022.

\bibitem{radford2021learning}
Alec Radford, Jong~Wook Kim, Chris Hallacy, Aditya Ramesh, Gabriel Goh,
  Sandhini Agarwal, Girish Sastry, Amanda Askell, Pamela Mishkin, Jack Clark,
  et~al.
\newblock Learning transferable visual models from natural language
  supervision.
\newblock In {\em International Conference on Machine Learning}, pages
  8748--8763. PMLR, 2021.

\bibitem{radford2015unsupervised}
Alec Radford, Luke Metz, and Soumith Chintala.
\newblock Unsupervised representation learning with deep convolutional
  generative adversarial networks.
\newblock {\em arXiv preprint arXiv:1511.06434}, 2015.

\bibitem{ramesh2022hierarchical}
Aditya Ramesh, Prafulla Dhariwal, Alex Nichol, Casey Chu, and Mark Chen.
\newblock Hierarchical text-conditional image generation with clip latents.
\newblock {\em arXiv preprint arXiv:2204.06125}, 2022.

\bibitem{ramesh2021zero}
Aditya Ramesh, Mikhail Pavlov, Gabriel Goh, Scott Gray, Chelsea Voss, Alec
  Radford, Mark Chen, and Ilya Sutskever.
\newblock Zero-shot text-to-image generation.
\newblock In {\em International Conference on Machine Learning}, pages
  8821--8831. PMLR, 2021.

\bibitem{rombach2022high}
Robin Rombach, Andreas Blattmann, Dominik Lorenz, Patrick Esser, and Bj{\"o}rn
  Ommer.
\newblock High-resolution image synthesis with latent diffusion models.
\newblock In {\em Proceedings of the IEEE/CVF Conference on Computer Vision and
  Pattern Recognition}, pages 10684--10695, 2022.

\bibitem{saharia2022photorealistic}
Chitwan Saharia, William Chan, Saurabh Saxena, Lala Li, Jay Whang, Emily
  Denton, Seyed Kamyar~Seyed Ghasemipour, Burcu~Karagol Ayan, S~Sara Mahdavi,
  Rapha~Gontijo Lopes, et~al.
\newblock Photorealistic text-to-image diffusion models with deep language
  understanding.
\newblock {\em arXiv preprint arXiv:2205.11487}, 2022.

\bibitem{saharia2021image}
Chitwan Saharia, Jonathan Ho, William Chan, Tim Salimans, David~J Fleet, and
  Mohammad Norouzi.
\newblock Image super-resolution via iterative refinement.
\newblock {\em arXiv preprint arXiv:2104.07636}, 2021.

\bibitem{oasis}
Edgar Sch{\"o}nfeld, Vadim Sushko, Dan Zhang, Juergen Gall, Bernt Schiele, and
  Anna Khoreva.
\newblock You only need adversarial supervision for semantic image synthesis.
\newblock In {\em International Conference on Learning Representations}, 2021.

\bibitem{shi2019variational}
Yuge Shi, Brooks Paige, Philip Torr, et~al.
\newblock Variational mixture-of-experts autoencoders for multi-modal deep
  generative models.
\newblock {\em Advances in Neural Information Processing Systems}, 32, 2019.

\bibitem{sohl2015deep}
Jascha Sohl-Dickstein, Eric Weiss, Niru Maheswaranathan, and Surya Ganguli.
\newblock Deep unsupervised learning using nonequilibrium thermodynamics.
\newblock In {\em International Conference on Machine Learning}, pages
  2256--2265. PMLR, 2015.

\bibitem{sohn2015learning}
Kihyuk Sohn, Honglak Lee, and Xinchen Yan.
\newblock Learning structured output representation using deep conditional
  generative models.
\newblock {\em Advances in neural information processing systems}, 28, 2015.

\bibitem{song2020denoising}
Jiaming Song, Chenlin Meng, and Stefano Ermon.
\newblock Denoising diffusion implicit models.
\newblock {\em arXiv preprint arXiv:2010.02502}, 2020.

\bibitem{song2019generative}
Yang Song and Stefano Ermon.
\newblock Generative modeling by estimating gradients of the data distribution.
\newblock {\em Advances in Neural Information Processing Systems}, 32, 2019.

\bibitem{sutter2021generalized}
Thomas~M Sutter, Imant Daunhawer, and Julia~E Vogt.
\newblock Generalized multimodal elbo.
\newblock {\em arXiv preprint arXiv:2105.02470}, 2021.

\bibitem{suzuki2016joint}
Masahiro Suzuki, Kotaro Nakayama, and Yutaka Matsuo.
\newblock Joint multimodal learning with deep generative models.
\newblock {\em arXiv preprint arXiv:1611.01891}, 2016.

\bibitem{tan2019efficientnet}
Mingxing Tan and Quoc Le.
\newblock Efficientnet: Rethinking model scaling for convolutional neural
  networks.
\newblock In {\em International conference on machine learning}, pages
  6105--6114. PMLR, 2019.

\bibitem{tan2021diverse}
Zhentao Tan, Menglei Chai, Dongdong Chen, Jing Liao, Qi Chu, Bin Liu, Gang Hua,
  and Nenghai Yu.
\newblock Diverse semantic image synthesis via probability distribution
  modeling.
\newblock In {\em Proceedings of the IEEE/CVF Conference on Computer Vision and
  Pattern Recognition}, pages 7962--7971, 2021.

\bibitem{tang2021attentiongan}
Hao Tang, Hong Liu, Dan Xu, Philip~HS Torr, and Nicu Sebe.
\newblock Attentiongan: Unpaired image-to-image translation using
  attention-guided generative adversarial networks.
\newblock {\em IEEE Transactions on Neural Networks and Learning Systems},
  2021.

\bibitem{wang2018pix2pixHD}
Ting-Chun Wang, Ming-Yu Liu, Jun-Yan Zhu, Andrew Tao, Jan Kautz, and Bryan
  Catanzaro.
\newblock High-resolution image synthesis and semantic manipulation with
  conditional gans.
\newblock In {\em Proceedings of the IEEE Conference on Computer Vision and
  Pattern Recognition}, 2018.

\bibitem{welling2011bayesian}
Max Welling and Yee~W Teh.
\newblock Bayesian learning via stochastic gradient langevin dynamics.
\newblock In {\em Proceedings of the 28th international conference on machine
  learning (ICML-11)}, pages 681--688, 2011.

\bibitem{wu2018multimodal}
Mike Wu and Noah Goodman.
\newblock Multimodal generative models for scalable weakly-supervised learning.
\newblock {\em Advances in Neural Information Processing Systems}, 31, 2018.

\bibitem{xia2021tedigan}
Weihao Xia, Yujiu Yang, Jing-Hao Xue, and Baoyuan Wu.
\newblock Tedigan: Text-guided diverse face image generation and manipulation.
\newblock In {\em Proceedings of the IEEE/CVF Conference on Computer Vision and
  Pattern Recognition}, pages 2256--2265, 2021.

\bibitem{zeng2021np}
Xiaohui Zeng, Raquel Urtasun, Richard Zemel, Sanja Fidler, and Renjie Liao.
\newblock Np-draw: A non-parametric structured latent variable model for image
  generation.
\newblock {\em arXiv preprint arXiv:2106.13435}, 2021.

\bibitem{zhang2018perceptual}
Richard Zhang, Phillip Isola, Alexei~A Efros, Eli Shechtman, and Oliver Wang.
\newblock The unreasonable effectiveness of deep features as a perceptual
  metric.
\newblock In {\em CVPR}, 2018.

\bibitem{zhang2021m6}
Zhu Zhang, Jianxin Ma, Chang Zhou, Rui Men, Zhikang Li, Ming Ding, Jie Tang,
  Jingren Zhou, and Hongxia Yang.
\newblock M6-ufc: Unifying multi-modal controls for conditional image
  synthesis.
\newblock {\em arXiv preprint arXiv:2105.14211}, 2021.

\bibitem{zheng2021farl}
Yinglin Zheng, Hao Yang, Ting Zhang, Jianmin Bao, Dongdong Chen, Yangyu Huang,
  Lu Yuan, Dong Chen, Ming Zeng, and Fang Wen.
\newblock General facial representation learning in a visual-linguistic manner.
\newblock {\em arXiv preprint arXiv:2112.03109}, 2021.

\end{thebibliography}
}
\newpage

\subsection{Generalized product of experts for differential weighing of different modalities}

As mentioned in section the conditional density in the presence of multiple conditioning strategies turns out to be
\begin{multline}
    P(z|{\bf{X}})=\frac{P(z)}{P({\bf{X}})}\prod_{i=1}^{N}P(x_i|z) 
    = KP(z) \frac{\prod_{i=1}^{N}P(z|x_i)}{\prod_{i=1}^{N}P(z)},
    \label{eq:POE5}
\end{multline}

Under Gaussian assumption where the unconditional densities and the conditional densities follow a Gaussian distribution, The above equation could be approximated using Generalized product of experts\cite{cao2014generalized} to
\begin{equation}
    P(z|{\bf{X}})
    = KP(z) \frac{\prod_{i=1}^{N}q(z|x_i)}{\prod_{i=1}^{N}q(z)},
    \label{eq:POE2}
\end{equation}

where $q(.)$ is an estimate of the original density and 
\begin{equation}
    q \sim N(\mu, \sigma)
\end{equation}
Generalized product of experts\cite{cao2014generalized} further allows stringer conditions to be applied to the individual conditional densities and reweigh the individual densities to favour some over other, represented by
\begin{equation}
    P(z|{\bf{X}})
    = KP(z) \frac{\prod_{i=1}^{N}q^{\beta_i}(z|x_i)}{\prod_{i=1}^{N}q(z)},
    \label{eq:POE3}
\end{equation}

Bringing the idea of reliable means in the equation, and assuming that a good estimate of the conditional densities can be made, Eq \ref{eq:POE3} changes to
\begin{equation}
    P(z|{\bf{X}})
    = (\prod_{i=1}^{N}P_{\delta_i}^{a_i}(z_t|\phi))\frac{\prod_{i=1}^{N}P_{\delta_i}^{\beta_i}(z_t|x_i)}{\prod_{i=1}^{N}(\prod_{j=1}^{N}P_{\delta_i}^{a_i}(z_t|\phi))^{\beta_i}},
    \label{eq:POE4}
\end{equation}
Hence the effective score becomes 
 \begin{equation}
\begin{split}
 \nabla_{z_t}\log  (z_t|{\bf{X}}) =   \\
\nabla_{z_t}\log  \biggl((\prod_{i=1}^{N}P_{\delta_i}^{a_i}(z_t|\phi))\frac{\prod_{i=1}^{N}P_{\delta_i}^{\beta_i}(z_t|x_i)}{\prod_{j=1}^{N}\prod_{i=1}^{N}P_{\delta_i}^{a_i\beta_i}(z_t|\phi)} \biggr)=\\ \sum_{i=1}^N  a_i \nabla_{z_t}\text{log}  P_{\delta_j}(z_t|\phi)+\\
     \sum_{i=1}^N \biggl( \beta_i\nabla_{z_t}\text{log} P_{\delta_i}(z_t|x_i) - \beta_i\sum_{j=1}^Na_j \nabla_{z_t}\text{log}  P_{\delta_j}(z_t|\phi)\biggr),
\end{split}
\end{equation}

The equivalent score can be represented by 
\begin{multline}
   \epsilon_{c}= \epsilon_{\theta}(z_t,{\bf{X}} ,t) =\sum_{i=1}^N a_i \epsilon_{i}(z_t,\phi,t) +\\
   \sum_{i=1}^N\beta_i\biggl( \epsilon_{i}(z_t,x_i,t)-
   \sum_{j=i}^N a_{j}\epsilon_j (z_t,\phi,t)\biggr),
   \label{eq:multimodal2}
\end{multline}
\subsection{Extension to cases with different variance schedules}
Regardless of the variance schedules, one key property in diffusion models is that adding noise equivalent to the $T$ timesteps should always converge to a standard normal distribution, i.e
\begin{equation}
    q(x_T|x_{0}) = \mathcal{N}(x_t; \sqrt{\bar{\alpha}_t} x_0, (1-\bar{\alpha}_t) \mathcal{I}).
    \label{eq:q_sample0}
\end{equation}

where $\bar{\alpha}$ is the cumulative product of the variance schedule $\beta$. Once the effective $\epsilon$ at each timestep is obtained the individual scores could be calculated using the equation for score calculation mentioned in the main paper and the image at the next timestep could be obtained using the the respective variance schedule.


\subsection{A new approach for text to image generation for facial images}
In all of our experiments defined in section we include text based description one of the modalities. The existing diffusion based approaches make use of millions of paired image-text pairs and train models for the task of text to Image generation. But in multimodal scenarios because of the availability of information from other modalities, we do not require such a powerful model. Hence we propose an novel lightweight text to image approach as follows. Consider a conditional diffusion model for text to image conversion defined by $g_{\phi}(z_t,\text{emb},t)$. For our model we make use of a image-text pretraining model like CLIP with the corresponding Image and text embedders  defined by $(h_{img}(.),h_{text}(.))$. During training, we condition using the low dimensional Image emdedding using adaptive  group normalization at stage of our network. The Image embedding is obtained by using the image embedder of clip on the incoming training image. The loss for optimization of the parameters $\phi$ of the network are defined by,
\begin{equation}
 L =E_{t \sim[1, T], \epsilon \sim \mathcal{N}(0, \mathbf{I})}\left[\left\|\epsilon-g_{\phi}\left(\mathbf{z}_t,h_{img}(z_0), t\right)\right\|^{2}\right]\\
\end{equation}

During the inference process, we make use of the text embedder of the vision-language model and obtain the text embeddings by passing the corresponding text conditioning $text_{in}$  sample using the process,
\begin{align*}
        z_{t-1} \leftarrow{} \frac{1}{\sqrt{1-\beta_t}} \left( z_t - \frac{ \beta_t}{\sqrt{1-\bar{\alpha}_t}}g_{\phi}(z_t,h_{text}(text_{in}), t) \right) \\ + \sigma_t^2 \boldsymbol{\eta}, \\
    \text{ where } \boldsymbol{\eta} = \mathcal{N}(\boldsymbol{0}, \boldsymbol{I}),
    \label{eq:cond_samp}
\end{align*}

The embeddings for image and text were obtained by using \cite{zheng2021farl}

\subsection{Text to Image generation:-}

For Multimodal text to Image generation, we comparing with existing works trained for text to image generation in CelebA-MM dataset. But unlike the exiting works that were trained with paired text-image prompts, our model perform zero-shot text to image generation and it has never seen any text prompts during training time. We utilize FARL\cite{zheng2021farl} for obtaining the image-text embeddings of the images. For evaluating the quality of text to image generation, we follow \cite{xia2021tedigan}. 

\subsection{Sketch to Face training:-}

For generating rough sketches of the faces, we utilize \cite{ding2001canny} edge detector to extract edges from the image and use this as the conditioning technique while training the network.
\begin{figure*}[tb!]
    \centering
    \setlength{\tabcolsep}{0.5pt}
    {\small
    \renewcommand{\arraystretch}{0.5} 
    \begin{tabular}{c c c c c c c c c c}
    \captionsetup{type=figure, font=scriptsize}
    \raisebox{0.1in}{\rotatebox{90}{\small \emph{$0.0$}
 }}
  \includegraphics[width=0.135\linewidth]{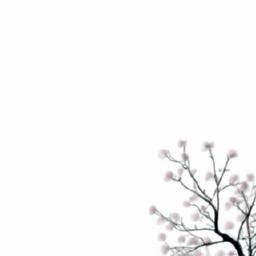}
  \includegraphics[width=0.135\linewidth]{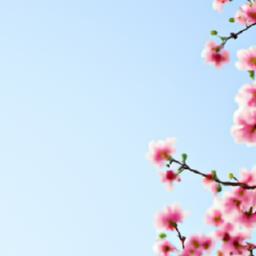}
  \includegraphics[width=0.135\linewidth]{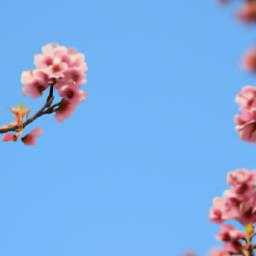}
  \includegraphics[width=0.135\linewidth]{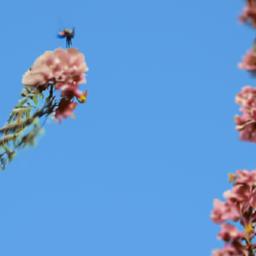}
  \includegraphics[width=0.135\linewidth]{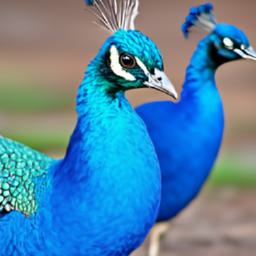}
  \includegraphics[width=0.135\linewidth]{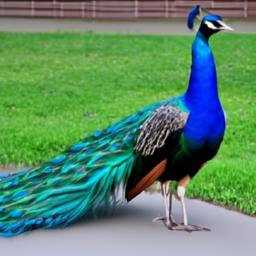}
  \includegraphics[width=0.135\linewidth]{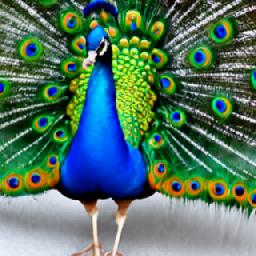}
 \tabularnewline
     \raisebox{0.1in}{\rotatebox{90}{\small \emph{$0.2$}
 }}
  \includegraphics[width=0.135\linewidth]{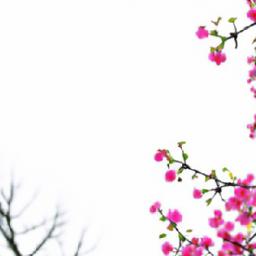}
  \includegraphics[width=0.135\linewidth]{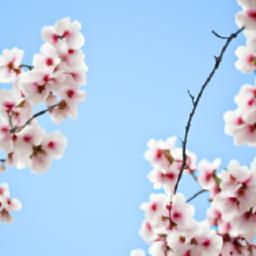}
  \includegraphics[width=0.135\linewidth]{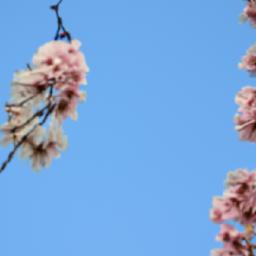}
  \includegraphics[width=0.135\linewidth]{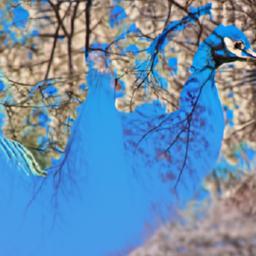}
  \includegraphics[width=0.135\linewidth]{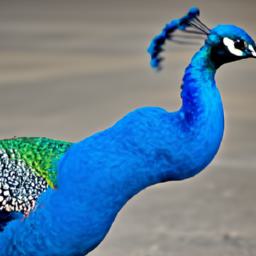}
  \includegraphics[width=0.135\linewidth]{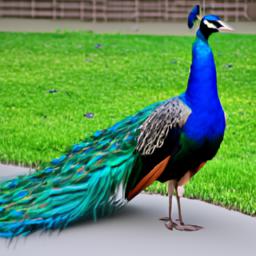}
  \includegraphics[width=0.135\linewidth]{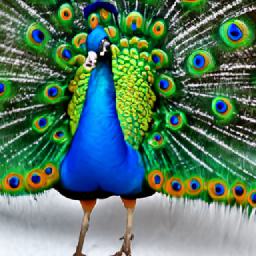}
\tabularnewline
    \raisebox{0.1in}{\rotatebox{90}{\small \emph{$0.4$}
 }}
  \includegraphics[width=0.135\linewidth]{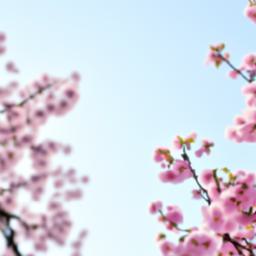}
  \includegraphics[width=0.135\linewidth]{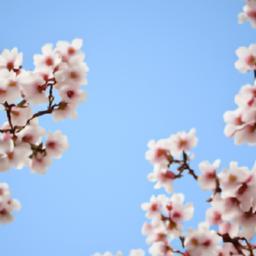}
  \includegraphics[width=0.135\linewidth]{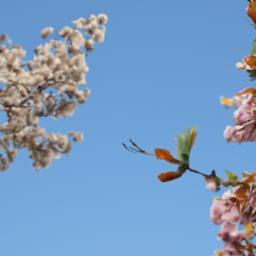}
  \includegraphics[width=0.135\linewidth]{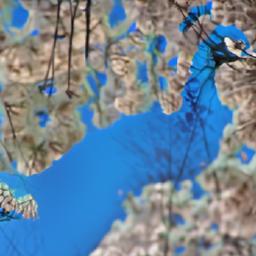}
  \includegraphics[width=0.135\linewidth]{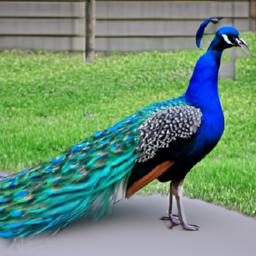}
  \includegraphics[width=0.135\linewidth]{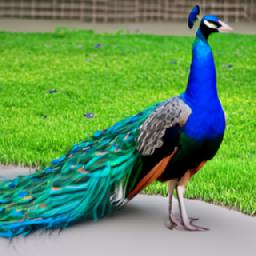}
  \includegraphics[width=0.135\linewidth]{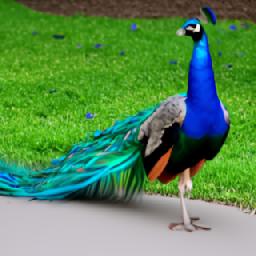}
\tabularnewline
    \raisebox{0.1in}{\rotatebox{90}{\small \emph{$0.5$}
 }}
  \includegraphics[width=0.135\linewidth]{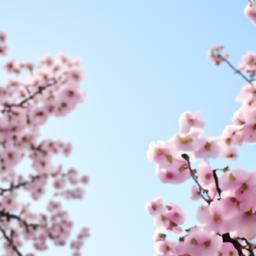}
  \includegraphics[width=0.135\linewidth]{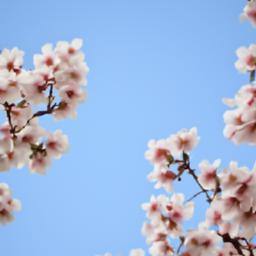}
  \includegraphics[width=0.135\linewidth]{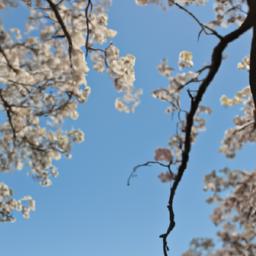}
  \includegraphics[width=0.135\linewidth]{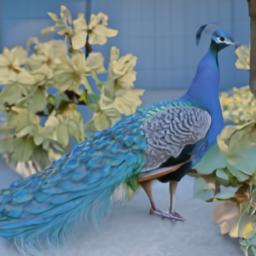}
  \includegraphics[width=0.135\linewidth]{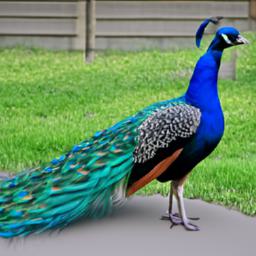}
  \includegraphics[width=0.135\linewidth]{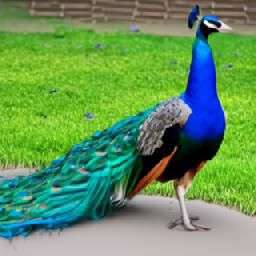}
  \includegraphics[width=0.135\linewidth]{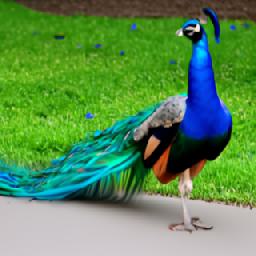}
\tabularnewline
    \raisebox{0.1in}{\rotatebox{90}{\small \emph{$0.6$}
 }}
  \includegraphics[width=0.135\linewidth]{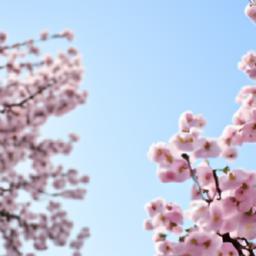}
  \includegraphics[width=0.135\linewidth]{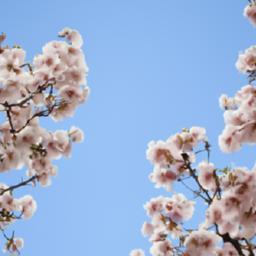}
  \includegraphics[width=0.135\linewidth]{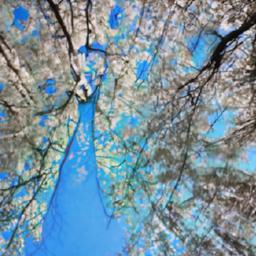}
  \includegraphics[width=0.135\linewidth]{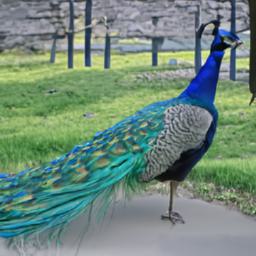}
  \includegraphics[width=0.135\linewidth]{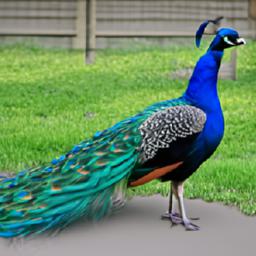}
  \includegraphics[width=0.135\linewidth]{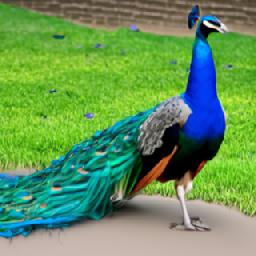}
  \includegraphics[width=0.135\linewidth]{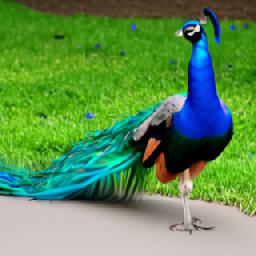}
\tabularnewline
    \raisebox{0.1in}{\rotatebox{90}{\small \emph{$0.8$}
 }}
  \includegraphics[width=0.135\linewidth]{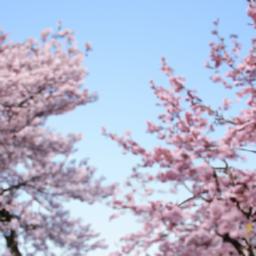}
  \includegraphics[width=0.135\linewidth]{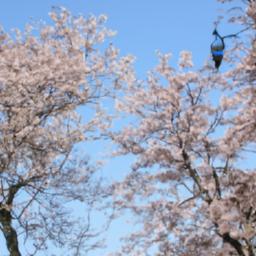}
   \includegraphics[width=0.135\linewidth]{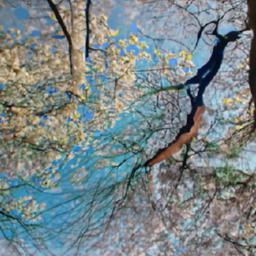}
  \includegraphics[width=0.135\linewidth]{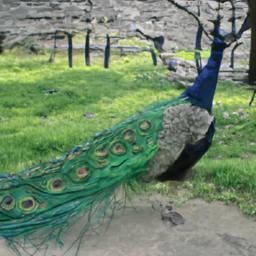}
  \includegraphics[width=0.135\linewidth]{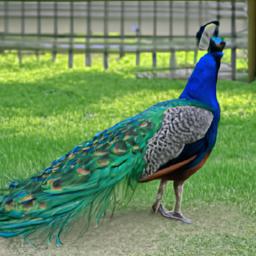}
  \includegraphics[width=0.135\linewidth]{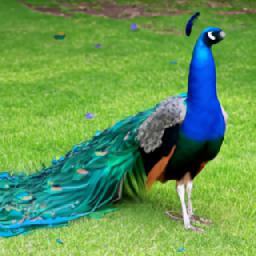}
  \includegraphics[width=0.135\linewidth]{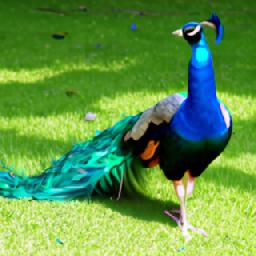}
\tabularnewline
    \raisebox{0.1in}{\rotatebox{90}{\small \emph{$1.0$}
 }}
  \includegraphics[width=0.135\linewidth]{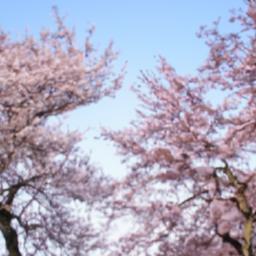}
  \includegraphics[width=0.135\linewidth]{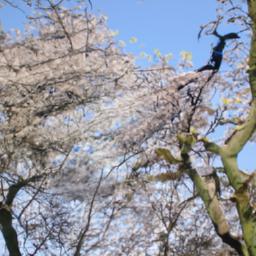}
   \includegraphics[width=0.135\linewidth]{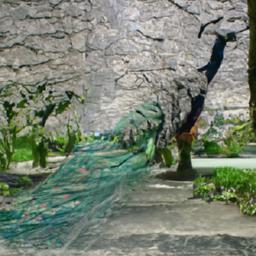}
  \includegraphics[width=0.135\linewidth]{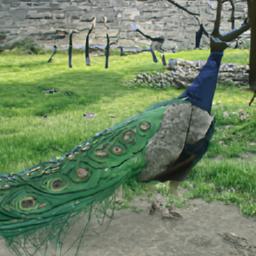}
  \includegraphics[width=0.135\linewidth]{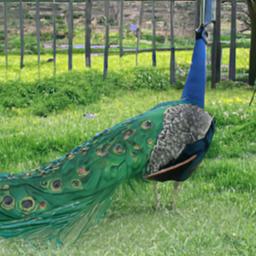}
  \includegraphics[width=0.135\linewidth]{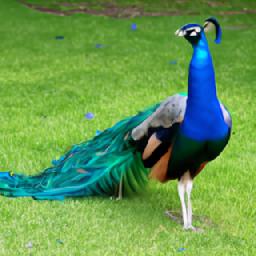}
  \includegraphics[width=0.135\linewidth]{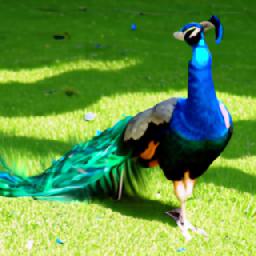}
    \tabularnewline
        \raisebox{0.1in}{\rotatebox{90}{
 }}
 \hspace{0.5mm}
   $\scriptsize{Text}\xrightarrow[\text{weighing factor}]{\hspace*{3cm}}\scriptsize{Class}$
  \tabularnewline
\vspace{2mm}
\vspace{-2\baselineskip}
\end{tabular}}
\vspace{-0.8cm}
\hspace{20pt}\captionof{figure}{\textbf{The role of reliability factor for multimodal generation.} Text Prompt is "A garden of cherry blossom trees, Class is "84: Peacock". y-axis, higher the value, higher the reliability for the text model}
\label{fig:glideablation12}
\vspace{-2mm}
\end{figure*}%

\begin{figure*}[tb!]
    \centering
    \setlength{\tabcolsep}{0.5pt}
    {\small
    \renewcommand{\arraystretch}{0.5} 
    \begin{tabular}{c c c c c c c c c c}
    \captionsetup{type=figure, font=scriptsize}
    \raisebox{0.1in}{\rotatebox{90}{\small \emph{$0.0$}
 }}
  \includegraphics[width=0.135\linewidth]{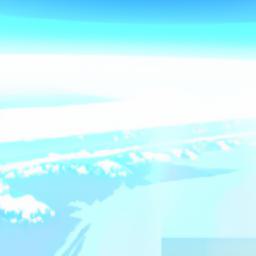}
  \includegraphics[width=0.135\linewidth]{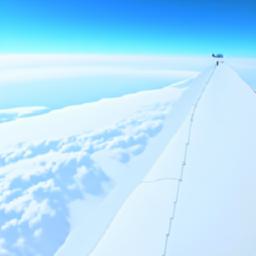}
  \includegraphics[width=0.135\linewidth]{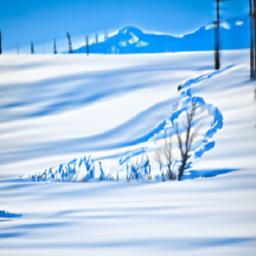}
  \includegraphics[width=0.135\linewidth]{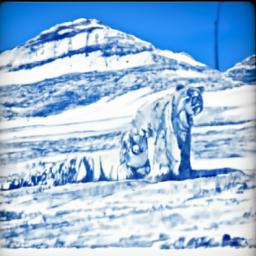}
  \includegraphics[width=0.135\linewidth]{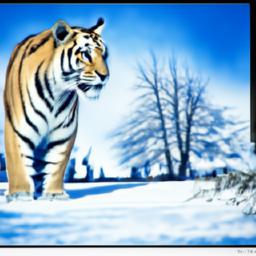}
  \includegraphics[width=0.135\linewidth]{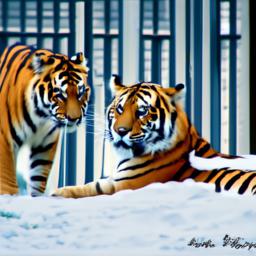}
  \includegraphics[width=0.135\linewidth]{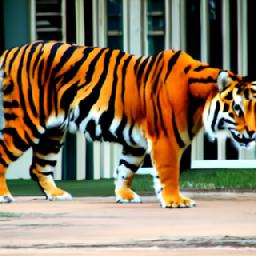}
 \tabularnewline
     \raisebox{0.1in}{\rotatebox{90}{\small \emph{$0.2$}
 }}
  \includegraphics[width=0.135\linewidth]{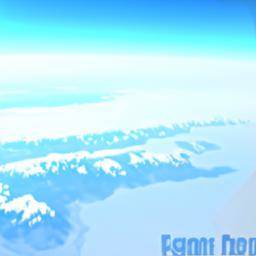}
  \includegraphics[width=0.135\linewidth]{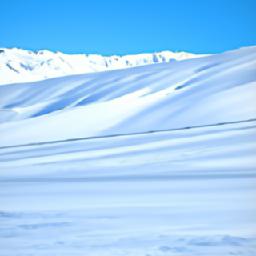}
  \includegraphics[width=0.135\linewidth]{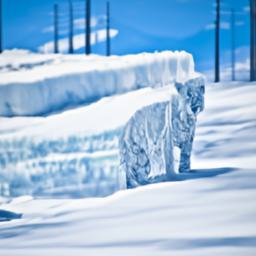}
  \includegraphics[width=0.135\linewidth]{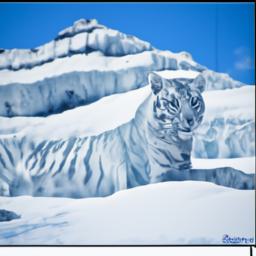}
  \includegraphics[width=0.135\linewidth]{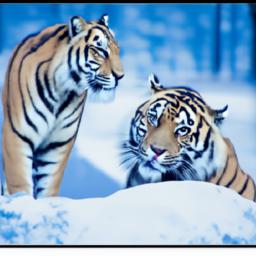}
  \includegraphics[width=0.135\linewidth]{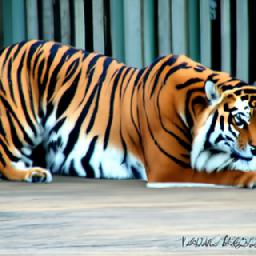}
  \includegraphics[width=0.135\linewidth]{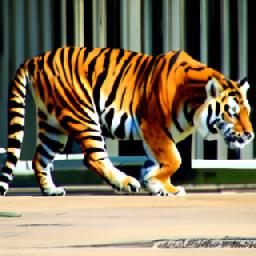}
\tabularnewline
    \raisebox{0.1in}{\rotatebox{90}{\small \emph{$0.4$}
 }}
  \includegraphics[width=0.135\linewidth]{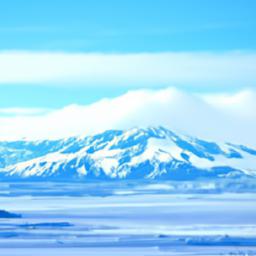}
  \includegraphics[width=0.135\linewidth]{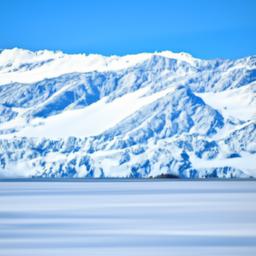}
  \includegraphics[width=0.135\linewidth]{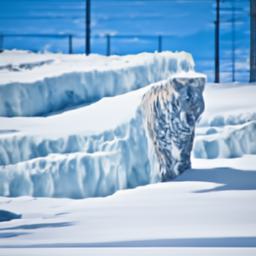}
  \includegraphics[width=0.135\linewidth]{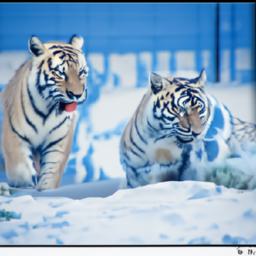}
  \includegraphics[width=0.135\linewidth]{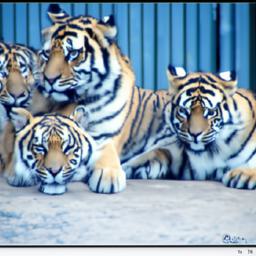}
  \includegraphics[width=0.135\linewidth]{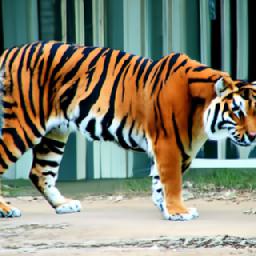}
  \includegraphics[width=0.135\linewidth]{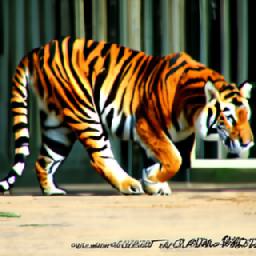}
\tabularnewline
    \raisebox{0.1in}{\rotatebox{90}{\small \emph{$0.5$}
 }}
  \includegraphics[width=0.135\linewidth]{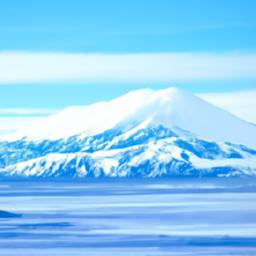}
  \includegraphics[width=0.135\linewidth]{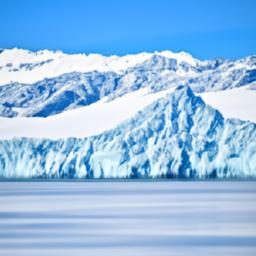}
  \includegraphics[width=0.135\linewidth]{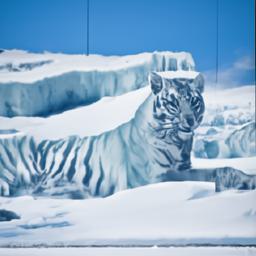}
  \includegraphics[width=0.135\linewidth]{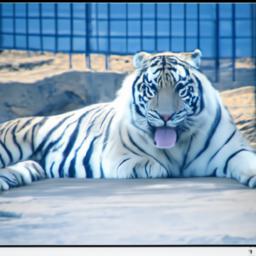}
  \includegraphics[width=0.135\linewidth]{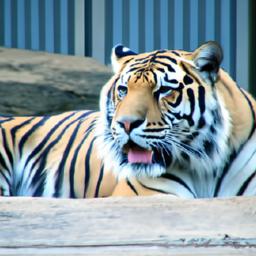}
  \includegraphics[width=0.135\linewidth]{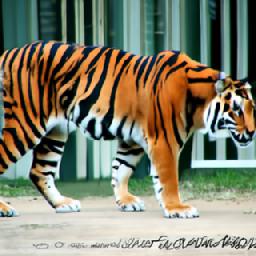}
  \includegraphics[width=0.135\linewidth]{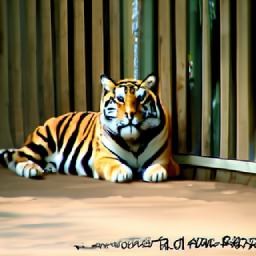}
\tabularnewline
    \raisebox{0.1in}{\rotatebox{90}{\small \emph{$0.6$}
 }}
  \includegraphics[width=0.135\linewidth]{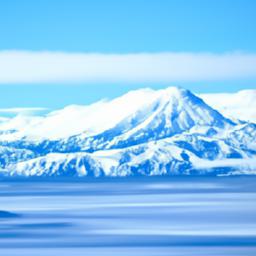}
  \includegraphics[width=0.135\linewidth]{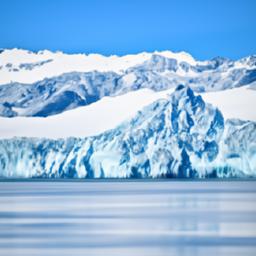}
  \includegraphics[width=0.135\linewidth]{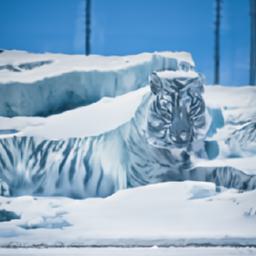}
  \includegraphics[width=0.135\linewidth]{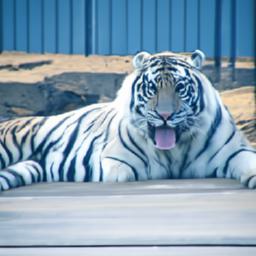}
  \includegraphics[width=0.135\linewidth]{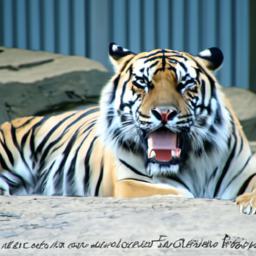}
  \includegraphics[width=0.135\linewidth]{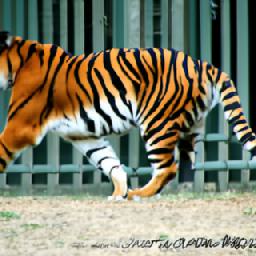}
  \includegraphics[width=0.135\linewidth]{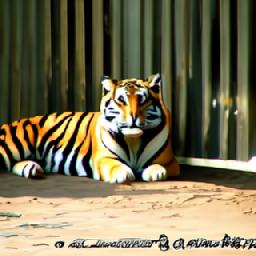}
\tabularnewline
    \raisebox{0.1in}{\rotatebox{90}{\small \emph{$0.8$}
 }}
  \includegraphics[width=0.135\linewidth]{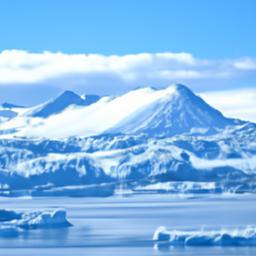}
  \includegraphics[width=0.135\linewidth]{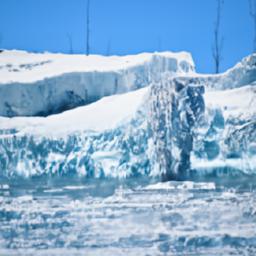}
   \includegraphics[width=0.135\linewidth]{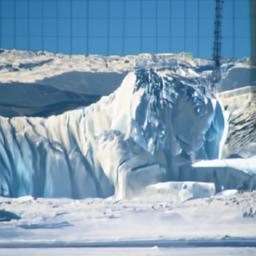}
  \includegraphics[width=0.135\linewidth]{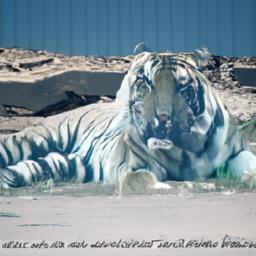}
  \includegraphics[width=0.135\linewidth]{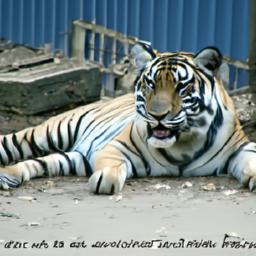}
  \includegraphics[width=0.135\linewidth]{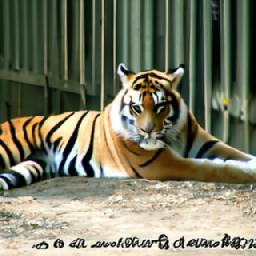}
  \includegraphics[width=0.135\linewidth]{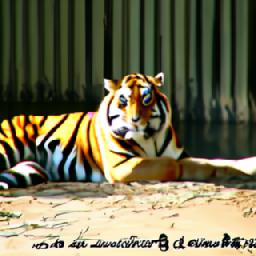}
\tabularnewline
    \raisebox{0.1in}{\rotatebox{90}{\small \emph{$1.0$}
 }}
  \includegraphics[width=0.135\linewidth]{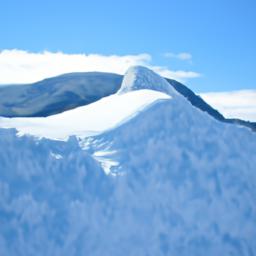}
  \includegraphics[width=0.135\linewidth]{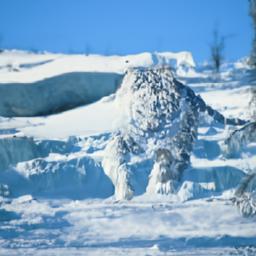}
   \includegraphics[width=0.135\linewidth]{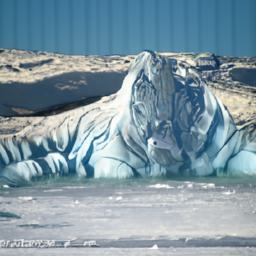}
  \includegraphics[width=0.135\linewidth]{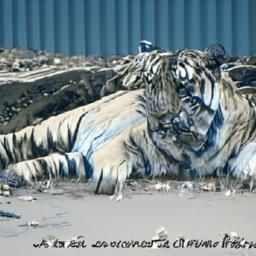}
  \includegraphics[width=0.135\linewidth]{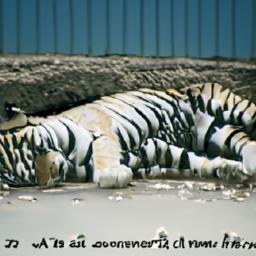}
  \includegraphics[width=0.135\linewidth]{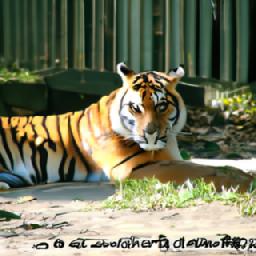}
  \includegraphics[width=0.135\linewidth]{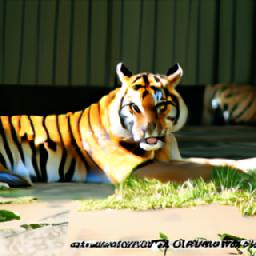}
    \tabularnewline
        \raisebox{0.1in}{\rotatebox{90}{
 }}
 \hspace{0.5mm}
   $\scriptsize{Text}\xrightarrow[\text{weighing factor}]{\hspace*{3cm}}\scriptsize{Class}$
  \tabularnewline
\vspace{2mm}
\vspace{-2\baselineskip}
\end{tabular}}
\vspace{-0.8cm}
\hspace{20pt}\captionof{figure}{\textbf{The role of reliability factor for multimodal generation.} Text Prompt is "A ice mountain, Class is "292: Tiger" y-axis, higher the value, higher the reliability for the text model}
\label{fig:glideablation13}
\vspace{-2mm}
\end{figure*}%

\begin{figure*}[tb!]
    \centering
    \setlength{\tabcolsep}{0.5pt}
    {\small
    \renewcommand{\arraystretch}{0.5} 
    \begin{tabular}{c c c c c c c c c c}
    \captionsetup{type=figure, font=scriptsize}
    \raisebox{0.1in}{\rotatebox{90}{\small \emph{$0.0$}
 }}
  \includegraphics[width=0.135\linewidth]{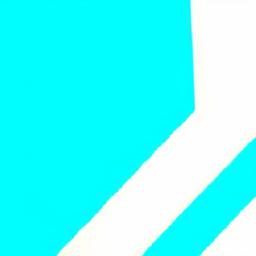}
  \includegraphics[width=0.135\linewidth]{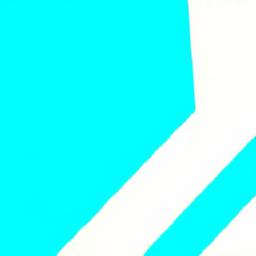}
  \includegraphics[width=0.135\linewidth]{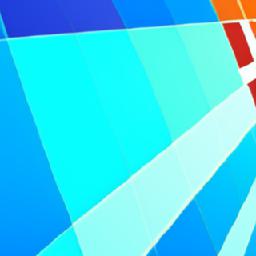}
  \includegraphics[width=0.135\linewidth]{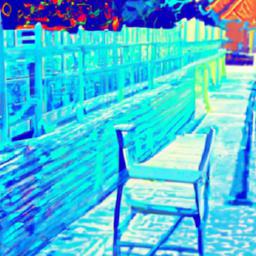}
  \includegraphics[width=0.135\linewidth]{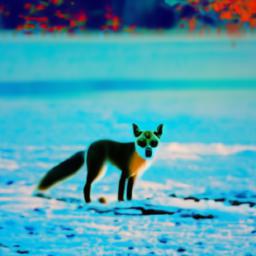}
  \includegraphics[width=0.135\linewidth]{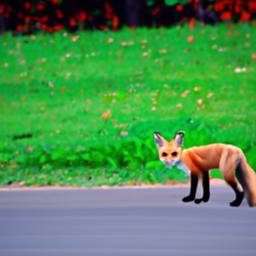}
  \includegraphics[width=0.135\linewidth]{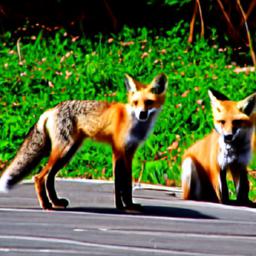}
 \tabularnewline
     \raisebox{0.1in}{\rotatebox{90}{\small \emph{$0.2$}
 }}
  \includegraphics[width=0.135\linewidth]{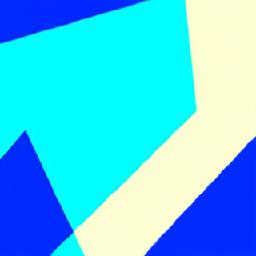}
  \includegraphics[width=0.135\linewidth]{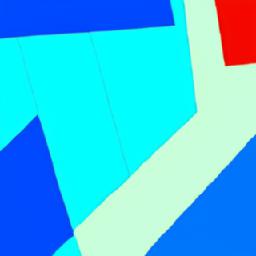}
  \includegraphics[width=0.135\linewidth]{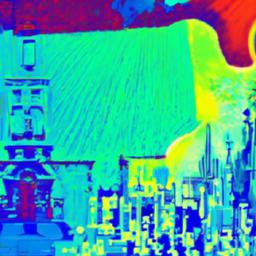}
  \includegraphics[width=0.135\linewidth]{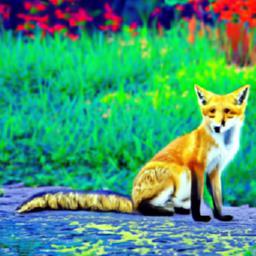}
  \includegraphics[width=0.135\linewidth]{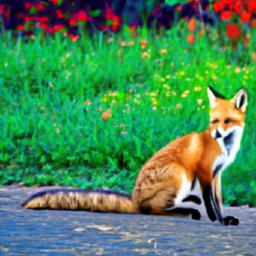}
  \includegraphics[width=0.135\linewidth]{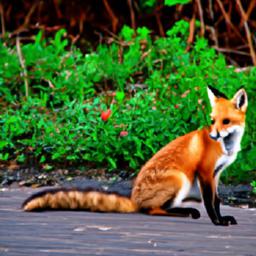}
  \includegraphics[width=0.135\linewidth]{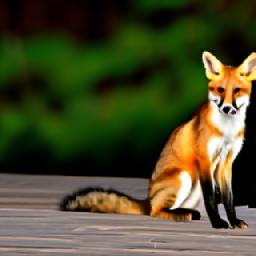}
\tabularnewline
    \raisebox{0.1in}{\rotatebox{90}{\small \emph{$0.4$}
 }}
  \includegraphics[width=0.135\linewidth]{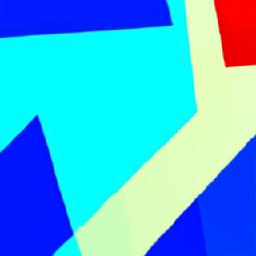}
  \includegraphics[width=0.135\linewidth]{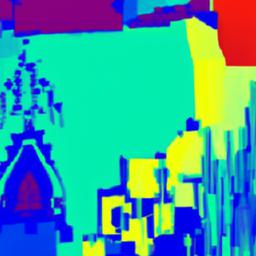}
  \includegraphics[width=0.135\linewidth]{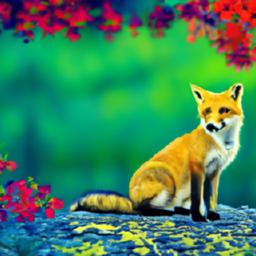}
  \includegraphics[width=0.135\linewidth]{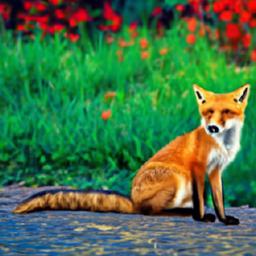}
  \includegraphics[width=0.135\linewidth]{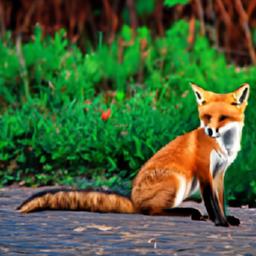}
  \includegraphics[width=0.135\linewidth]{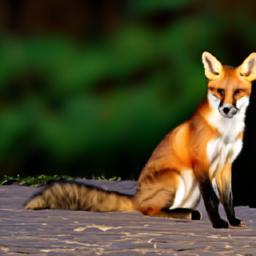}
  \includegraphics[width=0.135\linewidth]{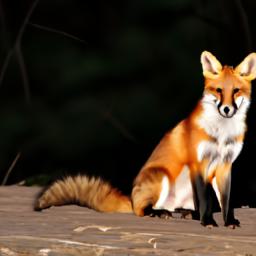}
\tabularnewline
    \raisebox{0.1in}{\rotatebox{90}{\small \emph{$0.5$}
 }}
  \includegraphics[width=0.135\linewidth]{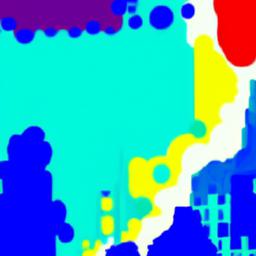}
  \includegraphics[width=0.135\linewidth]{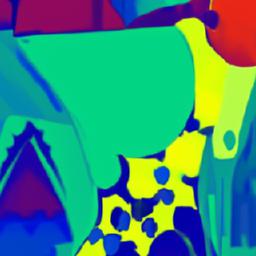}
  \includegraphics[width=0.135\linewidth]{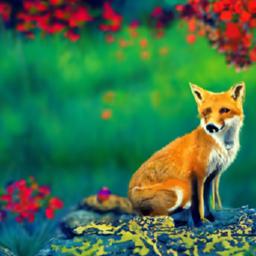}
  \includegraphics[width=0.135\linewidth]{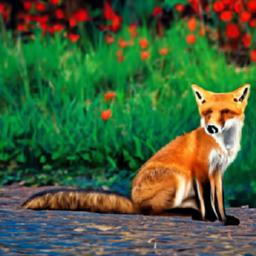}
  \includegraphics[width=0.135\linewidth]{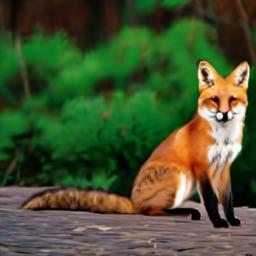}
  \includegraphics[width=0.135\linewidth]{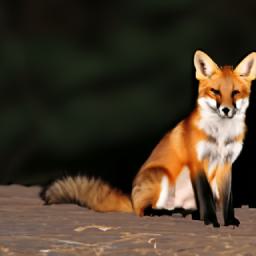}
  \includegraphics[width=0.135\linewidth]{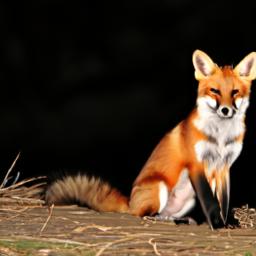}
\tabularnewline
    \raisebox{0.1in}{\rotatebox{90}{\small \emph{$0.6$}
 }}
  \includegraphics[width=0.135\linewidth]{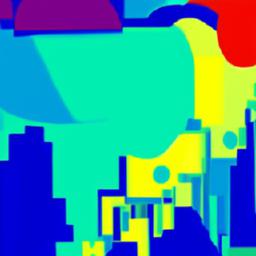}
  \includegraphics[width=0.135\linewidth]{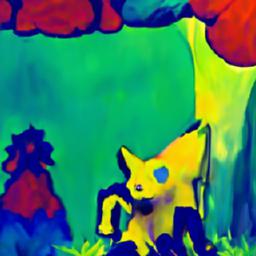}
  \includegraphics[width=0.135\linewidth]{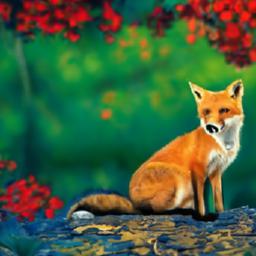}
  \includegraphics[width=0.135\linewidth]{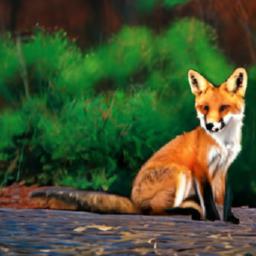}
  \includegraphics[width=0.135\linewidth]{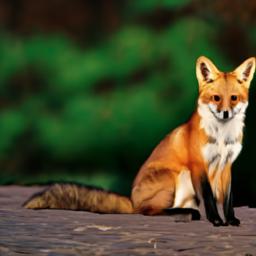}
  \includegraphics[width=0.135\linewidth]{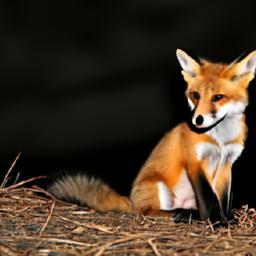}
  \includegraphics[width=0.135\linewidth]{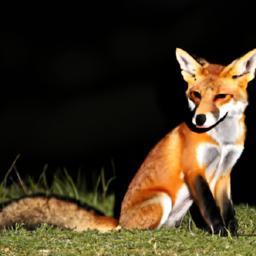}
\tabularnewline
    \raisebox{0.1in}{\rotatebox{90}{\small \emph{$0.8$}
 }}
  \includegraphics[width=0.135\linewidth]{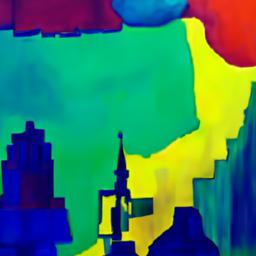}
  \includegraphics[width=0.135\linewidth]{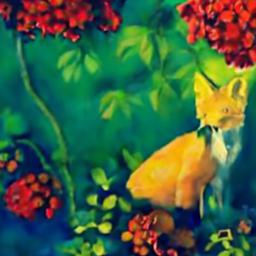}
   \includegraphics[width=0.135\linewidth]{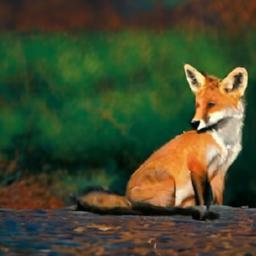}
  \includegraphics[width=0.135\linewidth]{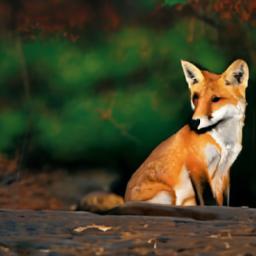}
  \includegraphics[width=0.135\linewidth]{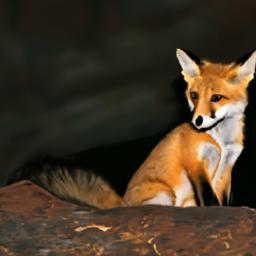}
  \includegraphics[width=0.135\linewidth]{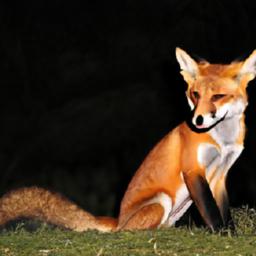}
  \includegraphics[width=0.135\linewidth]{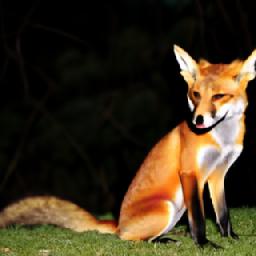}
\tabularnewline
    \raisebox{0.1in}{\rotatebox{90}{\small \emph{$1.0$}
 }}
  \includegraphics[width=0.135\linewidth]{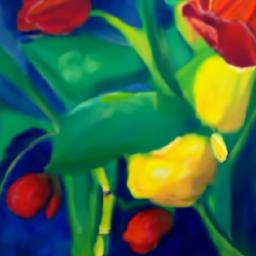}
  \includegraphics[width=0.135\linewidth]{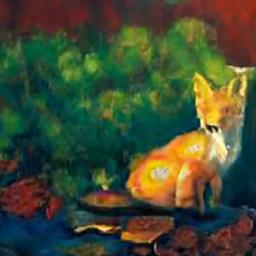}
   \includegraphics[width=0.135\linewidth]{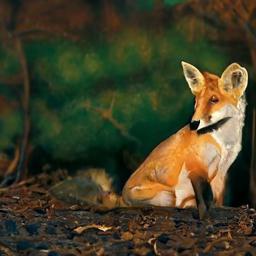}
  \includegraphics[width=0.135\linewidth]{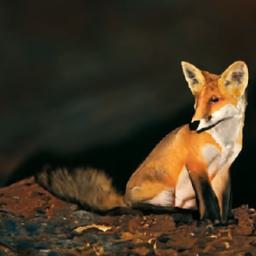}
  \includegraphics[width=0.135\linewidth]{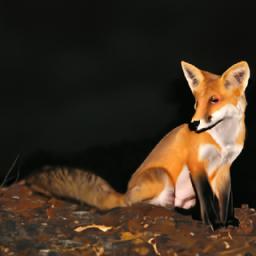}
  \includegraphics[width=0.135\linewidth]{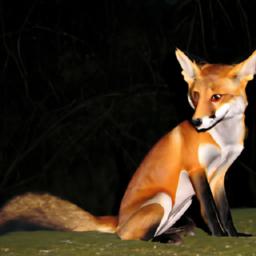}
  \includegraphics[width=0.135\linewidth]{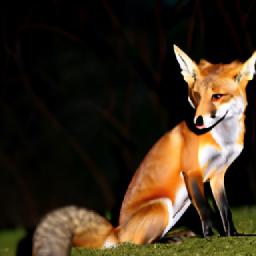}
    \tabularnewline
        \raisebox{0.1in}{\rotatebox{90}{
 }}
 \hspace{0.5mm}
   $\scriptsize{Text}\xrightarrow[\text{weighing factor}]{\hspace*{3cm}}\scriptsize{Class}$
  \tabularnewline
\vspace{2mm}
\vspace{-2\baselineskip}
\end{tabular}}
\vspace{-0.8cm}
\hspace{20pt}\captionof{figure}{\textbf{The role of reliability factor for multimodal generation.} Text Prompt is "An oil painting, Class is "277: Red Fox". y-axis, higher the value, higher the reliability for the text model}
\label{fig:glideablation15}
\vspace{-2mm}
\end{figure*}%

\begin{figure*}[tb!]
    \centering
    \setlength{\tabcolsep}{0.5pt}
    {\small
    \renewcommand{\arraystretch}{0.5} 
    \begin{tabular}{c c c c c c c c c c}
    \captionsetup{type=figure, font=scriptsize}
    \raisebox{0.1in}{\rotatebox{90}{\small \emph{$0.0$}
 }}
  \includegraphics[width=0.135\linewidth]{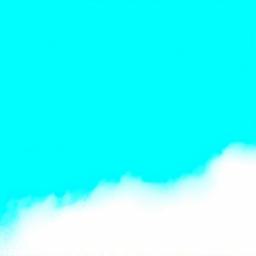}
  \includegraphics[width=0.135\linewidth]{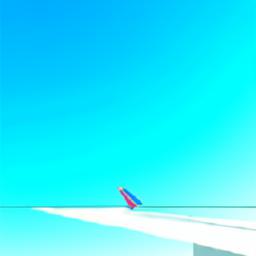}
  \includegraphics[width=0.135\linewidth]{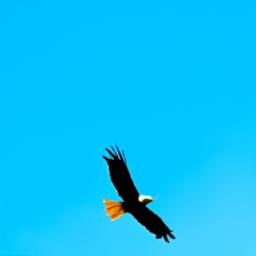}
  \includegraphics[width=0.135\linewidth]{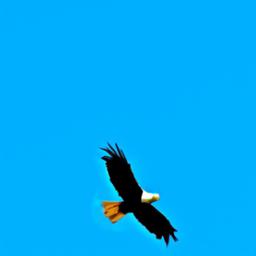}
  \includegraphics[width=0.135\linewidth]{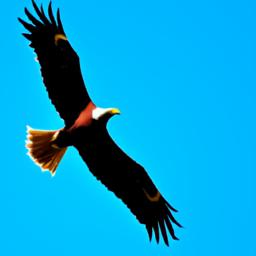}
  \includegraphics[width=0.135\linewidth]{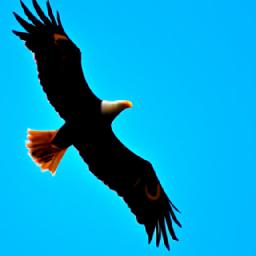}
  \includegraphics[width=0.135\linewidth]{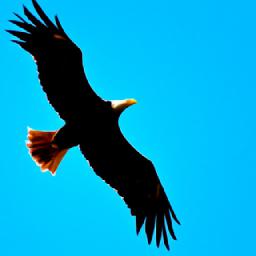}
 \tabularnewline
     \raisebox{0.1in}{\rotatebox{90}{\small \emph{$0.2$}
 }}
  \includegraphics[width=0.135\linewidth]{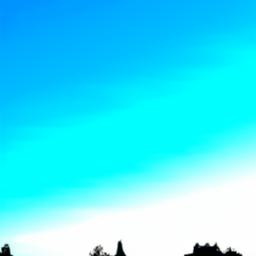}
  \includegraphics[width=0.135\linewidth]{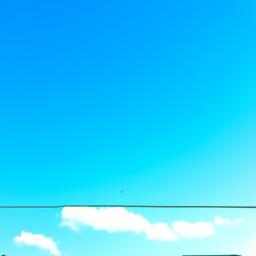}
  \includegraphics[width=0.135\linewidth]{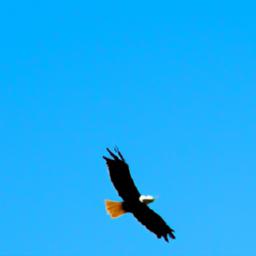}
  \includegraphics[width=0.135\linewidth]{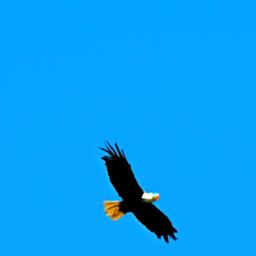}
  \includegraphics[width=0.135\linewidth]{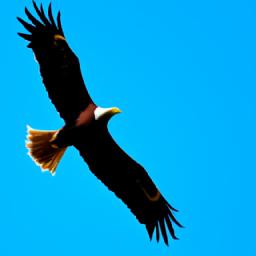}
  \includegraphics[width=0.135\linewidth]{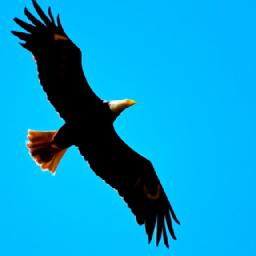}
  \includegraphics[width=0.135\linewidth]{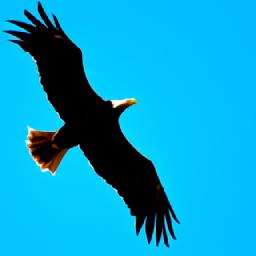}
\tabularnewline
    \raisebox{0.1in}{\rotatebox{90}{\small \emph{$0.4$}
 }}
  \includegraphics[width=0.135\linewidth]{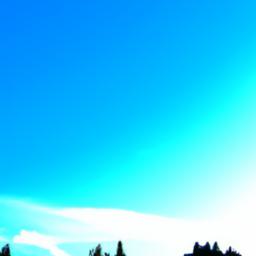}
  \includegraphics[width=0.135\linewidth]{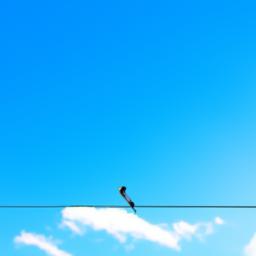}
  \includegraphics[width=0.135\linewidth]{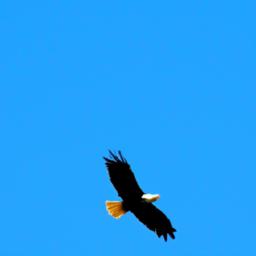}
  \includegraphics[width=0.135\linewidth]{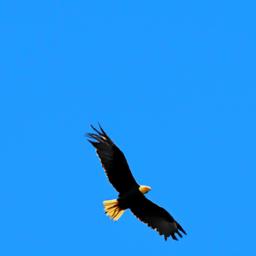}
  \includegraphics[width=0.135\linewidth]{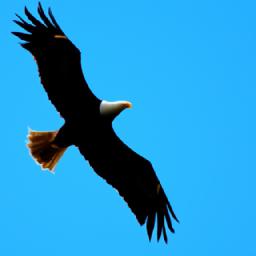}
  \includegraphics[width=0.135\linewidth]{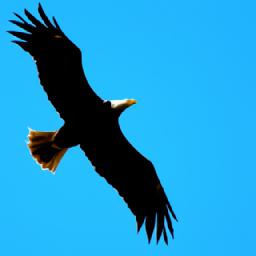}
  \includegraphics[width=0.135\linewidth]{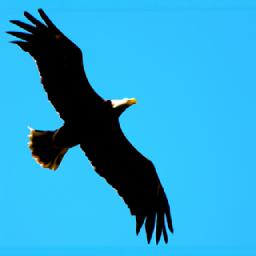}
\tabularnewline
    \raisebox{0.1in}{\rotatebox{90}{\small \emph{$0.5$}
 }}
  \includegraphics[width=0.135\linewidth]{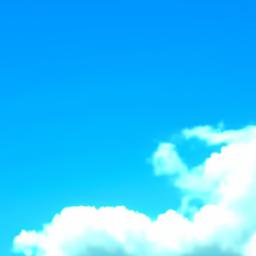}
  \includegraphics[width=0.135\linewidth]{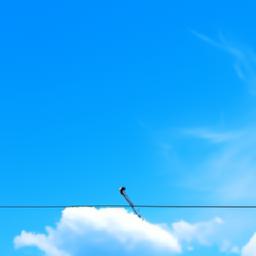}
  \includegraphics[width=0.135\linewidth]{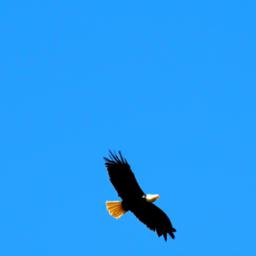}
  \includegraphics[width=0.135\linewidth]{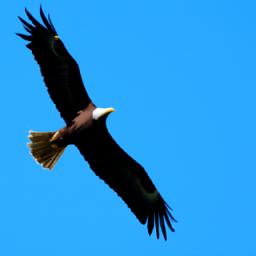}
  \includegraphics[width=0.135\linewidth]{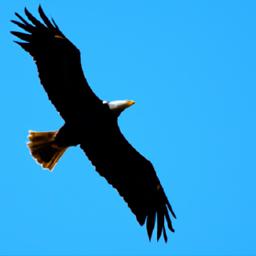}
  \includegraphics[width=0.135\linewidth]{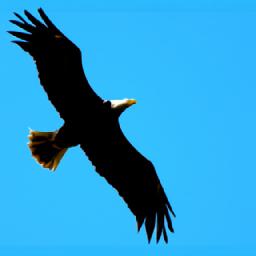}
  \includegraphics[width=0.135\linewidth]{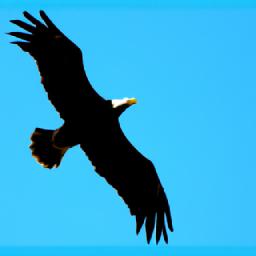}
\tabularnewline
    \raisebox{0.1in}{\rotatebox{90}{\small \emph{$0.6$}
 }}
  \includegraphics[width=0.135\linewidth]{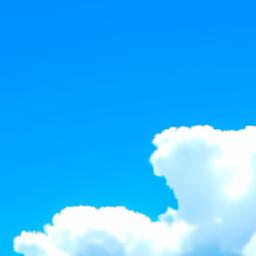}
  \includegraphics[width=0.135\linewidth]{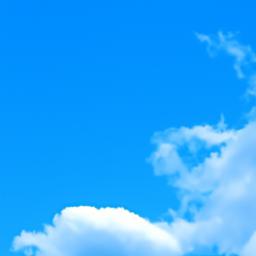}
  \includegraphics[width=0.135\linewidth]{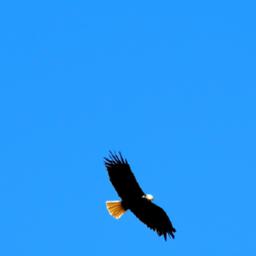}
  \includegraphics[width=0.135\linewidth]{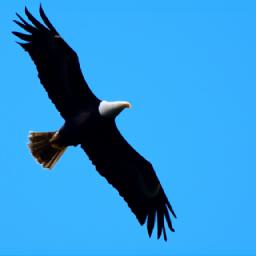}
  \includegraphics[width=0.135\linewidth]{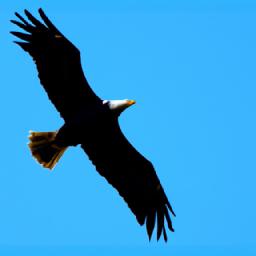}
  \includegraphics[width=0.135\linewidth]{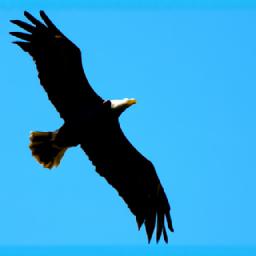}
  \includegraphics[width=0.135\linewidth]{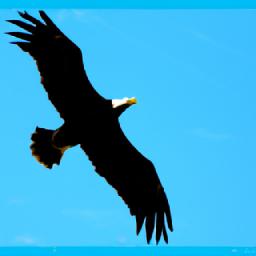}
\tabularnewline
    \raisebox{0.1in}{\rotatebox{90}{\small \emph{$0.8$}
 }}
  \includegraphics[width=0.135\linewidth]{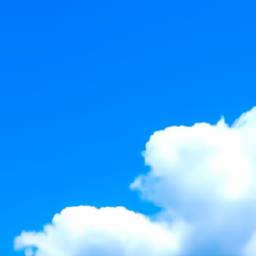}
  \includegraphics[width=0.135\linewidth]{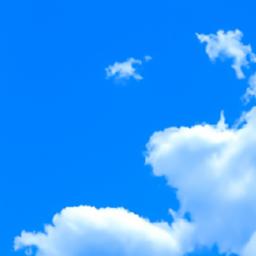}
   \includegraphics[width=0.135\linewidth]{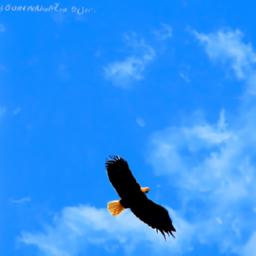}
  \includegraphics[width=0.135\linewidth]{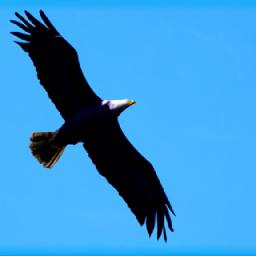}
  \includegraphics[width=0.135\linewidth]{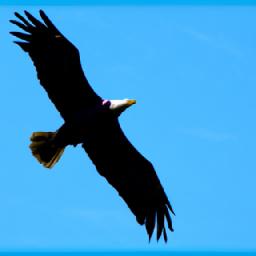}
  \includegraphics[width=0.135\linewidth]{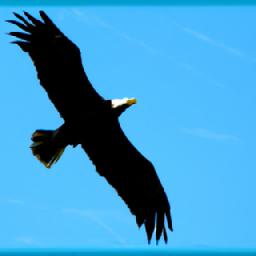}
  \includegraphics[width=0.135\linewidth]{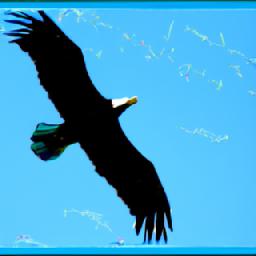}
\tabularnewline
    \raisebox{0.1in}{\rotatebox{90}{\small \emph{$1.0$}
 }}
  \includegraphics[width=0.135\linewidth]{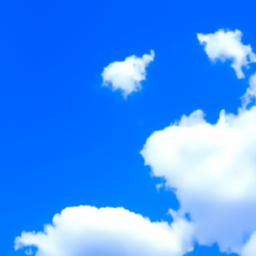}
  \includegraphics[width=0.135\linewidth]{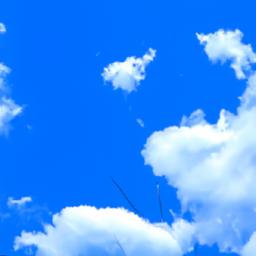}
   \includegraphics[width=0.135\linewidth]{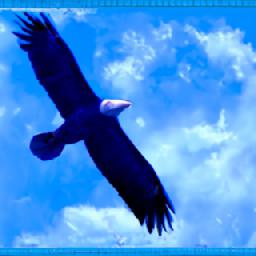}
  \includegraphics[width=0.135\linewidth]{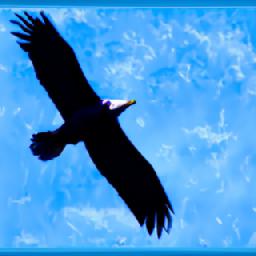}
  \includegraphics[width=0.135\linewidth]{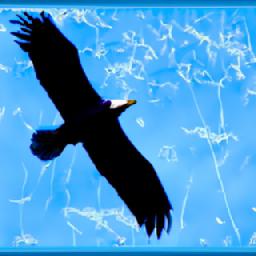}
  \includegraphics[width=0.135\linewidth]{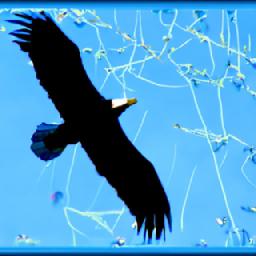}
  \includegraphics[width=0.135\linewidth]{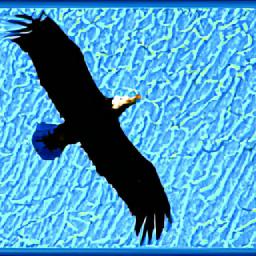}
    \tabularnewline
        \raisebox{0.1in}{\rotatebox{90}{
 }}
 \hspace{0.5mm}
   $\scriptsize{Text}\xrightarrow[\text{weighing factor}]{\hspace*{3cm}}\scriptsize{Class}$
  \tabularnewline
\vspace{2mm}
\vspace{-2\baselineskip}
\end{tabular}}
\vspace{-0.8cm}
\hspace{20pt}\captionof{figure}{\textbf{The role of reliability factor for multimodal generation.} Text Prompt is "A blue sky with clouds, Class is "22: Eagle". y-axis, higher the value, higher the reliability for the text model}
\label{fig:glideablation14}
\vspace{-2mm}
\end{figure*}%

 \begin{figure*}[t!]
	\centering
			\includegraphics[width=\textwidth]{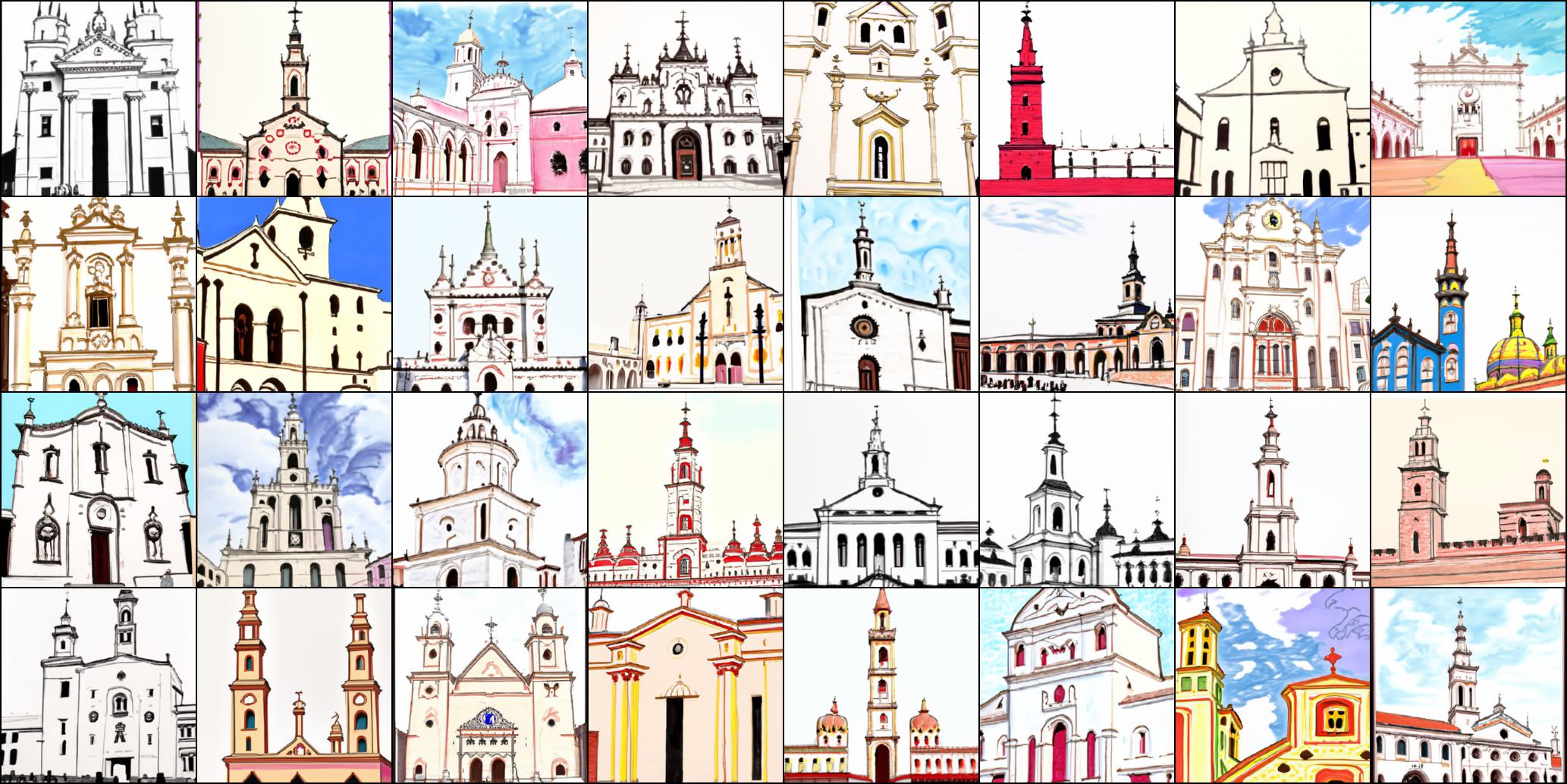}
	\centering
	\caption{Results for text "A crayon drawing" and ImageNet class "663: Monastry".Not cherry-picked}
	\label{fig:t2im3243}
	\vskip -10pt
\end{figure*}
 \begin{figure*}[t!]
	\centering
			\includegraphics[width=\textwidth]{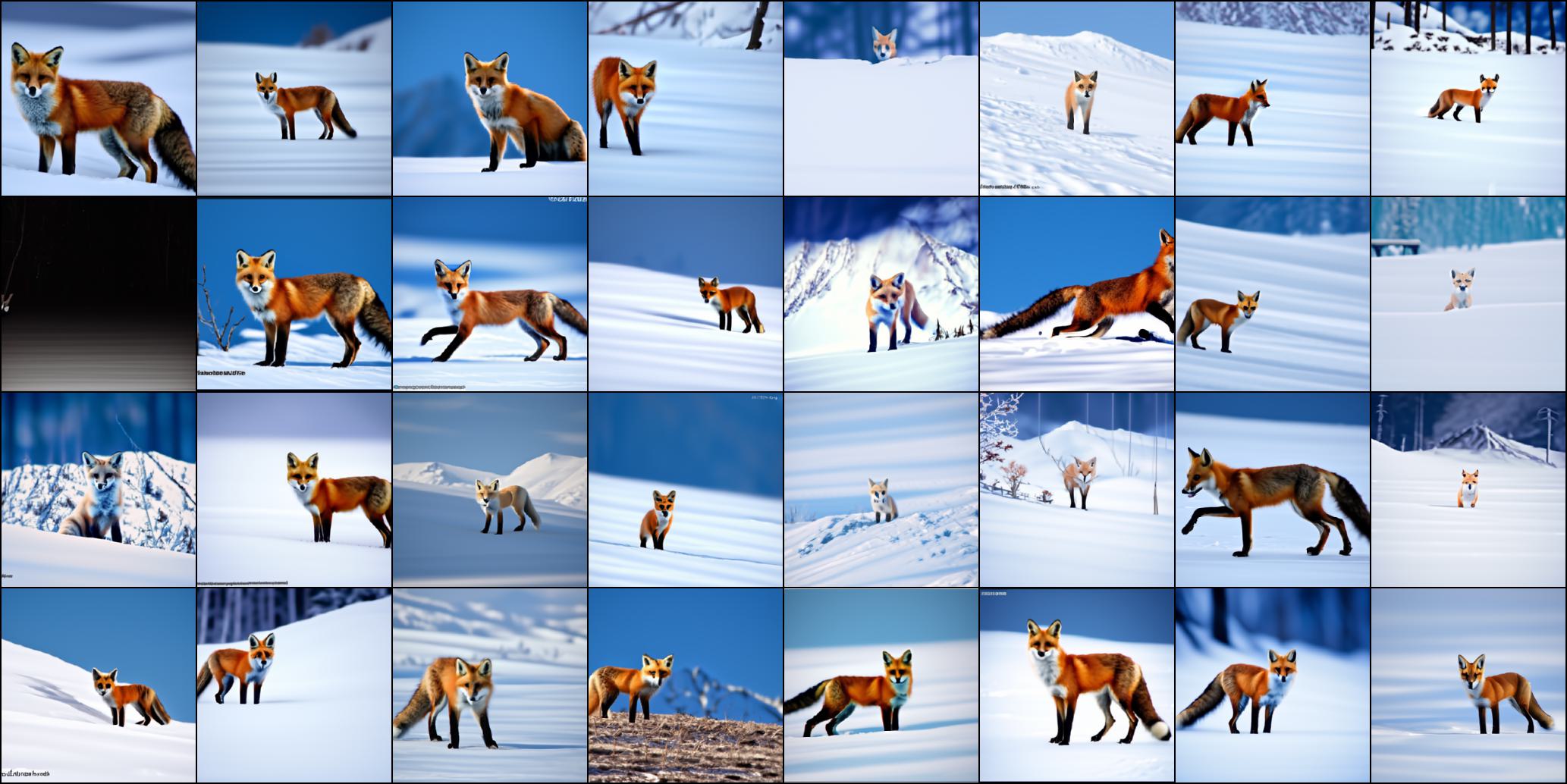}
	\centering
	\caption{Results for text "A snow mountain" and ImageNet class "277:- Red fox".Not cherry-picked}
	\label{fig:t2im3223}
	\vskip -10pt
\end{figure*}
 \begin{figure*}[t!]
	\centering
			\includegraphics[width=\textwidth]{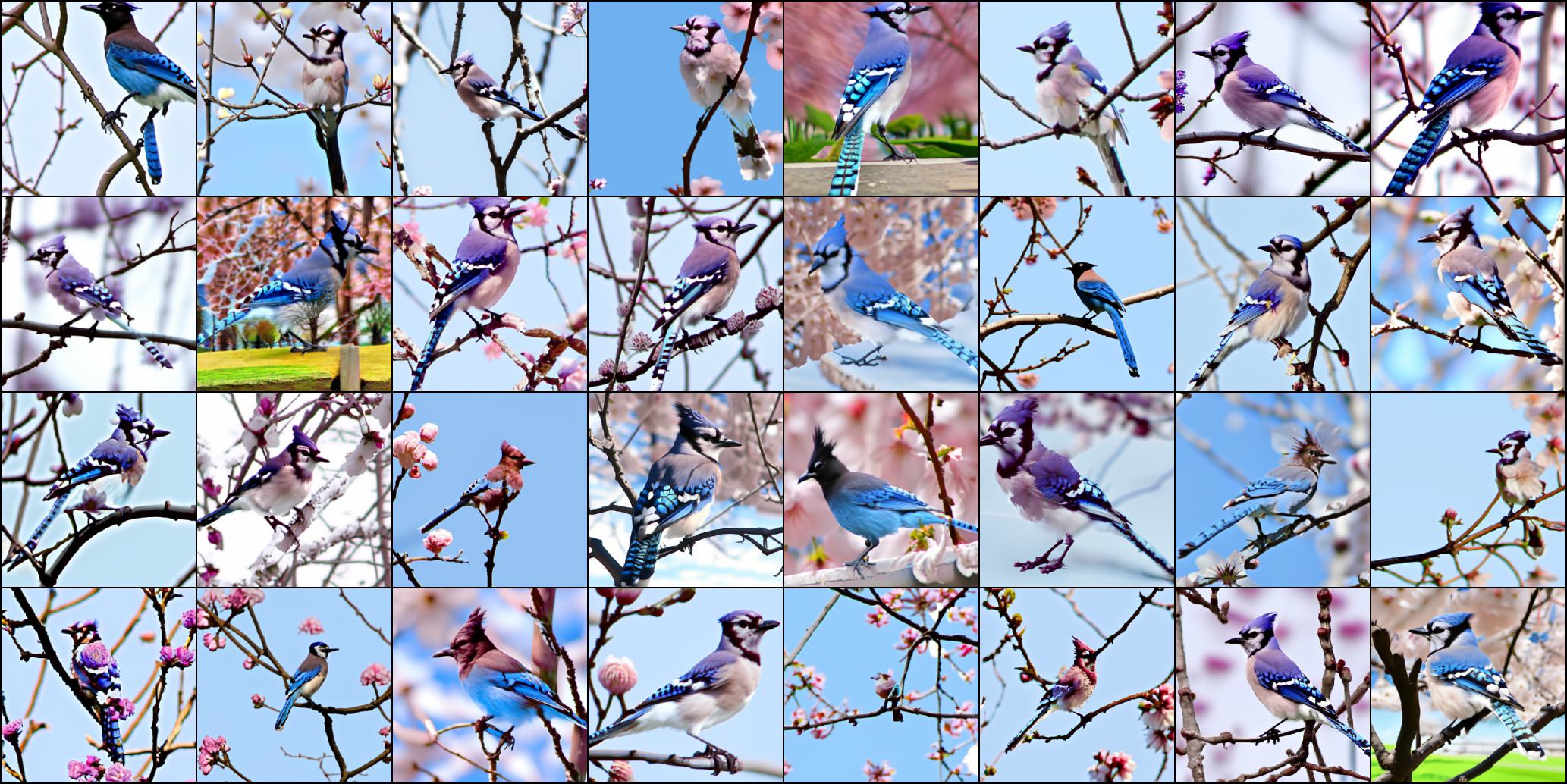}
	\centering
	\caption{Results for text "A cherry blossom tree" and ImageNet class "17: Jay". Not cherry-picked}
	\label{fig:t2im154}
	\vskip -10pt
\end{figure*}
 \begin{figure*}[t!]
	\centering
			\includegraphics[width=\textwidth]{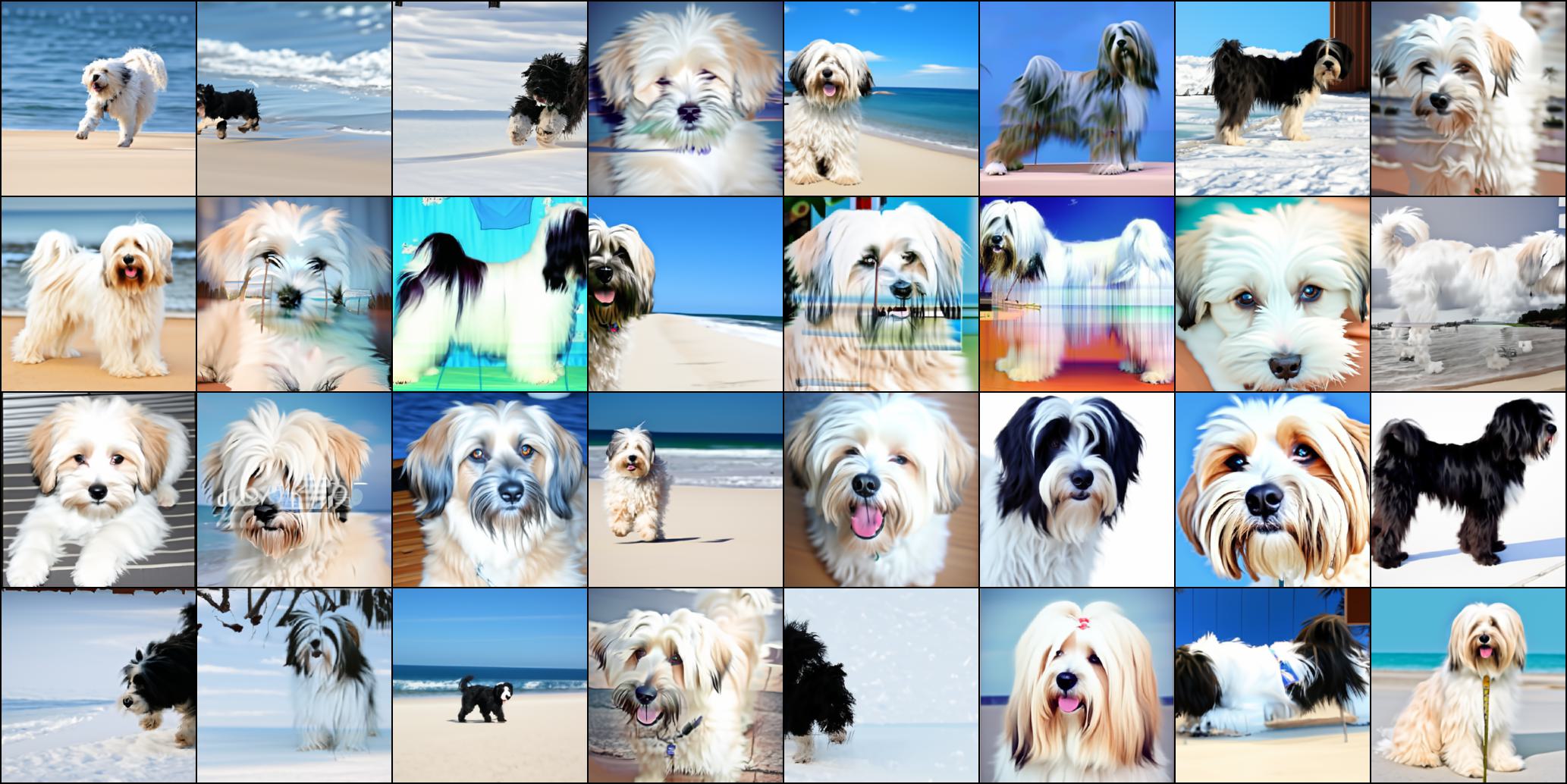}
	\centering
	\caption{Results for text "Photo of a beach" and ImageNet class "200:- Tibetian Terrier".Not cherry-picked}
	\label{fig:t2im143}
	\vskip -10pt
\end{figure*}
 \begin{figure*}[t!]
	\centering
			\includegraphics[width=\textwidth]{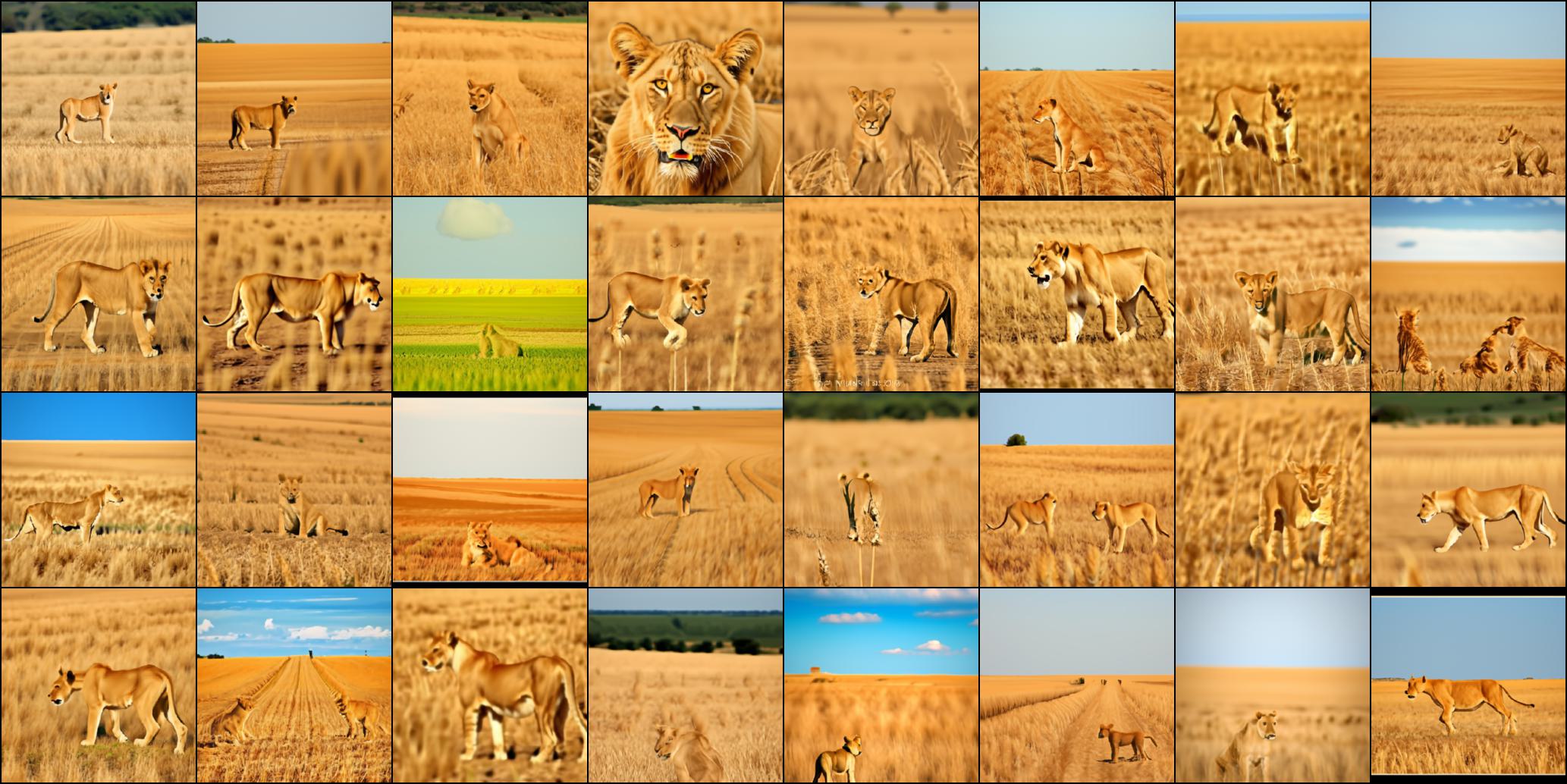}
	\centering
	\caption{Results for text "A wheat field" and ImageNet class "291:- lion".Not cherry-picked}
	\label{fig:t2im432}
	\vskip -10pt
\end{figure*}
 \begin{figure*}[t!]
	\centering
			\includegraphics[width=\textwidth]{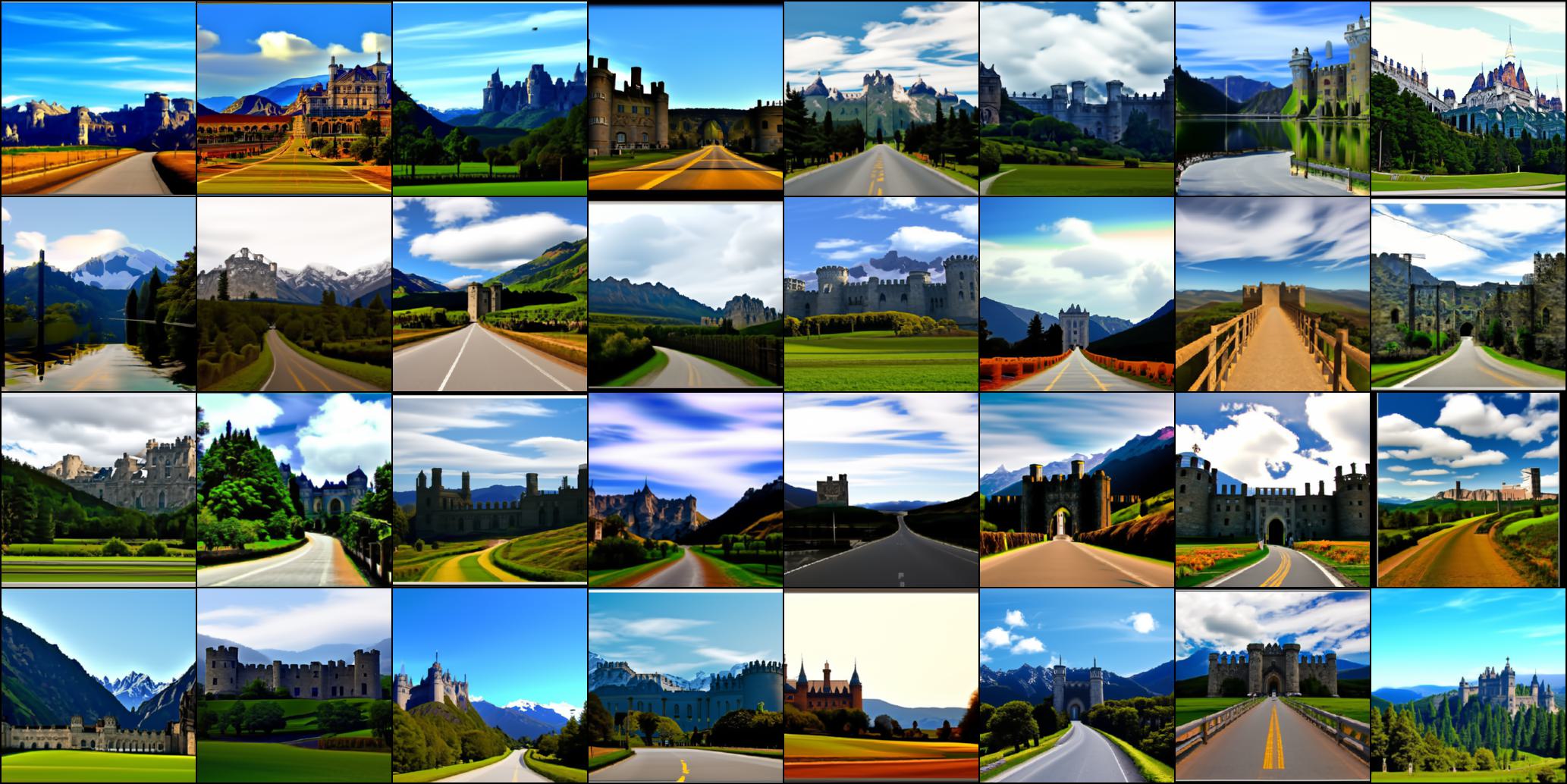}
	\centering
	\caption{Results for text "A road leading into mountains" and ImageNet class "483: A castle". Not cherry-picked}
	\label{fig:t2im132}
	\vskip -10pt
\end{figure*}

 \begin{figure*}[t!]
	\centering
			\includegraphics[width=\textwidth]{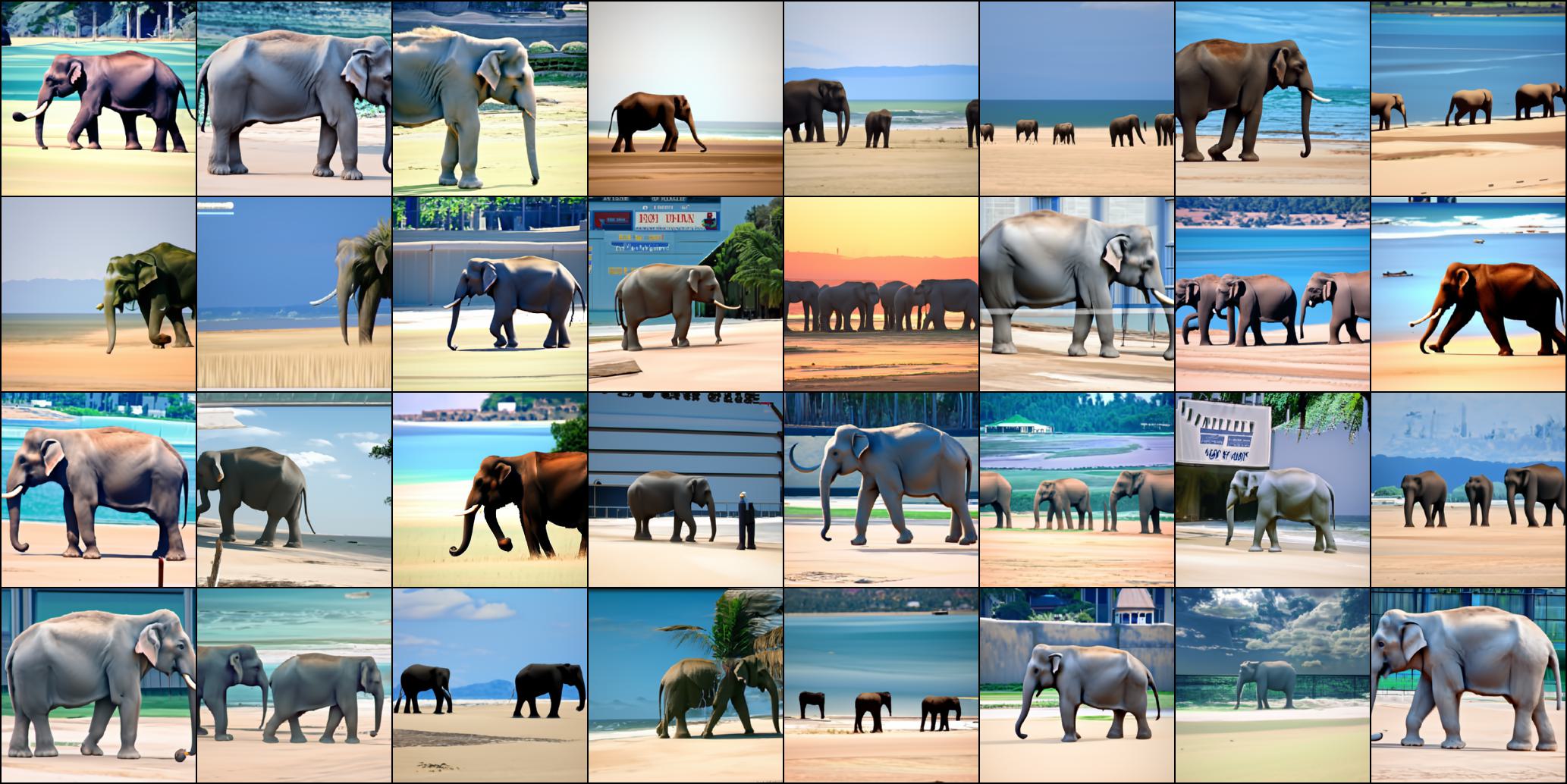}
	\centering
	\caption{Results for text "Photo of a beach" and ImageNet class "385: Elephant". Not cherry-picked}
	\label{fig:t2im121}
	\vskip -10pt
\end{figure*}
 \begin{figure*}[t!]
	\centering
			\includegraphics[width=\textwidth]{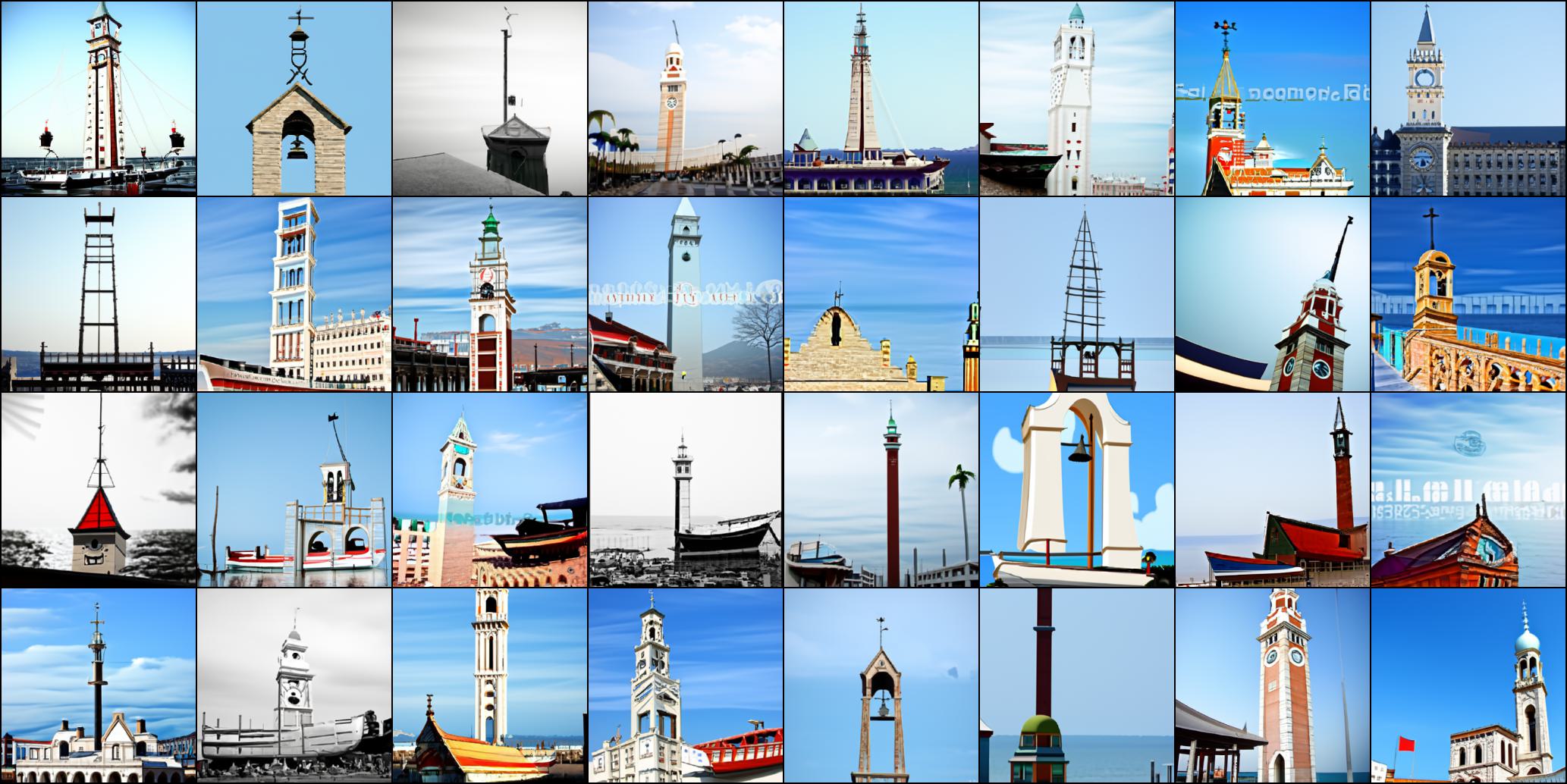}
	\centering
	\caption{Results for text "A canoe on the sea" and ImageNet class "442: A bell cote". Not cherry-picked}
	\label{fig:t2im1111}
	\vskip -10pt
\end{figure*}
 \begin{figure*}[t!]
	\centering
			\includegraphics[width=\textwidth]{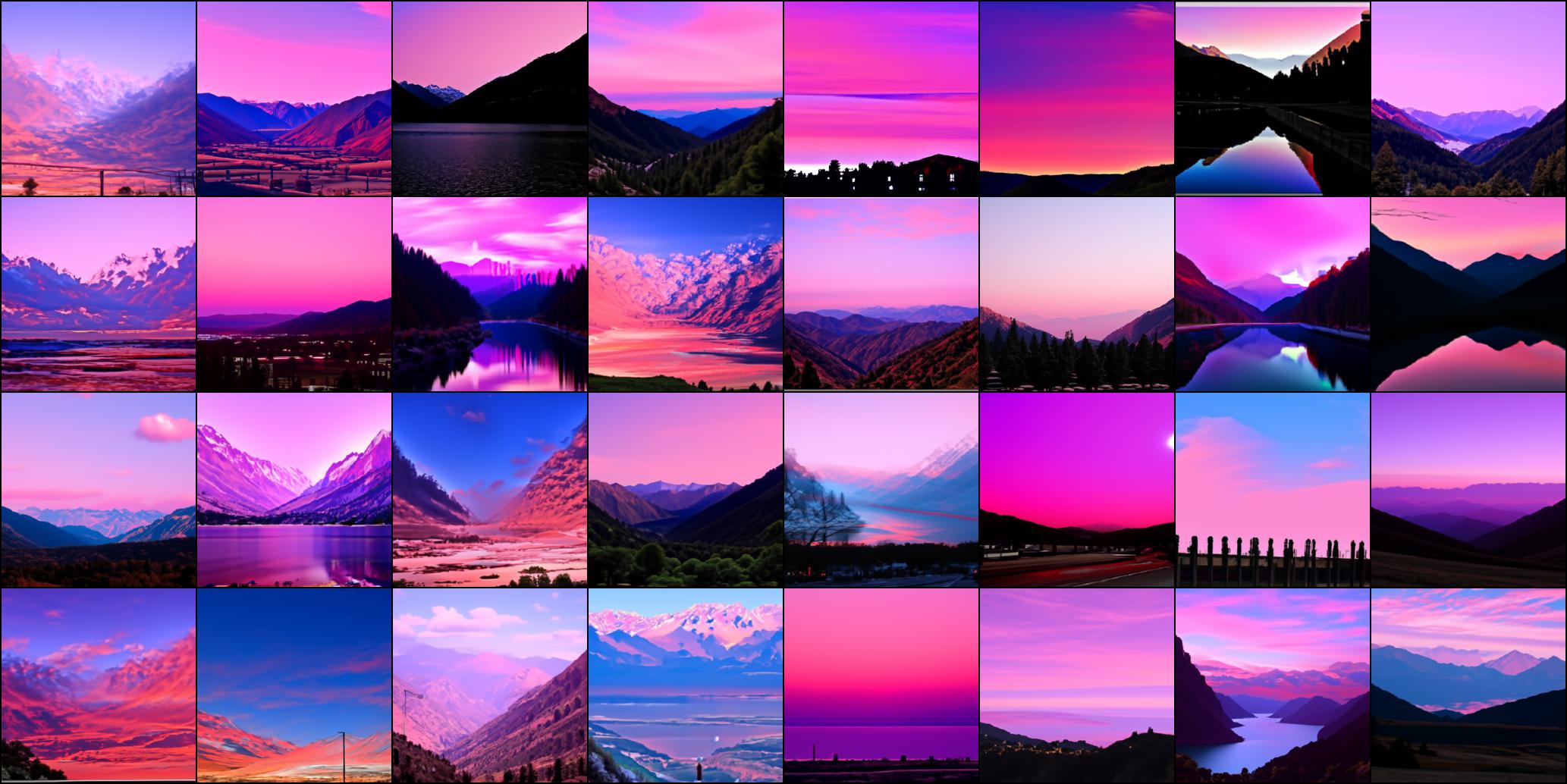}
	\centering
	\caption{Results for text "A pink sky" and ImageNet class "979: A valley". Not cherry-picked}
	\label{fig:t2im111}
	\vskip -10pt
\end{figure*}
 \begin{figure*}[t!]
	\centering
			\includegraphics[width=\textwidth]{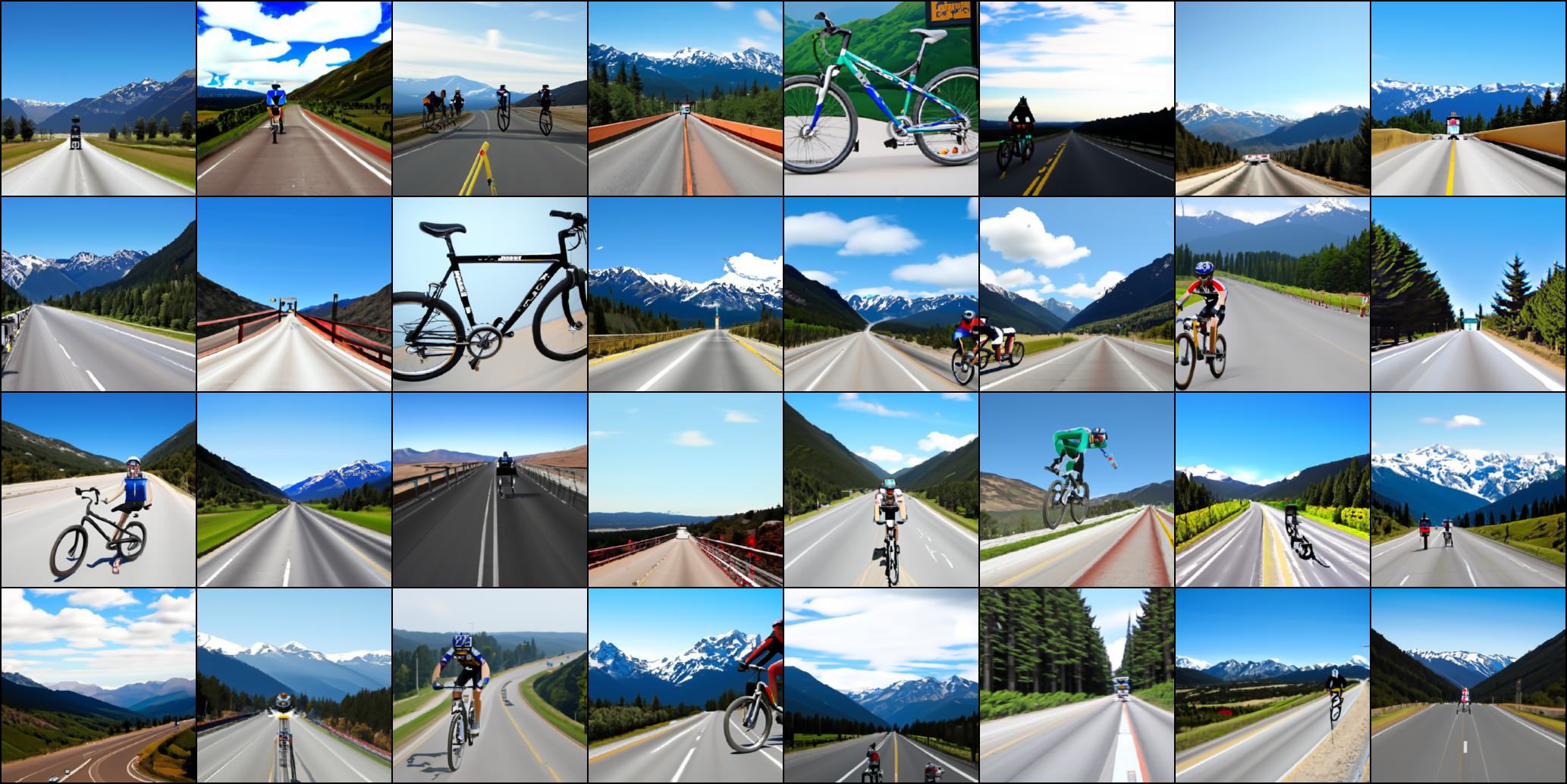}
	\centering
	\caption{Results for text "A highway" and ImageNet class "671: A bike". Not cherry-picked}
	\label{fig:t2im11}
	\vskip -10pt
\end{figure*}
 \begin{figure*}[t!]
	\centering
			\includegraphics[width=\textwidth]{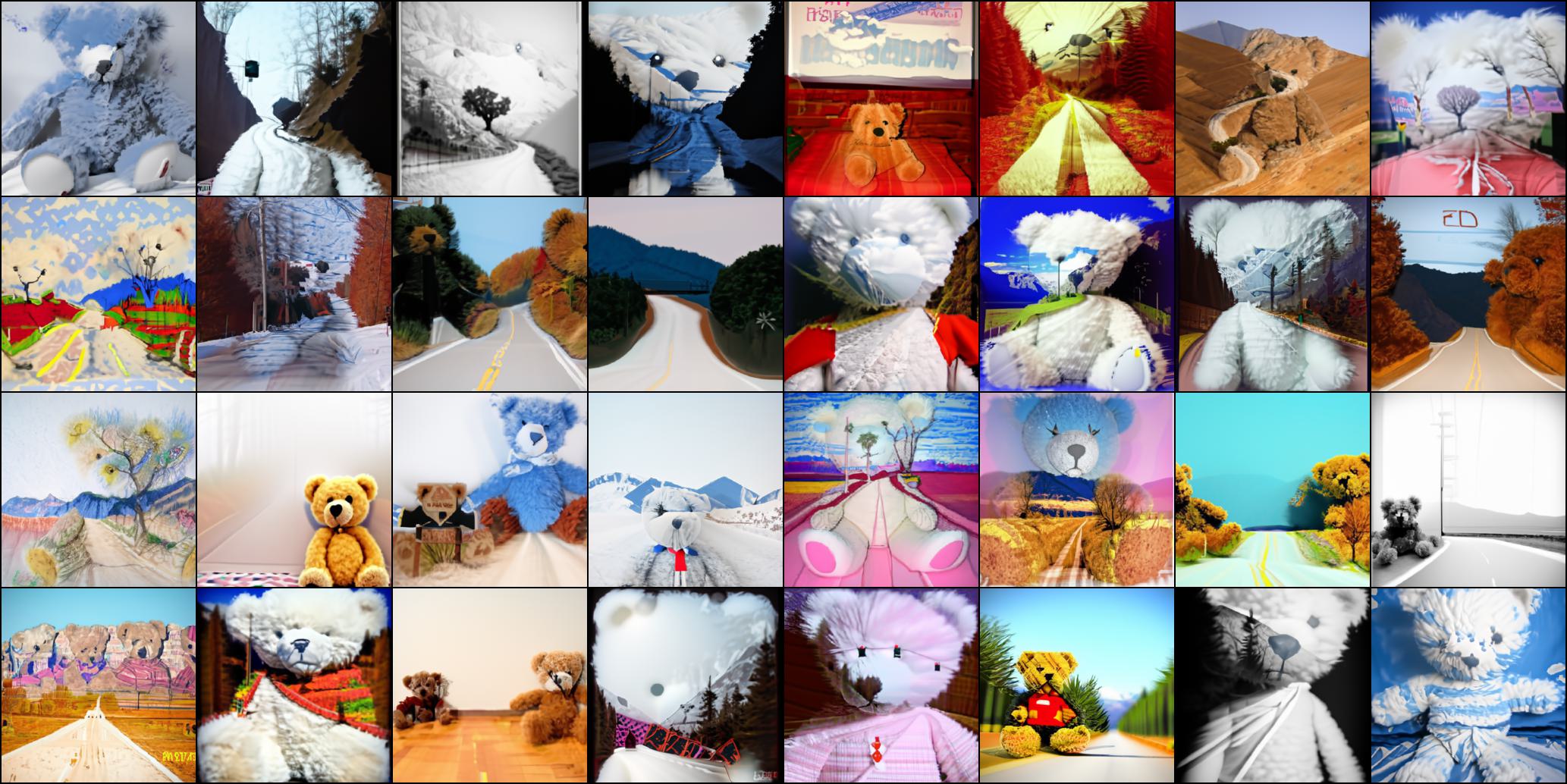}
	\centering
	\caption{Results for text "A road leading to mountains" and ImageNet class "850: Teddy bear". Not cherry-picked}
	\label{fig:t2im1a}
	\vskip -10pt
\end{figure*}
 \begin{figure*}[t!]
	\centering
			\includegraphics[width=\textwidth]{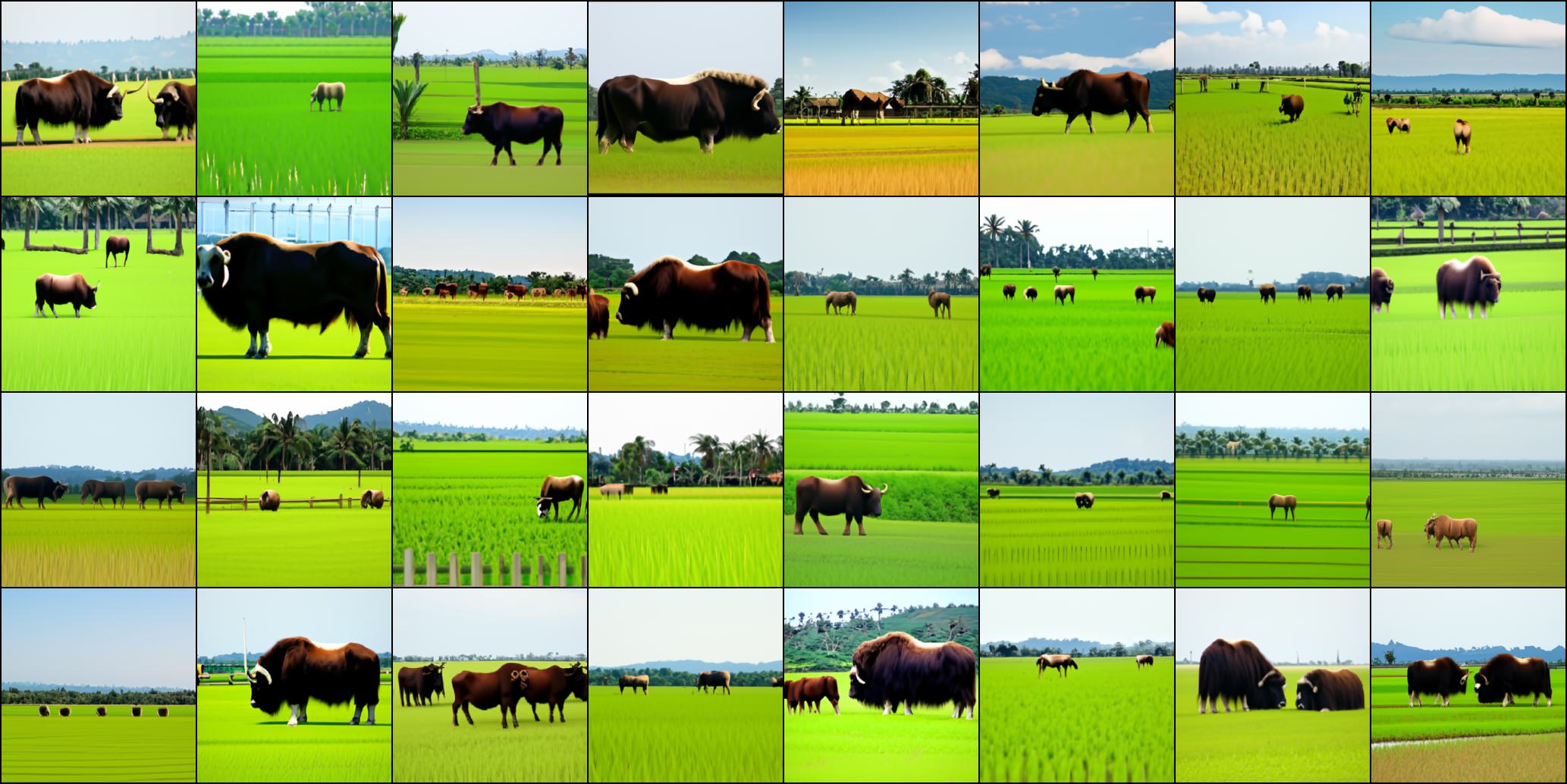}
	\centering

	\caption{Results for text "A paddy field" and ImageNet class "345: ox". Not cherry-picked}
	\label{fig:t2im1}
	\vskip -10pt
\end{figure*}
 \begin{figure*}[t!]
	\centering
			\includegraphics[width=\textwidth]{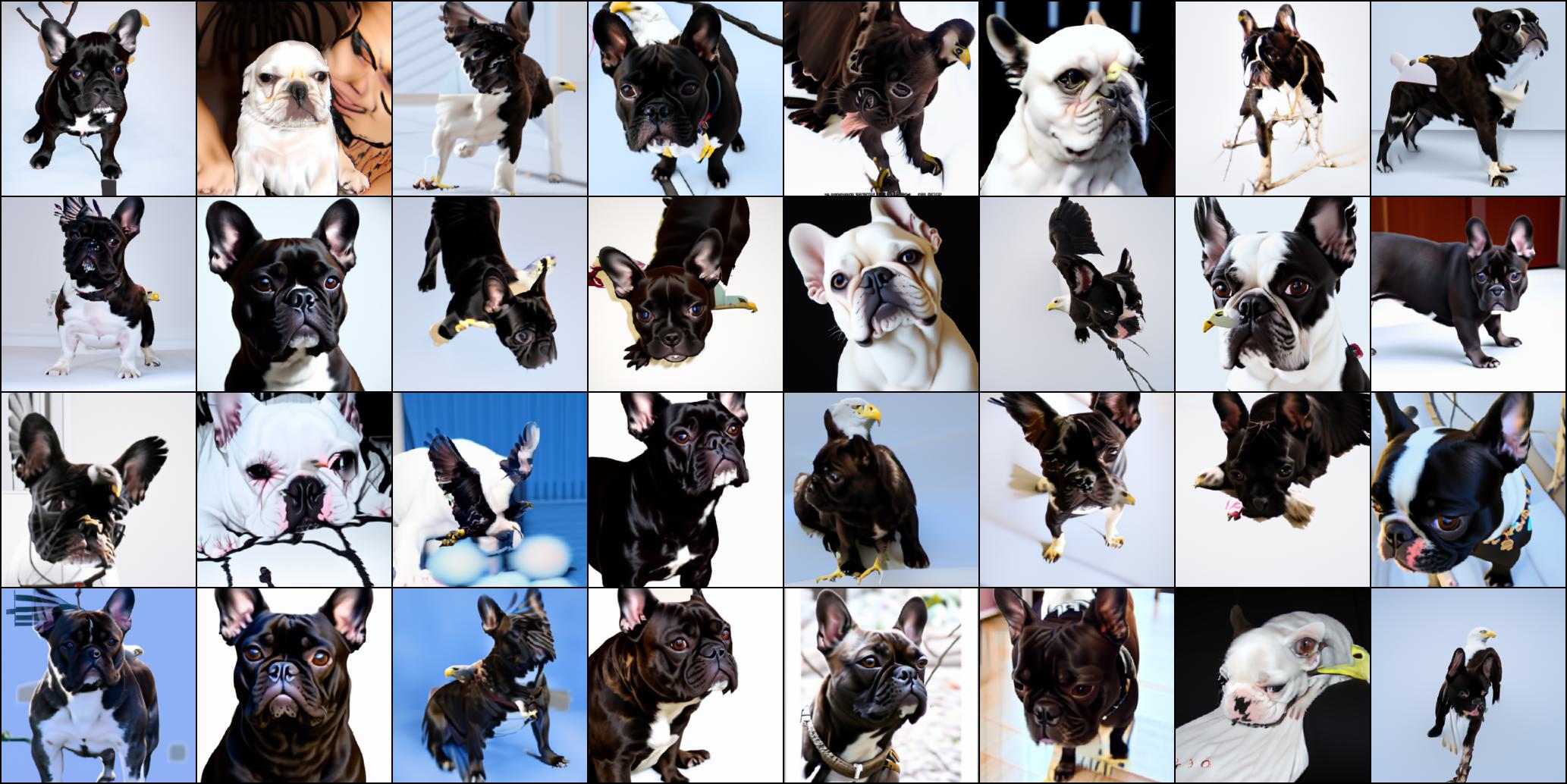}
	\centering

	\caption{Failure case: Contradictory inputs "An Eagle" and ImageNet class "245: Bulldog". }
	\label{fig:t2im1aa}
	\vskip -10pt
\end{figure*}
 \begin{figure*}[t!]
	\centering
			\includegraphics[width=\textwidth]{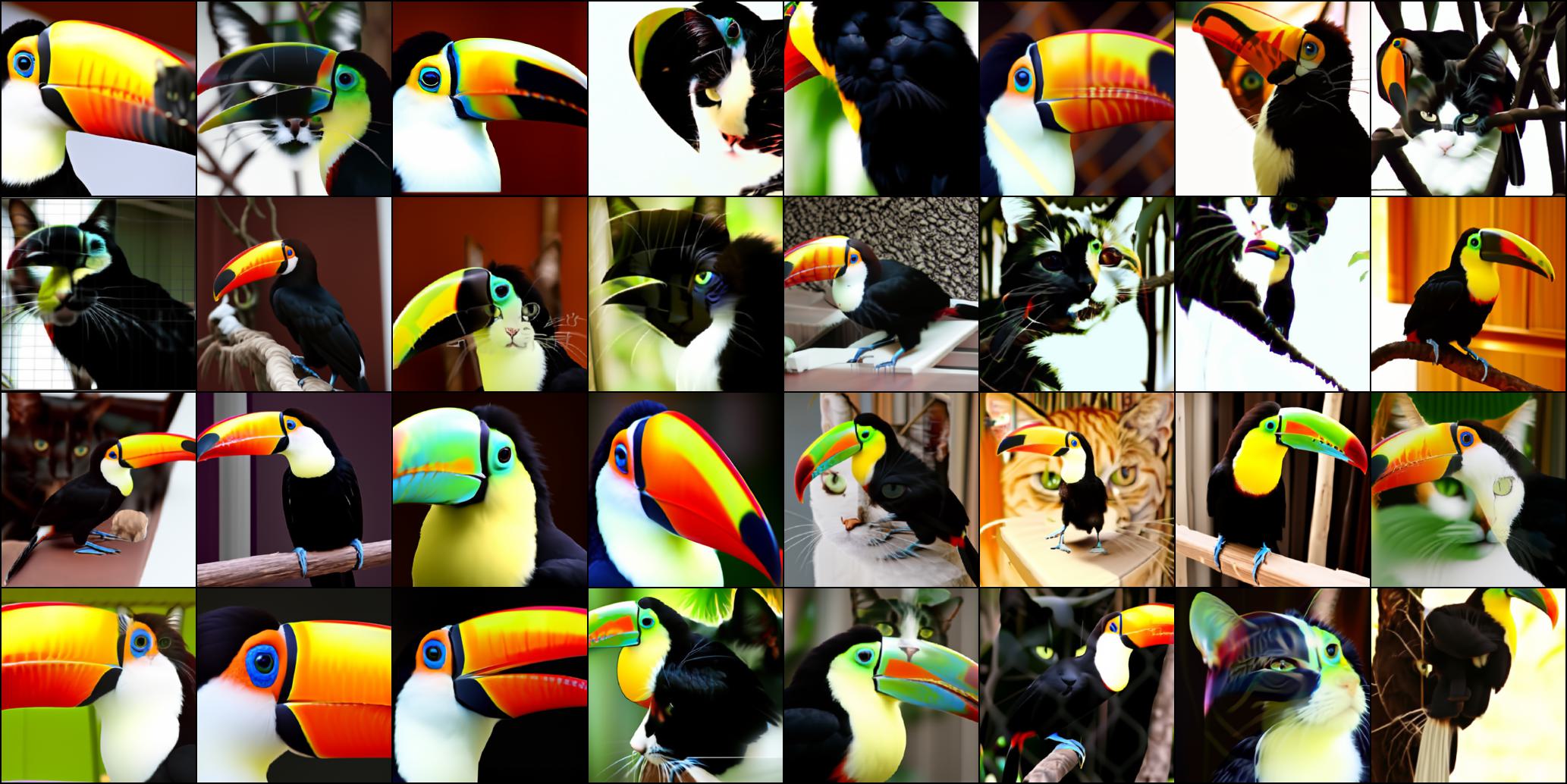}
	\centering

	\caption{Failure case: Contradictory input  "Photo of a cat" and ImageNet class "96: Toucan". }
	\label{fig:t2im1ab}
	\vskip -10pt
\end{figure*}
\begin{figure*}[t!]
    \centering
    \begin{subfigure}[t]{0.137\linewidth}
      \captionsetup{justification=centering, labelformat=empty, font=scriptsize}
  \includegraphics[width=1\linewidth]{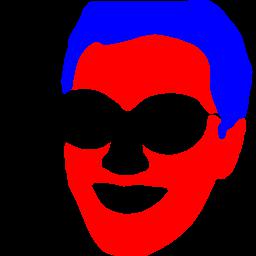}
   \includegraphics[width=1\linewidth]{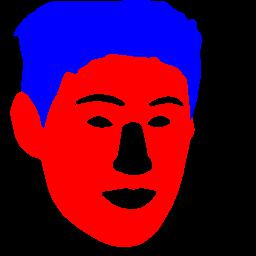}
      \includegraphics[width=1\linewidth]{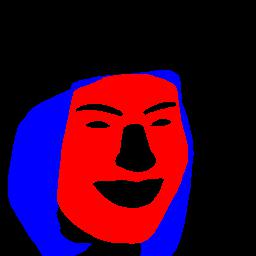}

        \includegraphics[width=1\linewidth]{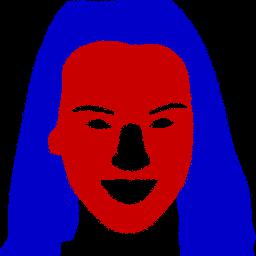}
      \includegraphics[width=1\linewidth]{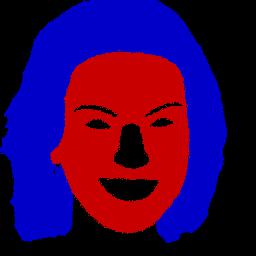}
      \caption{Semantic label}
    \end{subfigure}
    \begin{subfigure}[t]{0.137\linewidth}
      \captionsetup{justification=centering, labelformat=empty, font=scriptsize}
     \includegraphics[width=1\linewidth]{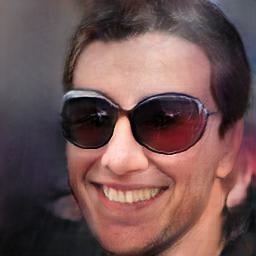}
   \includegraphics[width=1\linewidth]{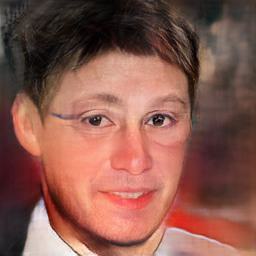}
      \includegraphics[width=1\linewidth]{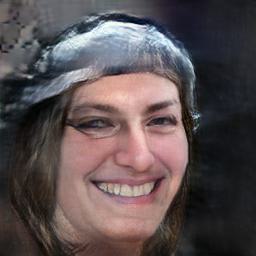}   
        \includegraphics[width=1\linewidth]{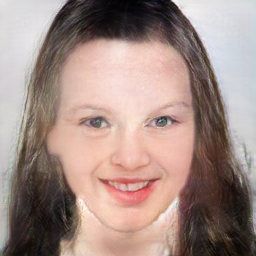}
      \includegraphics[width=1\linewidth]{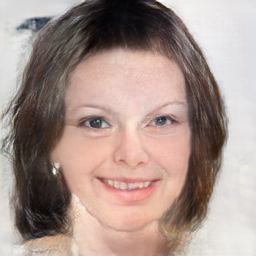}
      \caption{SPADE\cite{park2019SPADE}}
    \end{subfigure}
    \begin{subfigure}[t]{0.137\linewidth}
      \captionsetup{justification=centering, labelformat=empty, font=scriptsize}
      \includegraphics[width=1\linewidth]{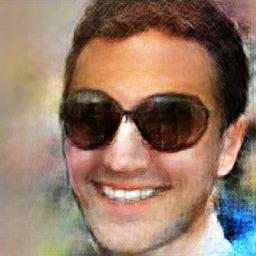}
   \includegraphics[width=1\linewidth]{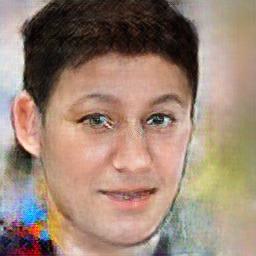}
      \includegraphics[width=1\linewidth]{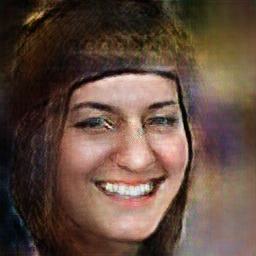}
        \includegraphics[width=1\linewidth]{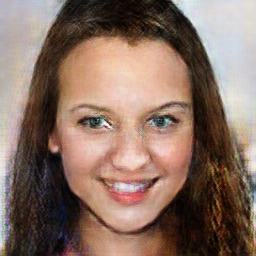}
      \includegraphics[width=1\linewidth]{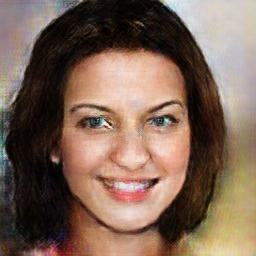}
      \caption{PIX2PIXHD\cite{wang2018pix2pixHD}}
    \end{subfigure}
    \begin{subfigure}[t]{0.137\linewidth}
      \captionsetup{justification=centering, labelformat=empty, font=scriptsize}
      \includegraphics[width=1\linewidth]{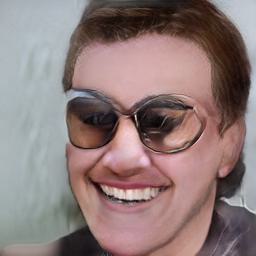}
   \includegraphics[width=1\linewidth]{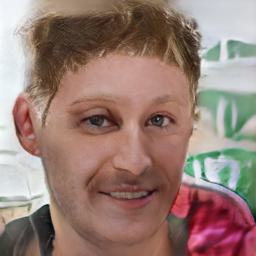}
      \includegraphics[width=1\linewidth]{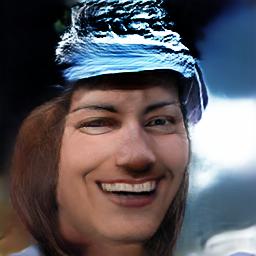}
         \includegraphics[width=1\linewidth]{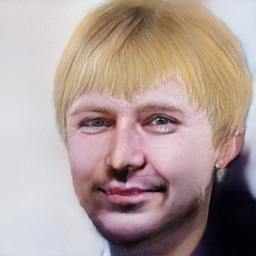}
      \includegraphics[width=1\linewidth]{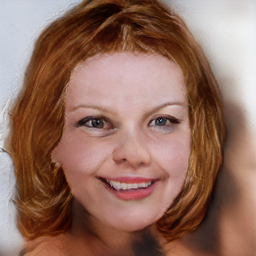}
      \caption{INADE\cite{tan2021diverse}}
    \end{subfigure}
    \begin{subfigure}[t]{0.137\linewidth}
      \captionsetup{justification=centering, labelformat=empty, font=scriptsize}
      \includegraphics[width=1\linewidth]{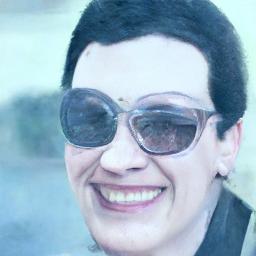}
   \includegraphics[width=1\linewidth]{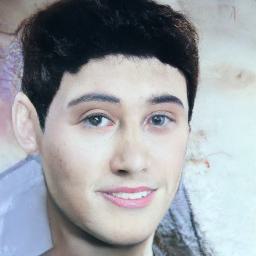}
      \includegraphics[width=1\linewidth]{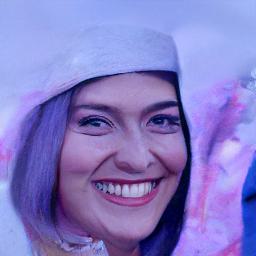}
        \includegraphics[width=1\linewidth]{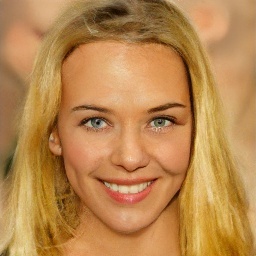}
      \includegraphics[width=1\linewidth]{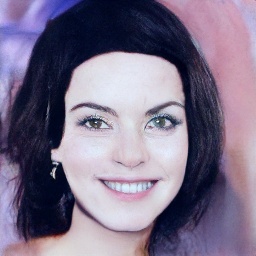}
      \caption{DDPM\cite{ho2020denoising}}
    \end{subfigure}
    \begin{subfigure}[t]{0.137\linewidth}
      \captionsetup{justification=centering, labelformat=empty, font=scriptsize}
        \includegraphics[width=1\linewidth]{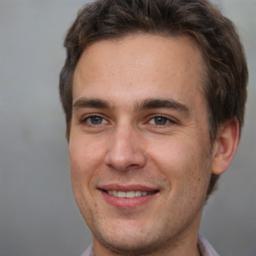}
   \includegraphics[width=1\linewidth]{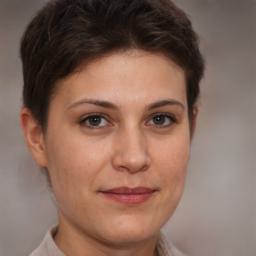}
      \includegraphics[width=1\linewidth]{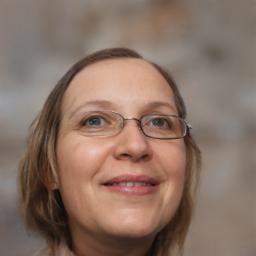}
        \includegraphics[width=1\linewidth]{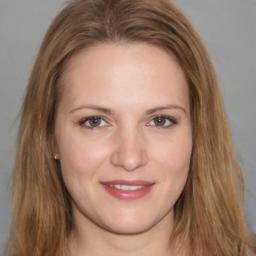}
      \includegraphics[width=1\linewidth]{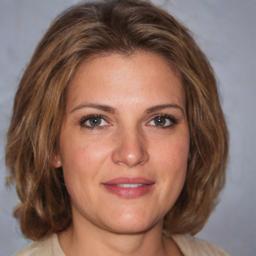}
      \caption{TediGAN\cite{xia2021tedigan}}
    \end{subfigure}
    \begin{subfigure}[t]{0.137\linewidth}
      \captionsetup{justification=centering, labelformat=empty, font=scriptsize}
        \includegraphics[width=1\linewidth]{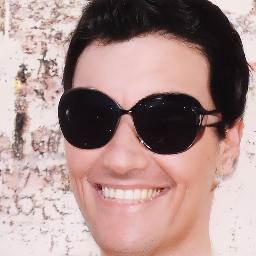}
   \includegraphics[width=1\linewidth]{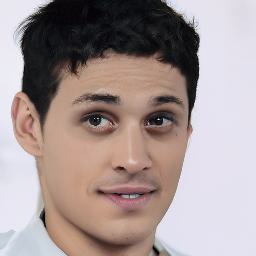}
      \includegraphics[width=1\linewidth]{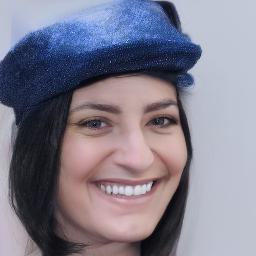}
        \includegraphics[width=1\linewidth]{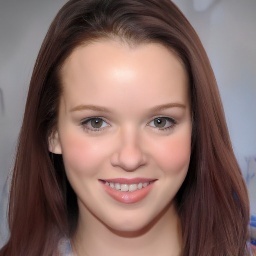}
      \includegraphics[width=1\linewidth]{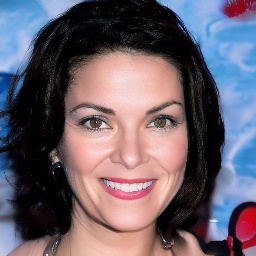}
      \caption{OURS}
    \end{subfigure}
    \vspace{-3mm}    \caption{\textbf{Qualitative comparisons for semantic to face generation.} In this case, a single model is trained by alternating different input datasets across different iterations. During Inference time all the modalities are taken from a single dataset and the proposed sampling technique is used.}
    \label{fig:facesematicsupp}
  \end{figure*}

\begin{figure*}[t!]
    \centering
    \begin{subfigure}[t]{\linewidth}
      \captionsetup{justification=centering, labelformat=empty, font=scriptsize}
  \includegraphics[width=1\linewidth]{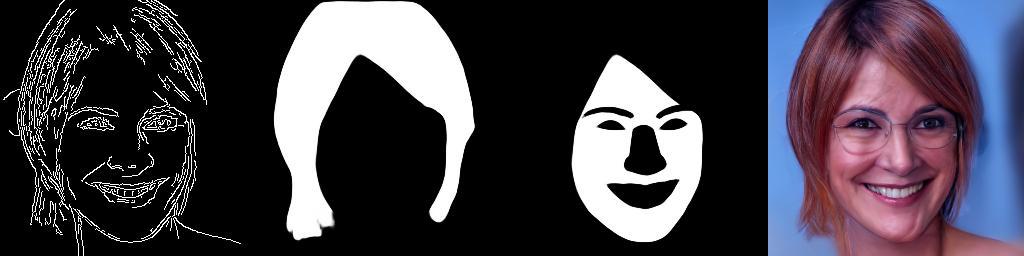}

    \end{subfigure}
       \begin{subfigure}[t]{\linewidth}
      \captionsetup{justification=centering, labelformat=empty, font=scriptsize}
  \includegraphics[width=1\linewidth]{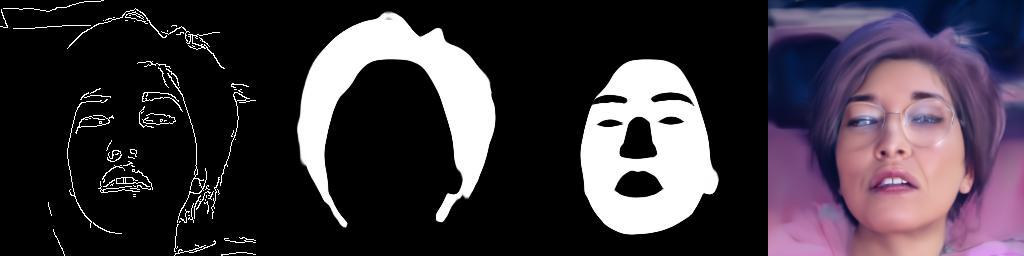}

    \end{subfigure}
      \begin{subfigure}[t]{\linewidth}
      \captionsetup{justification=centering, labelformat=empty, font=scriptsize}
  \includegraphics[width=1\linewidth]{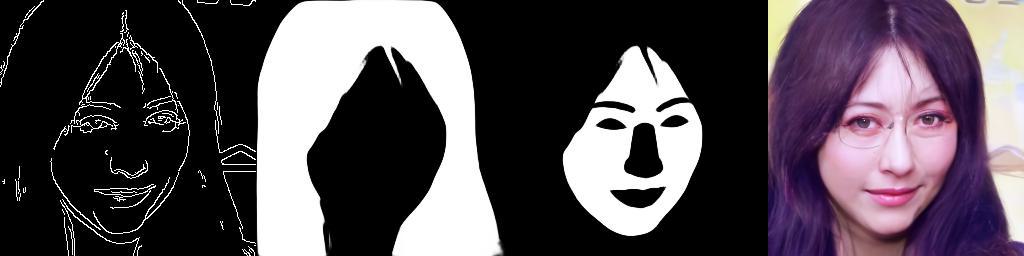}

    \end{subfigure}
     \begin{subfigure}[t]{\linewidth}
      \captionsetup{justification=centering, labelformat=empty, font=scriptsize}
  \includegraphics[width=1\linewidth]{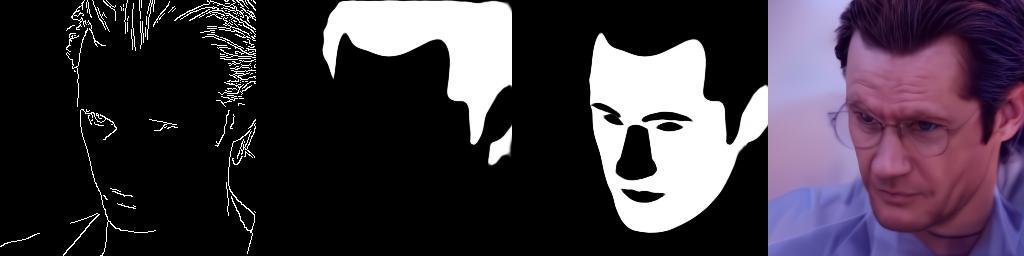}

    \end{subfigure}
\begin{subfigure}[t]{\linewidth}
      \captionsetup{justification=centering, labelformat=empty, font=scriptsize}
  \includegraphics[width=1\linewidth]{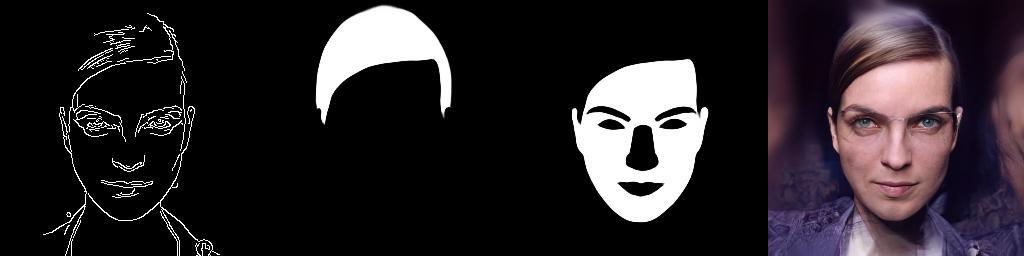}

    \end{subfigure}

    \vspace{-3mm}    \caption{\textbf{Multimodal face generation using four modalities} Text used: "An old person with  eyeglasses" }
    \label{fig:facesematicsupp3}
  \end{figure*}
\begin{figure*}[t!]
    \centering
    \begin{subfigure}[t]{\linewidth}
      \captionsetup{justification=centering, labelformat=empty, font=scriptsize}
  \includegraphics[width=1\linewidth]{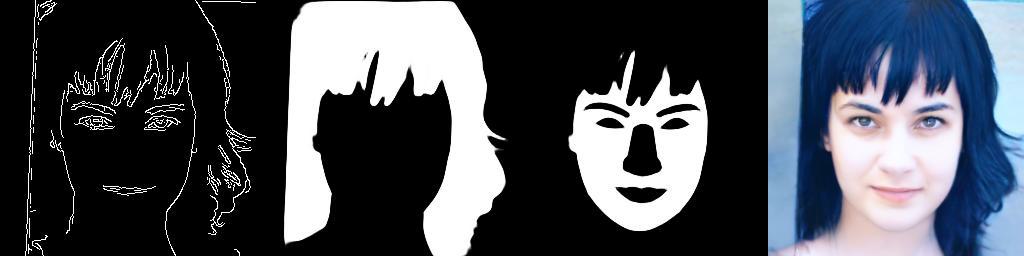}

    \end{subfigure}
       \begin{subfigure}[t]{\linewidth}
      \captionsetup{justification=centering, labelformat=empty, font=scriptsize}
  \includegraphics[width=1\linewidth]{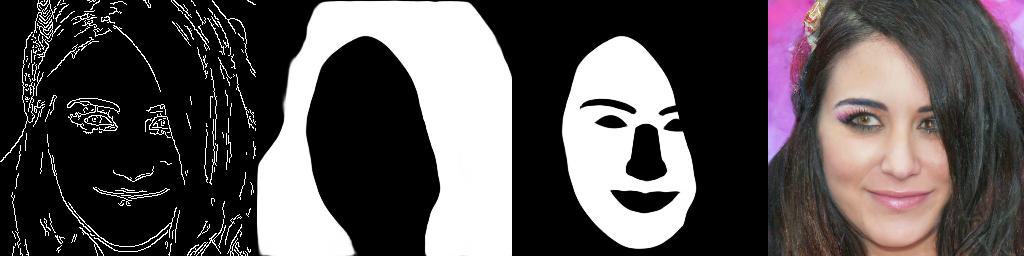}

    \end{subfigure}
      \begin{subfigure}[t]{\linewidth}
      \captionsetup{justification=centering, labelformat=empty, font=scriptsize}
  \includegraphics[width=1\linewidth]{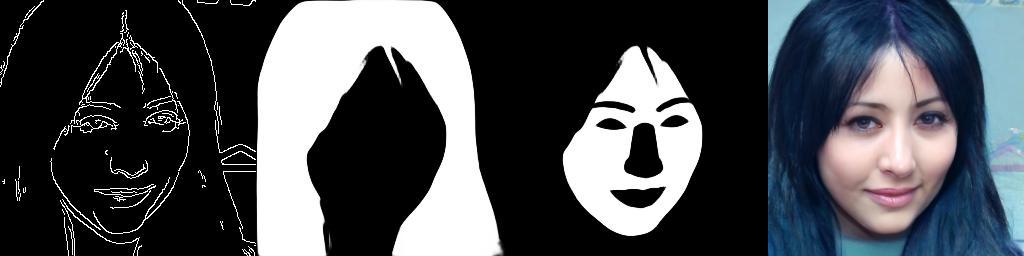}

    \end{subfigure}
     \begin{subfigure}[t]{\linewidth}
      \captionsetup{justification=centering, labelformat=empty, font=scriptsize}
  \includegraphics[width=1\linewidth]{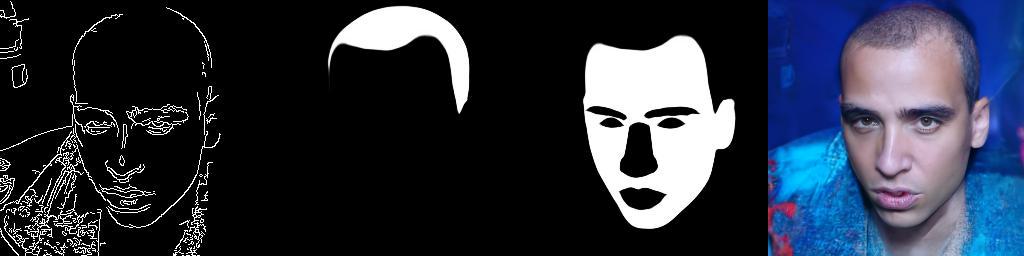}

    \end{subfigure}
\begin{subfigure}[t]{\linewidth}
      \captionsetup{justification=centering, labelformat=empty, font=scriptsize}
  \includegraphics[width=1\linewidth]{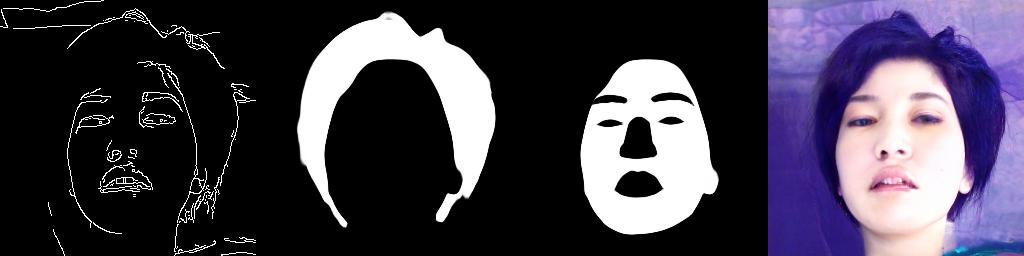}

    \end{subfigure}

    \vspace{-3mm}    \caption{\textbf{Multimodal face generation using four modalities} Text used: "A person with black hair" }
    \label{fig:facesematicsupp5}
  \end{figure*}

\begin{figure*}[t!]
    \centering
    \begin{subfigure}[t]{\linewidth}
      \captionsetup{justification=centering, labelformat=empty, font=scriptsize}
  \includegraphics[width=1\linewidth]{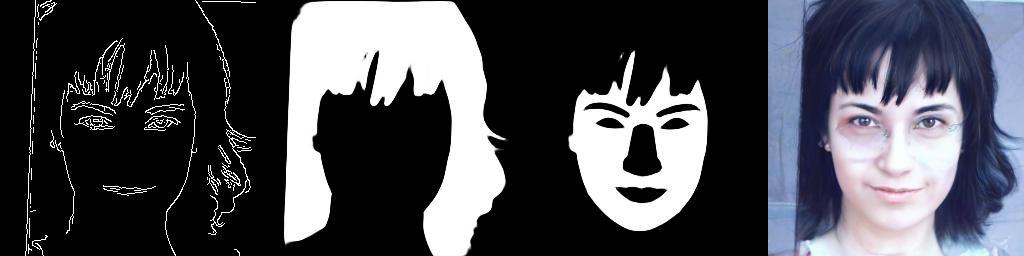}

    \end{subfigure}
       \begin{subfigure}[t]{\linewidth}
      \captionsetup{justification=centering, labelformat=empty, font=scriptsize}
  \includegraphics[width=1\linewidth]{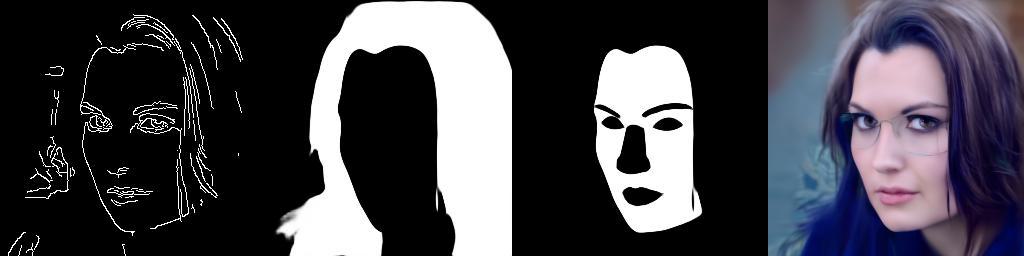}

    \end{subfigure}
      \begin{subfigure}[t]{\linewidth}
      \captionsetup{justification=centering, labelformat=empty, font=scriptsize}
  \includegraphics[width=1\linewidth]{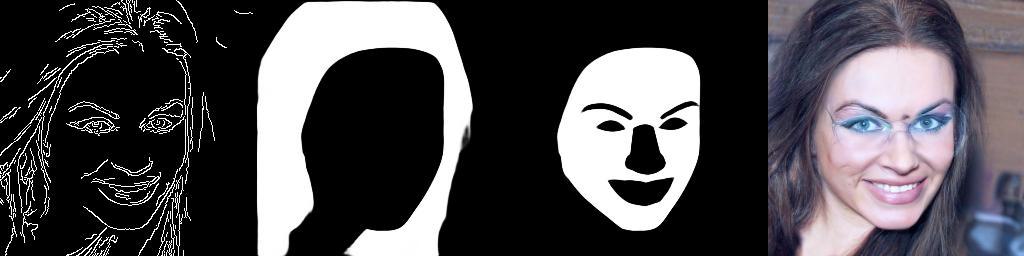}

    \end{subfigure}
     \begin{subfigure}[t]{\linewidth}
      \captionsetup{justification=centering, labelformat=empty, font=scriptsize}
  \includegraphics[width=1\linewidth]{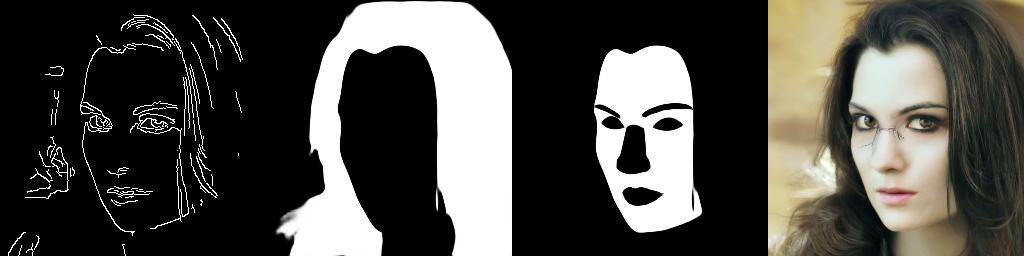}

    \end{subfigure}
\begin{subfigure}[t]{\linewidth}
      \captionsetup{justification=centering, labelformat=empty, font=scriptsize}
  \includegraphics[width=1\linewidth]{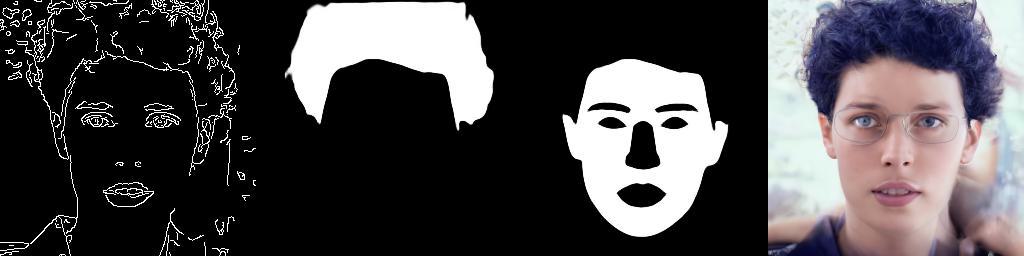}

    \end{subfigure}

    \vspace{-3mm}    \caption{\textbf{Multimodal face generation using four modalities} Text used: "A person with eyeglasses" }
    \label{fig:facesematicsupp1}
  \end{figure*}
\begin{figure*}[t!]
    \centering
    \begin{subfigure}[t]{\linewidth}
      \captionsetup{justification=centering, labelformat=empty, font=scriptsize}
  \includegraphics[width=1\linewidth]{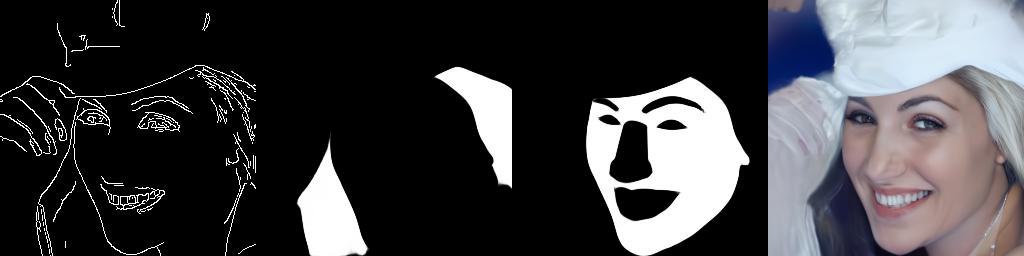}

    \end{subfigure}
       \begin{subfigure}[t]{\linewidth}
      \captionsetup{justification=centering, labelformat=empty, font=scriptsize}
  \includegraphics[width=1\linewidth]{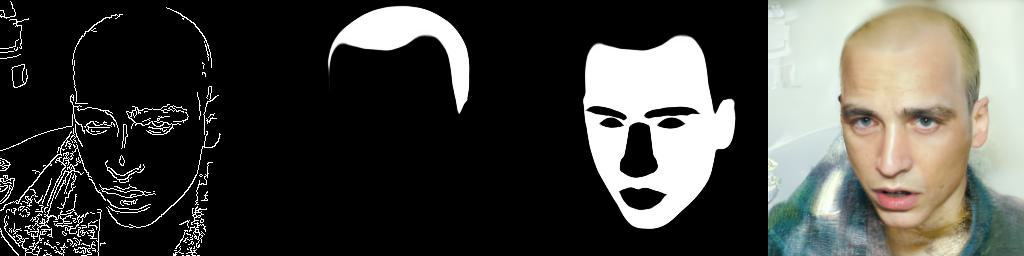}

    \end{subfigure}
      \begin{subfigure}[t]{\linewidth}
      \captionsetup{justification=centering, labelformat=empty, font=scriptsize}
  \includegraphics[width=1\linewidth]{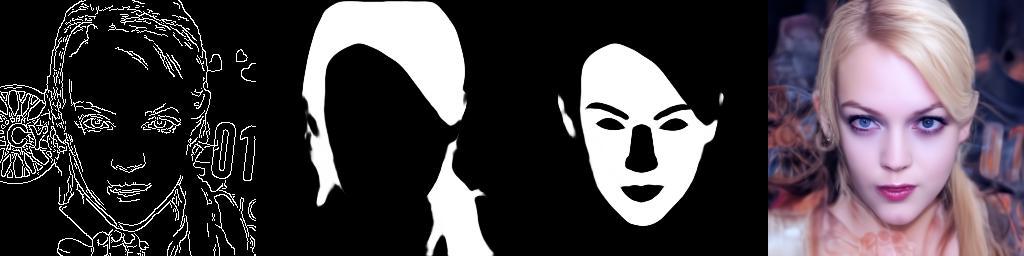}

    \end{subfigure}
     \begin{subfigure}[t]{\linewidth}
      \captionsetup{justification=centering, labelformat=empty, font=scriptsize}
  \includegraphics[width=1\linewidth]{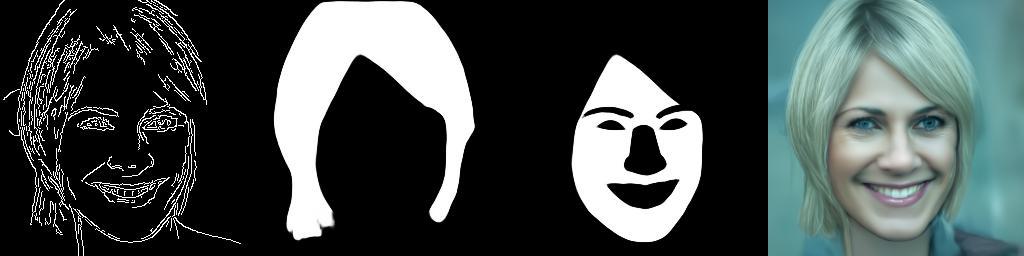}

    \end{subfigure}
\begin{subfigure}[t]{\linewidth}
      \captionsetup{justification=centering, labelformat=empty, font=scriptsize}
  \includegraphics[width=1\linewidth]{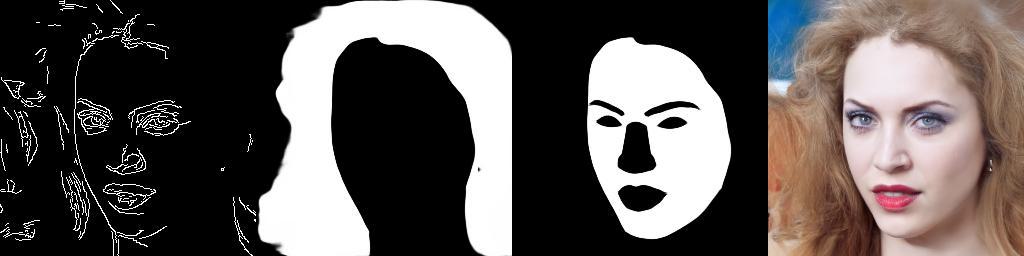}

    \end{subfigure}

    \vspace{-3mm}    \caption{\textbf{Multimodal face generation using four modalities} Text used: "A person with blonde hair" }
    \label{fig:facesematicsupp2}
  \end{figure*}
\begin{figure*}[t!]
    \centering
    \begin{subfigure}[t]{\linewidth}
      \captionsetup{justification=centering, labelformat=empty, font=scriptsize}
  \includegraphics[width=1\linewidth]{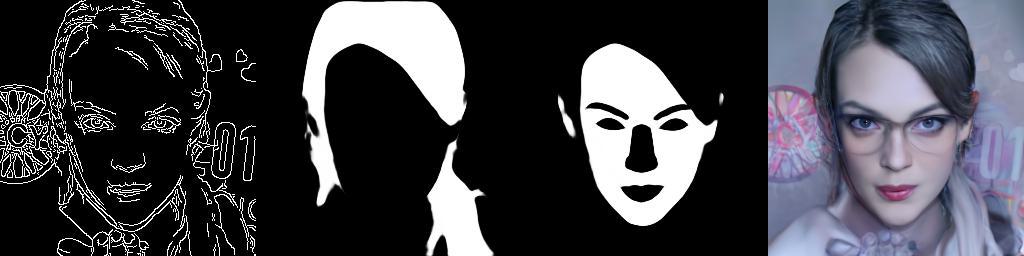}

    \end{subfigure}
       \begin{subfigure}[t]{\linewidth}
      \captionsetup{justification=centering, labelformat=empty, font=scriptsize}
  \includegraphics[width=1\linewidth]{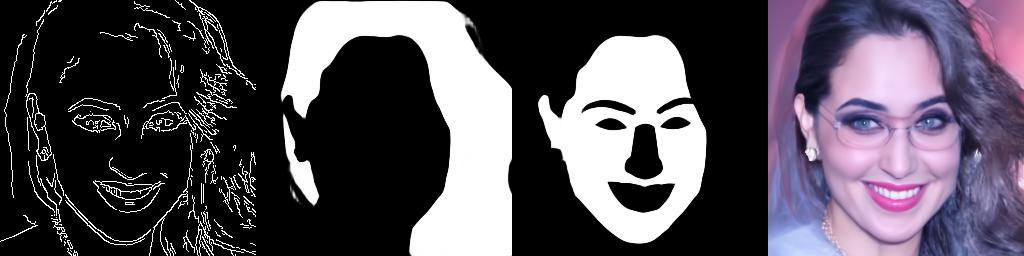}

    \end{subfigure}
      \begin{subfigure}[t]{\linewidth}
      \captionsetup{justification=centering, labelformat=empty, font=scriptsize}
  \includegraphics[width=1\linewidth]{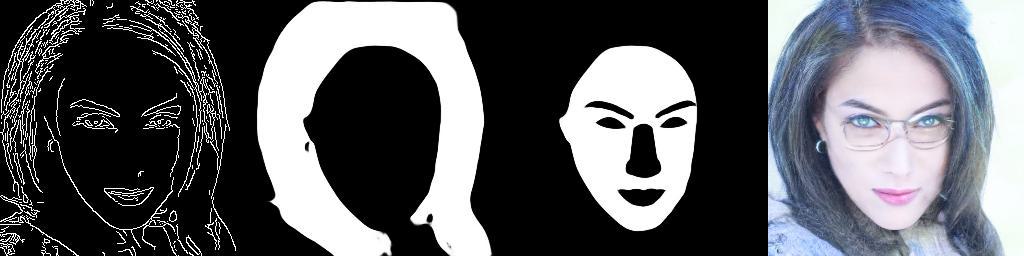}

    \end{subfigure}
    
    \vspace{-3mm}    \caption{\textbf{Multimodal face generation using four modalities} Text used: "This person has gray hair and wears eyeglasses" }
    \label{fig:facesematicsupp4}
  \end{figure*}
 \begin{figure*}[t!]
	\centering
			\includegraphics[width=\textwidth]{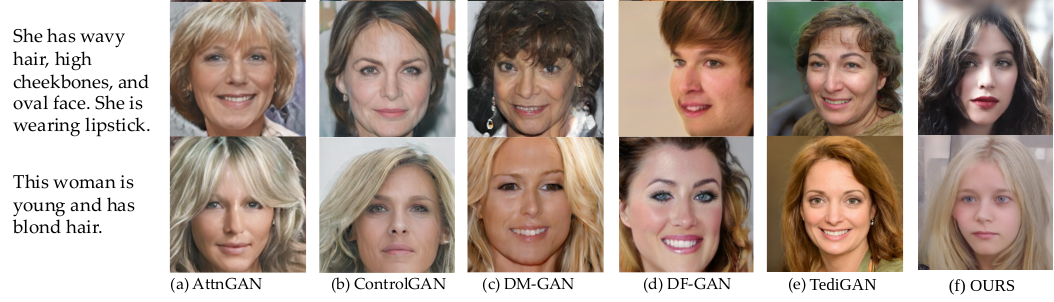}
	\centering
	\caption{Results for text to image generation on CelebA-HQ dataset.}
	\label{fig:t2im}
	\vskip -10pt
\end{figure*}

\end{document}